\documentclass[10pt]{article}
\usepackage[preprint]{tmlr}

\usepackage[T1]{fontenc}
\usepackage[utf8]{inputenc}
\usepackage{microtype}
\usepackage{amssymb}
\usepackage{inconsolata}
\usepackage{wrapfig}
\usepackage{algorithm}
\usepackage{float}
\usepackage{algorithmic}
\usepackage{booktabs}
\usepackage{amsmath}
\usepackage{mathtools}
\usepackage{graphicx}
\usepackage{subcaption}
\usepackage{enumitem}
\usepackage{booktabs}
\usepackage{multirow}
\usepackage{comment}

\usepackage{listings}
\usepackage[dvipsnames,svgnames,table]{xcolor}
\usepackage[colorlinks,allcolors=MidnightBlue]{hyperref}
\usepackage{url}
\usepackage{soul}
\usepackage{mdframed}

\definecolor{parametercolor}{RGB}{213,232,212}
\definecolor{querycolor}{RGB}{255,230,204}
\definecolor{contextcolor}{RGB}{248,206,204}

\definecolor{codegreen}{rgb}{0,0.6,0}
\definecolor{codegray}{rgb}{0.5,0.5,0.5}
\definecolor{codepurple}{rgb}{0.58,0,0.82}
\definecolor{backcolour}{rgb}{0.95,0.95,0.92}

\lstdefinestyle{mystyle}{
  commentstyle=\color{codegreen},
  keywordstyle=\color{magenta},
  numberstyle=\tiny\color{codegray},
  stringstyle=\color{codepurple},
  basicstyle=\ttfamily\footnotesize,
  breakatwhitespace=false,
  literate={\#}{\#}1,
  breaklines=true,
  captionpos=b,
  keepspaces=true,
  showspaces=false,
  showstringspaces=false,
  showtabs=false,
  tabsize=2,
  frame=lines,
  framerule=0.4pt,
  rulesepcolor=\color{gray},
  escapeinside={\%*}{*)}      
}
\usepackage[noabbrev,capitalize]{cleveref}

\crefname{page}{page}{pages}
\crefname{footnote}{footnote}{footnotes}
\crefname{equation}{equation}{equations}
\crefname{corollary}{Corollary}{Corollaries}
\crefname{line}{line}{lines}
\crefrangeformat{line}{lines #3#1#4--#5#2#6}
\crefname{lstlsting}{Listing}{Listings}
\crefname{section}{\S}{\S\S}
\Crefname{section}{\S}{\S\S}
\crefformat{section}{#2\S#1#3}
\Crefformat{section}{#2\S#1#3}
\crefrangeformat{section}{\S\S#3#1#4--#5#2#6}
\Crefrangeformat{section}{\S\S#3#1#4--#5#2#6}
\crefmultiformat{section}{\S#2#1#3}{ and~\S#2#1#3}{, \S#2#1#3}{ and~\S#2#1#3}
\Crefmultiformat{section}{\S#2#1#3}{ and~\S#2#1#3}{, \S#2#1#3}{ and~\S#2#1#3}
\crefrangemultiformat{section}{\S\S#3#1#4--#5#2#6}{ and~\S\S#3#1#4--#5#2#6}{, \S\S#3#1#4--#5#2#6}{ and~\S\S#3#1#4--#5#2#6}
\Crefrangemultiformat{section}{\S\S#3#1#4--#5#2#6}{ and~\S\S#3#1#4--#5#2#6}{, \S\S#3#1#4--#5#2#6}{ and~\S\S#3#1#4--#5#2#6}

\usepackage{xcolor}

\usepackage[disable]{todonotes}%
\makeatletter
\newcommand*\iftodonotes{\if@todonotes@disabled\expandafter\@secondoftwo\else\expandafter\@firstoftwo\fi}
\makeatother
\newcommand{\noindentaftertodo}{\iftodonotes{\noindent}{}\ignorespaces}
\newlength{\extramargin}
\usepackage{calc}
\iftodonotes{\usepackage[paperwidth=\paperwidth+\extramargin*2+10mm,marginparwidth=\marginparwidth+\extramargin,width=\textwidth,height=\textheight]{geometry}}{}

\newcommand{\safedrafttodo}[2][]{\ifinner\todo[inline,#1]{#2}\else\todo[#1]{#2}\fi}

\newcommand{\Fixme}[2][]{\noindentaftertodo}
\newcommand{\Notewho}[3][]{\noindentaftertodo}
\newcommand{\Jason}[2][]{\noindentaftertodo}
\newcommand{\Prabhav}[2][]{\noindentaftertodo}
\newcommand{\Haojun}[2][]{\noindentaftertodo}
\newcommand{\Tom}[2][]{\noindentaftertodo}
\newcommand{\Claude}[2][]{\noindentaftertodo}
\newcommand{\Cutforspace}[2][]{\noindentaftertodo}

\usepackage{xspace}

\newcommand{\mgruns}{10\xspace}
\newcommand{\defn}[1]{\textbf{#1}}
\newcommand{\localdefn}[1]{\textit{#1}}

\usepackage{bm}

\renewcommand{\vec}[1]{{\boldsymbol{\mathbf{#1}}}}
\newcommand{\vi}{\vec{i}}
\newcommand{\vX}{\vec{X}}
\newcommand{\vx}{\vec{x}}
\newcommand{\vR}{\vec{R}}
\newcommand{\vr}{\vec{r}}
\newcommand{\vS}{\vec{S}}
\newcommand{\vs}{\vec{s}}
\newcommand{\ve}{\vec{e}}
\newcommand{\vh}{\vec{h}}
\newcommand{\vw}{\vec{w}}
\newcommand{\vtheta}{\vec{\theta}}
\newcommand{\ptheta}{{p_\vtheta}}
\newcommand{\pthetahat}{{p_{\hat{\vtheta}}}}
\newcommand{\vphi}{\vec{\phi}}
\newcommand{\qphi}{{q_\vphi}}
\newcommand{\qphihat}{{q_{\hat{\vphi}}}}
\newcommand{\qphistar}{{q_{\vphi^*}}}
\newcommand{\vpsi}{\vec{\psi}}
\newcommand{\vpsihat}{\hat{\vpsi}}
\newcommand{\pprior}{p_{\mathrm{prior}}}

\newcommand{\edge}[1]{\raisebox{-1pt}{$\,\stackrel{#1}{\longrightarrow}\,$}}
\newcommand{\Rho}{\mathcal{R}}
\newcommand{\pa}{\mathrm{pa}}
\newcommand{\vZ}{\vec{Z}}
\newcommand{\vz}{\vec{z}}
\newcommand{\vt}{\vec{t}}

\newcommand{\Dom}{\mathcal{D}}
\newcommand{\taskloss}{\mathrm{loss}}
\newcommand{\risk}{\mathrm{risk}}
\newcommand{\voi}{\mathrm{VOI}}
\newcommand{\Test}{\textsf{Test}}
\newcommand{\Train}{\textsf{Train}}

\DeclareMathOperator*{\argmin}{argmin}
\DeclareMathOperator*{\argmax}{argmax}
\DeclarePairedDelimiter\parens{(}{)}
\DeclarePairedDelimiterX{\divx}[2]{(}{)}{#1\;\delimsize\|\;#2}

\newcommand{\defeq}{\mathrel{\stackrel{\textnormal{\tiny def}}{=}}}
\newcommand{\psup}[1]{^{(#1)}}
\newcommand{\E}[2][]{\mathop{\mathbb{E}}_{{#1}}\left[#2\right]}

\renewcommand{\H}[1]{\mathrm{H}\parens{#1}}
\newcommand{\KL}[2]{\mathrm{KL}\divx{#1}{#2}}
\DeclarePairedDelimiter\norm{\lVert}{\rVert}

\newcommand{\obs}[1]{\check{#1}}
\newcommand{\mask}[1]{\check{\check{#1}}}
\newcommand{\wildcard}{\top}

\title{\textsc{Marformer}: A Transformer for Predicting\\ Missing Data Distributions}

\author{\name Prabhav Singh\thanks{Equal contribution, order determined randomly.  The work was conducted while all authors were at Johns Hopkins University.} \email prabhavs@cs.utexas.edu \\
   \addr School of Computing, College of Natural Sciences\\
   The University of Texas at Austin, Austin TX
   \AND
   \name Xiheng Tom Wang\footnotemark[1] \email xwang397@jh.edu \\
   \addr Department of Computer Science \\
   Johns Hopkins University, Baltimore MD
   \AND
   \name Haojun Shi\footnotemark[1] \email haojun.shi@yale.edu \\
   \addr Department of Computer Science \\
   Yale University, New Haven, CT
   \AND
   \name Jason Eisner
   \email jason@cs.jh.edu \\
   \addr Department of Computer Science \\
   Johns Hopkins University, Baltimore MD}

\def\month{MM}
\def\year{YYYY}
\def\openreview{\url{https://openreview.net/forum?id=XXXX}}

\begin{document}

\maketitle

\begin{abstract}
Real decisions are made under incomplete information.  If we observe only some of the random variables we need, we can predict the others.
The \defn{conditional marginals} over the missing variables are the key ingredient for computing Bayes risk and Value of Information (VOI), the expected gain from acquiring one more observation before deciding. We present the \textsc{Marformer}, a Transformer trained to directly predict conditional marginals
given any set of observed values.
Like \textsc{BERT}, which is trained to predict missing words from context, the \textsc{Marformer} constructs a hidden-vector representation for each distribution $p(X_i)$
and iteratively refines it through attention to other distributions $p(X_j)$. Unlike generative approaches, the \textsc{Marformer} does not model the full joint distribution,
requires no domain knowledge of the data-generating process, and makes all predictions in a single forward pass.
We evaluate across three synthetic domains with missing data---Bayesian networks, discretized multivariate Gaussians, and structured annotation data.  The \textsc{Marformer} can match or outperform classical missing-data methods, even when those methods are given the true model family and prior that generated the synthetic data.
We also evaluate on a real annotation dataset, where the \textsc{Marformer} outperforms the evaluated baselines at the largest training size. In both cases,
the \textsc{Marformer} is substantially faster than
the evaluated generative baselines.
\end{abstract}

\section{Introduction}
\label{sec:intro}

Incomplete data is the rule, not the exception, in scientific and applied
settings.
A recommender system has a user's opinions of only a few movies.
A standardized testing agency has only a few graders' opinions of each essay.
The incomplete data may be acquired adaptively:
A doctor does just enough tests to choose a treatment confidently.
An engineer runs cheap simulations to decide which candidate molecules or hyperparameter settings are worthy of further measurements.
An educational tester or job interviewer probes adaptively for strengths and weaknesses, using a limited budget.
An evaluator of an AI system adaptively selects
examples based on observed performance, using the results to predict performance on
other examples \citep{li2024active}.
In each case, the decision-maker needs not a point estimate of the missing values but a predictive \emph{distribution} over them
given the available evidence.
With such distributions, one can quantify residual uncertainty, predict the risk of a possible action,
or determine whether gathering more data before acting would be worth the cost.

In this paper we focus on  \emph{marginal} predictions.
Our goal is to train a neural network that can predict the marginal distribution of any \emph{single} missing value conditioned on \emph{any subset} of the remaining values.  We fit these conditional marginals directly by conditional maximum likelihood, a frequentist approach that requires no prior over the model parameters.  As we explain in \cref{sec:why-marginals}, such a model is directly useful for deciding among real-world actions and for choosing what additional information (missing values) to acquire before this decision.  Our architecture is generic and offers fast, amortized, trainable inference of all the \emph{individual} missing values in parallel.
We also point out in \cref{sec:why-marginals} that such a model can jointly impute \emph{multiple} missing values, if needed, by sampling them sequentially.  Thus, our (possibly incoherent) collection of conditional distributions contains as much information as a joint distribution---but in a form designed to support conditional inference.

The classical \defn{Expectation-Maximization} approach
proceeds in the opposite direction: from joint to conditional.  It \emph{first} obtains a joint distribution over all values by fitting a full generative model to the incomplete data, using the maximum likelihood principle \citep[EM:][]{dempster-em} or the maximum a posteriori principle if a prior is available \citep[MAP-EM:][\S1.6.1]{McLachlan+Krishnan:1997}.
One then performs inference as needed to extract the desired conditional marginals from this joint distribution.

Unfortunately, EM or MAP-EM (i)~requires domain expertise to specify the generative model family and any prior; (ii)~can encounter local optima when fitting the model parameters; and (iii)~may require intractable inference, necessitating variational or sampling-based approximations that can be slow or inaccurate.

A more sophisticated variant is \defn{Bayesian posterior imputation} \citep{tanner-wong-1987,gelfand-smith-1990}, which integrates out the posterior uncertainty about the generative model's parameters, typically by using MCMC to (approximately) sample them jointly with the missing values.  The samples of the missing data can then be used to estimate the desired conditional marginal distributions.\footnote{Known in this Bayesian context as posterior predictive distributions.}  Training (ii)
is thus absorbed into inference (iii), whose MCMC will escape local optima if run long enough.  However, training/inference is still expensive to do accurately, and model specification (i) is still challenging.

In short, these classical methods have serious practical weaknesses---as we will demonstrate experimentally.  The model
is commonly \emph{misspecified} because the domain is poorly understood or challenging
to model.  Such violations of (i) can lead to inaccurate results.  (ii) and (iii) mean that even a \emph{correctly specified} model family may produce poor results.
Finally, (iii) slows down training as well as runtime prediction.  This is particularly burdensome for online training: e.g., if the model is used to guide the acquisition of missing observations (as we discuss in \cref{sec:why-marginals}), it is natural to immediately use these to improve the model parameters, so training continues indefinitely.

\begin{figure}[h!]
  \centering
  \includegraphics[width=0.8\textwidth]{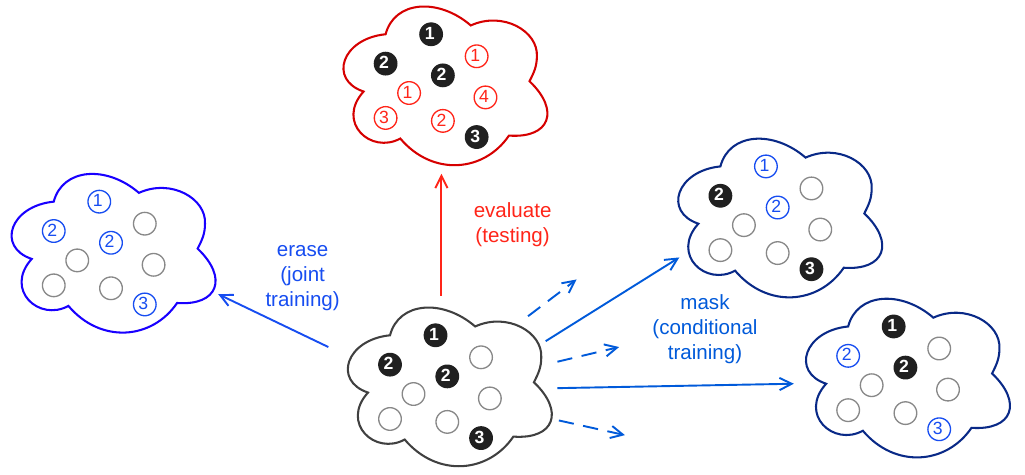}
  \caption{The complete instance $\vx$ at top reveals values only for attributes in $\vr$ (black), where $(\vx,\vr) \sim P^*$.  Thus, our methods receive only the incomplete instance $\vx_\vr$ at bottom.  In this example, $\vx_\vr$ is used both as a test instance and as a training instance (transductive learning).  At test time, a model is evaluated (top) by its ability to predict individual missing values (red) from the observed values $\vx_\vr$ (black).  It predicts a distribution for each red attribute and is rewarded if this distribution assigns high probability to the true value.  At training time, the \textsc{Marformer} (right) trains to predict masked values (blue) from seen values $\vx_\vs$ (black).  Many different masks $\vs$ are used.  In contrast, a classical generative model (left) trains to jointly predict all observed variables (blue) from nothing, using the missing variables (white) as latent variables.}\label{fig:train-eval}
\end{figure}

This is why we propose to abandon the hypothesized generative model---even when it is correct!---and simply fit all the conditional marginals directly.  For this we use a generic neural model with enough capacity to do the job (one can select the neural network's size via cross-validation).
Deep learning has been remarkably successful at fitting naturally occurring distributions. In contrast to the classical methods, training an overparameterized neural network does not adopt restrictive modeling assumptions,\footnote{This resembles the argument for nonparametric statistical tests: ``[E]ven when the parametric assumptions hold perfectly true, we will see that nonparametric methods are only slightly less powerful [i.e., statistically efficient] \ldots Furthermore, if the parametric assumptions about
the data fail to hold, only the nonparametric method is valid'' \citep{kvam2023nonparametric}.}
may bring useful inductive bias of its own \citep{kawaguchi2022generalization}, tends not to get trapped in bad local minima \citep{sun2020global}, and can be designed to directly predict the target distributions, making inference fast and end-to-end trainable.

To flesh out this program, we propose a specific neural architecture that we hope is suitable for many domains: a \pmb{\textsc{Marformer}} (Conditional \underline{Mar}ginal Trans\underline{former}). The architecture is illustrated in \cref{fig:marformer}.
Its key properties are:

\begin{enumerate}[nosep]
\item \textbf{Observation of values.}  Each input to the \textsc{Marformer} specifies a set of random variables together with their values, some of which may be missing.

\item \textbf{Prediction of parameters.} The \textsc{Marformer} outputs marginal distributions for all variables, conditioned on the observations in the input.  More precisely, for each variable, it outputs a parameter vector that will be interpreted as specifying a predictive distribution over that variable's domain.

\item \textbf{Conditional log-likelihood maximization.}  The training objective is conditional log-likelihood.  That is, the predicted marginal over an unknown quantity should assign a high log-probability (or log-density) to its true value.  This objective encourages predictions to match the true conditional distributions, which can subsequently be used to evaluate decisions under any task-specific loss \citep{gneiting2007strictly}. We discuss empirical calibration in \cref{sec:results-annot}.

\item \textbf{Amortized, single-pass inference.} Once trained, the \textsc{Marformer} predicts all missing marginals simultaneously in a single forward pass, sharing computation across the entire set of missing values.  Its architecture (\cref{fig:marformer-arch,fig:marformer-block}) generalizes BERT \citep{devlin-etal-2019-bert}, which specifically predicts conditional marginals over all missing words in a natural-language sentence.  As in BERT, each variable can attend to the evolving predictions of the other variables as their representations are transformed layer by layer.

\item \textbf{End-to-end differentiable training.}
Inference is fully defined by this forward pass.  Thus, training the \textsc{Marformer} to maximize conditional log-likelihood is explicitly optimizing the parameters to make inference accurate \citep{pmlr-v15-stoyanov11a,hershey2014deep}.  In contrast, classical approaches train a joint model without considering the errors that may later be introduced by approximate inference.

\item \textbf{Support for relational domains.}  Many machine learning methods train and test on \emph{many} partially observed samples that are assumed i.i.d.  However, a \defn{relational} task domain may have only a \emph{single} partially observed sample with very many interrelated variables: the observed variables are observed only once, and the missing variables are never observed.  Predicting the missing values is still possible, via shared parameters that model recurring relationships among variables.  For \emph{joint} modeling of relational domains, many suitable formalisms have been developed \citep{SRL-2007}, although inference is generally expensive because it requires some form of message passing among the many variables.  We extend the \textsc{Marformer} architecture (\cref{sec:architecture}) to also handle such domains, using a \defn{relational attention} mechanism \citep{shaw2018self} whereby related variables know their relationships and can learn to attend to one another.

\end{enumerate}

Now the obvious question: How can we train a \textsc{Marformer}?  Specifically, to supervise the prediction of missing data, where do we acquire the true values?
The self-supervised learning recipe
\citep[e.g.,][]{devlin-etal-2019-bert,he2021maskedautoencodersscalablevision}
randomly masks out some variables, and trains a model to predict their masked values.  The traditional recipe starts with \emph{complete} data, and often trains prediction only for its side effect of representation learning.  In contrast, we start with \emph{incomplete} data, and predicting missing values is our actual goal.  Yet \cite[like][]{du2023remaskerimputingtabulardata} we can follow the same masking recipe---randomly mask out some of the observed values and train the model to predict them from the remaining observed values (\cref{fig:train-eval}).  We hope that the \textsc{Marformer} model will generalize from predicting these masked values to predicting the originally missing values.  As we show in \cref{sec:training}, this is reasonable to expect \emph{provided that} the originally missing variables are missing completely at random (MCAR) and we mask additional variables using a missing-at-random (MAR) mechanism.  (We review missingness mechanisms in \cref{sec:missingdefs}.)

Our main synthetic evaluations use MCAR data, for which this masking recipe has the
justification given in \cref{sec:training}. We also report exploratory MAR evaluations in
\cref{sec:exp-mar}, where that justification does not automatically apply, and evaluate on
a naturally sparse real annotation dataset (\cref{sec:data-real}), for which MCAR is an unverified assumption.  In a sequel paper, we will generalize to training {\textsc{Marformer}}s on MAR data---the same setting where EM works---using a more complex training recipe that is beyond the scope of the present paper.\footnote{Like EM, that recipe
uses the current model to impute missing values.  We will impute using the \textsc{Marformer} itself.%
}

The present paper suffices to study the \textsc{Marformer} architecture itself, showing that it can successfully model many conditional distributions at once.  It can learn from modestly sized datasets, and it generalizes well from predicting masked values at training time to predicting the actually missing values at test time.  Most important, it is competitive with the classical approaches
even when they know the true model and prior.

\Cref{sec:problem-formulation} motivates and formalizes the problem. \Cref{sec:methods} describes the architecture and training. \Cref{sec:domain-bn,sec:domain-mg,sec:domain-annot} present experiments in three synthetic domains of increasing complexity. \Cref{sec:domain-real} evaluates the method on real annotation data. \Cref{sec:related} discusses other related work, and \cref{sec:conclusions} discusses limitations and future directions.

\section{Problem Setting}
\label{sec:problem-formulation}

\subsection{Notation for Attributes and Values}
\label{sec:notation}

We assume that our training and test instances are i.i.d.\@ samples from some unknown probability distribution $P^*$. To describe an instance's attributes, the underlying
sample space is equipped with a collection $\{X_i\}_{i \in \vi}$ of random variables named by the elements $i$ of the fixed index set $\vi$.
Each random variable $X_i$
has some domain $\Dom_i$.
We write $\vX$ for the entire random collection $\{X_i\}_{i \in \vi}$, which has domain $\prod_{i\in\vi} \Dom_i$.  For example, if $\vi=\{1,2,3\}$, then $\vX$ can be identified with the random tuple $(X_1,X_2,X_3)$ with values of the form $\vx=(x_1,x_2,x_3) \in \Dom_1 \times \Dom_2 \times \Dom_3$.

For $\vx$ in the domain of $\vX$ and
$\vs \subseteq \vi$, let $\vx_\vs$ denote the subcollection $\{x_i\}_{i \in \vs}$: that is, a dictionary or function that maps each \defn{attribute} $i\in\vs$
to its \defn{value} $x_i$.  (Note that one may recover $\vs$ from $\vx_\vs$ as the domain of this map.)  Applying this subscripting operation to $\vX$, we obtain the corresponding random subcollection $\vX_\vs$ with domain $\prod_{i\in\vs} \Dom_i$.  Note that $\vx_\vi=\vx$ and $\vX_\vi=\vX$.

By abuse of notation, even when $\vx$ does not refer to any object, we may use $\vx_\vs$ to refer to a possible value of $\vX_\vs$ (and $x_i$ to refer to a possible value of $X_i$), implying that $\vx_\vs \in \prod_{i\in\vs} \Dom_i$ (and $x_i \in \Dom_i$).

\subsection{Conditional Marginals and Their Uses}\label{sec:why-marginals}\label{sec:afa}

Our goal is to train a single \textsc{Marformer} to model \emph{all} of the conditional marginal distributions $P^*(X_i \mid \vX_\vs=\vx_\vs)$, where $\vs \subsetneq \vi$, $i \in \vi \setminus \vs$, and $\vx_\vs \in \prod_{j\in\vs} \Dom_j$.  The conditioning values $\vx_\vs$ are said to be \defn{seen} and are provided to the \textsc{Marformer} in its input.\footnote{This generalizes the task of modeling $P^*(X_i \mid X_1=x_1, \dots, X_{i-1}=x_{i-1})$---the central problem of \defn{prequential statistics} \citep{dawid1984prequential}, which shares our focus on predicting one unseen variable but is limited to forecasting the next step in a time series.\looseness=-1}

Why?  First, it is simple to extract a point prediction of a variable $X_i$ from its conditional marginal.  The \defn{Bayes-optimal} prediction is the one with minimum expected loss (\defn{Bayes risk}) given the observations $\vx_\vs$ so far:\looseness=-1
\begin{equation}\label{eq:point-prediction}
\hat{x_i} = \argmin_{x_i} \sum_{x_i^* \in \Dom_i} P^*(X_i=x_i^* \mid \vX_\vs=\vx_\vs) \cdot \taskloss(x_i \mid x_i^*)
\end{equation}
Under $L_2$ or $L_1$ loss, for example, $\hat{x_i}$ is the conditional mean or conditional median of $X_i$.\looseness=-1

Marginal distributions predict only one variable at a time.  Fortunately, once we can condition on arbitrary subsets of observations, we can \defn{chain} these conditional marginal predictions to obtain conditional joint predictions as needed.  The additional conditioning values in this chain may be hypothetical or sampled rather than observed.  For example, for any distinct $j,k \notin \vs$,
\begin{equation}\label{eq:chain}
P^*(X_j=x_j, X_k=x_k \mid \vX_\vs=\vx_\vs) = P^*(X_j=x_j \mid \vX_\vs=\vx_\vs)\cdot P^*(X_k=x_k \mid \vX_\vs=\vx_\vs, X_j=x_j)
\end{equation}
and we can drop in \textsc{Marformer} estimates of both factors. This generalizes autoregressive prediction, in a manner previously explored by orderless NADE \citep{uria2016neural} and XLNet \citep{yang2019xlnet}.  The generalization is important: because the \textsc{Marformer} specifies \emph{all} conditional distributions, it is fast to evaluate the two factors of \labelcref{eq:chain}, or to sample $x_j$ and then $x_k$ from them.  In contrast, under the autoregressive (``factored regression'') models widely used in the previous missing data literature \citep{lipsitz1996conditional,enders2025missing}, those operations would require expensive marginalization or importance reweighting, except when the fixed autoregressive order happens to include the specific factors that appear in \labelcref{eq:chain}.

Making a point prediction like \labelcref{eq:point-prediction} can be regarded as choosing an action. One can also use probabilities to evaluate \emph{real-world} actions with task-specific losses. The application defines the available actions and their losses; the \textsc{Marformer} supplies the conditional predictions used to evaluate them. After observing $\vX_\vs=\vx_\vs$, the distribution \labelcref{eq:chain} suffices to choose among actions whose losses depend on the state of the world only through the state's attributes $X_j$ and $X_k$ (and optionally $\vX_s$).
The Bayes risk of a possible action $a$ is again its expected loss, now under the conditional joint distribution \labelcref{eq:chain}:
\begin{equation}\label{eq:bayes-risk}
\mathrm{risk}(a \mid \vX_s=\vx_s) = \sum_{x_j,x_k} P^*(X_j=x_j,X_k=x_k \mid \vX_\vs=\vx_\vs) \cdot \taskloss(a \mid \vX_s=\vx_s, X_j=x_j, X_k=x_k)
\end{equation}
The Bayes-optimal action is $\hat{a} = \argmin_a \mathrm{risk}(a \mid \vX_\vs=\vx_\vs)$.  For example, a doctor may wish to assess each candidate treatment $a$ after observing symptoms $\vx_\vs$.  \Cref{eq:bayes-risk} does so by marginalizing over an unseen disease $X_j$ and an unseen patient preference $X_k$, both of which are imputed from $\vx_\vs$ using the \textsc{Marformer}.  It may be convenient to approximate the expectation \labelcref{eq:bayes-risk} by Monte Carlo, sampling $X_j$ and $X_k$ as mentioned above.  The same samples can be reused to compute the Bayes risks of alternative candidate actions $a'$.

An important speedup is available here.  Since the loss of $a$ is a deterministic function of random variables such as $X_j$ and $X_k$, it is a random variable itself.  Suppose we call it $X_i$.
Then $\mathrm{risk}(a \mid \vX_s=\vx_s)$ is the conditional expectation
$\sum_{x_i} P^*(X_i=x_i \mid \vX_\vs=\vx_\vs) \cdot x_i$, as in \cref{eq:point-prediction}.  We can again drop in the \textsc{Marformer}'s estimate of the conditional marginal here---provided that we had the forethought to include $X_i$ in the \textsc{Marformer}'s training, and provided that the \textsc{Marformer} can directly predict $X_i$ as accurately as it estimates it via \cref{eq:chain,eq:bayes-risk}.  Given that investment in training, this approach is much faster at prediction time.  It no longer requires an expectation over many chained $(x_j,x_k)$ pairs.  A \emph{single} call to the \textsc{Marformer} is now enough to predict the Bayes risk of $a$ and also, in parallel, other candidate actions $a'$.

Is it worth gathering more information before choosing an action?  Conditional marginals enable \defn{active feature acquisition}~\citep{doi:10.1287/mnsc.1080.0952}, which repeatedly chooses a next attribute to observe until further acquisition would no longer improve expected outcomes enough to justify its cost.
The \defn{value of information} (VOI) of observing $X_j$ in the preceding example is the expected reduction in loss, considering
how this new information changes the chosen action from $\hat{a}$ to the possibly different $\hat{a}_{x_j}$ \citep{howard1965voi,howard1966voi}:
\begin{align}
\voi(X_j \mid \vX_\vs=\vx_\vs) &= \sum_{x_j} P^*(x_j \mid \vX_\vs=\vx_\vs) \cdot (\risk(\hat{a}\mid \vX_\vs, X_j=x_j) - \risk(\hat{a}_{x_j} \mid \vX_\vs, X_j=x_j)) \\
& \text{where }\hat{a}=\argmin_a \risk(a \mid \vX_\vs)\text{ and } \hat{a}_{x_j} =\argmin_a \risk(a \mid \vX_\vs, X_j=x_j)
\end{align}
This can be computed using exactly the same conditional marginals as in \cref{eq:bayes-risk}.
$X_j$ is worth observing if its VOI exceeds the corresponding \defn{cost of information} (COI), that is, the expected cost of observing $X_j$.  (This cost is often a constant known in advance, but in general it is just another random variable for the \textsc{Marformer} to predict.)  Heuristically, one may choose a next variable to observe by maximizing $\frac{\mathrm{VOI}-\mathrm{COI}}{\mathrm{COI}}$ \citep{doi:10.1287/mnsc.1080.0952}, which represents the immediate \defn{rate of return on investment}  (ROI).\footnote{This is a \localdefn{myopic} (i.e., greedy) heuristic in the sense that it only considers the benefit and cost of the single next acquisition.  In a sequel paper, we will train a \textsc{Marformer}-based information acquisition policy by reinforcement learning, which can look farther into the future if desired.  Fully discounting the future turns out to retrieve the myopic ROI heuristic, or more precisely, an improved version that replaces the ratio-of-expectations with an expected ratio.}

In short, conditional marginals serve to guide decision-making about \emph{which action to take} and \emph{which information to gather first}.

In this paper, we focus on making conditional marginal estimation straightforward, fast, and accurate, using a generic model rather than a bespoke model for the domain.  For this purpose we will assume fixed training and test distributions.  In a sequel paper, we will attempt to train a policy for incrementally acquiring conditioning variables that usefully inform some downstream decision, along with training all conditional marginal distributions that the policy consults on the data that it gathers \citep{li-et-al-afa}.

\subsection{Predicting From Conditional Marginals on Incomplete Data}\label{sec:missingdefs}\label{sec:prediction}

Often we must work with incompletely observed data---for example, whatever active feature acquisition has collected so far.  To model incomplete datasets with stochastically missing attributes \citep{rubin1976,little2019}, the probability space from \cref{sec:notation} is also equipped with a set-valued random variable $\vR$ that specifies which attributes are \defn{observed} (``revealed'').
The other attributes are \defn{missing}.
Thus, an \defn{instance} $(\vx,\vr) \sim P^*(\vX,\vR)$ has $\vr \subseteq \vi$ and gives rise to the \defn{observed instance} $\vx_\vr$ that appears in training or test data.  We refer to $\vx$ as the \defn{complete instance}.  $P^*(\vX,\vR)$ may be factored into the \defn{complete-data distribution} $P^*(\vX)$ and the \defn{missingness mechanism} $P^*(\vR\mid\vX)$.

Incomplete data methods can have subtle bugs.  The rest of this section ensures correct prediction on incomplete \emph{test} instances.  The next section (\cref{sec:training}) ensures asymptotically correct estimation from incomplete \emph{training} instances.  Some readers may prefer to skip this material on a first reading and proceed to \cref{sec:experimental-setup}.

Example: Let $\vi=\{1,2,3,4\}$.  If nature ($P^*$) provides $\vx$ and then chooses to reveal attributes
$\vr={\{2,4\}}$,
then we observe the instance $\vx_\vr
= \{2 \mapsto x_2, 4\mapsto x_4\}$.
We then expect the missing value $x_1$ to be distributed according to the conditional marginal $P^*(X_1 \mid \vX_\vR = \vx_\vr) = P^*(X_1 \mid \vR=\{2,4\}, X_2=x_2, X_4=x_4)$.

The prediction in this test example takes care to condition on $\vR=\vr$, since the observation pattern $\vr$ is itself observed and is potentially useful---the fact that $X_1$ and $X_3$ were stochastically missing may provide an additional clue about their values. In contrast, the subtly different notation $P^*(X_1 \mid \vX_\vr = \vx_\vr)$ refers to the simpler distribution $P^*(X_1 \mid X_2=x_2, X_4=x_4)$, which is derived from only $P^*(\vX)$, with no dependence on the missingness mechanism $P^*(\vR\mid\vX)$.  This notation does not mention $\vR$: it treats $\vr$ as a given constant, rather than as a realization of the random variable $\vR$.

In this paper, we train \textsc{Marformer} models $\ptheta$ (as well as classical models) of this simpler distribution.  Both distributions marginalize over the unmentioned variables ($X_3$ in the example).  They can be related through Bayes' Theorem:
\begin{equation}\label{eq:bayes}
\underbrace{P^*(X_i=x_i \mid \vX_\vR=\vx_\vr)}_{\text{predictive distribution}} = P^*(X_i=x_i \mid \vR=\vr, \vX_\vr=\vx_\vr) \propto \underbrace{P^*(X_i=x_i \mid \vX_\vr=\vx_\vr)}_{\text{modeled distribution}} \cdot \underbrace{P^*(\vR=\vr \mid X_i=x_i, \vX_\vr=\vx_\vr)}_{\text{missingness likelihood}}
\end{equation}

\textbf{MNAR:} In the general \defn{missing not at random} (MNAR) case, predicting missing values requires not only our trained model but also knowledge of the missingness likelihood factor in \labelcref{eq:bayes}.  That factor cannot be learned from the observed instances alone \citep{molenberghs2008every}.

\textbf{IMAR:} However, the missingness likelihood factor can clearly be dropped if it is known to be constant with respect to $x_i$.  Let us say that the missingness mechanism $P^*(\vR\mid\vX)$ is \defn{individually missing at random} (IMAR) if the probability of the specific observation pattern $\vR=\vr$ is independent of the individual missing value $x_i$, given the observed values:
$P^*(\vR=\vr \mid X_i=x_i, \vX_\vr=\vx_\vr) = P^*(\vR=\vr \mid \vX_\vr=\vx_\vr)$, for all $\vr \subseteq \vi$, $\vx_\vr \in \prod_{j\in\vr} \Dom_j$, and $i \in \vi\setminus\vr$.
On the resulting incomplete instances $\vx_\vr$, the predictive distribution in \cref{eq:bayes} is always identical to the modeled distribution.  So an IMAR mechanism is \defn{ignorable} for our task $P^*(X_i=x_i \mid \vX_\vR=\vx_\vr)$ of predicting an \emph{individual} variable: the mechanism's factor can always be dropped from \cref{eq:bayes} without knowing anything else about it.  This is a new definition.

\textbf{MAR:} The traditional \defn{missing at random} (MAR) property \citep{rubin1976,seaman2013missing} strengthens IMAR to guarantee ignorability for the more difficult task $P^*(\vX_{\vi\setminus\vr} \mid \vX_\vR=\vx_\vr)$ of predicting \emph{all} missing variables jointly.  Simply replace $X_i$ with $\vX_{\vi\setminus\vr}$ throughout \labelcref{eq:bayes}.  The ``modeled distribution'' factor would now be defined by chaining as in \cref{eq:chain}.  The missingness likelihood factor becomes $P^*(\vR=\vr \mid \vX_{\vi\setminus\vr}=\vx_{\vi\setminus\vr}, \vX_\vr=\vx_\vr)$, and MAR states that it is always constant with respect to all of $\vx_{\vi\setminus\vr}$, so that it can again be dropped.  That is, given the observed values, the probability of the observation pattern  $\vR=\vr$ does not depend on any function of the missing values $\vx_{\vi\setminus\vr}$ (e.g., whether $x_1=x_2$).  More concisely, MAR states that $P^*(\vR=\vr \mid \vX=\vx) = P^*(\vR=\vr \mid \vX_\vr=\vx_\vr)$.\footnote{For an example of a mechanism that is not MAR but is still IMAR: Suppose $P^*(X_1,X_2\mid X_3)$ assigns equal probability to $(0,0),(0,1),(1,0)$, and $(1,1)$, and both $X_1$ and $X_2$ are missing if $X_1=X_2$, but otherwise neither is missing.  Observing that both variables are missing carries information about the \emph{pair} $(X_1,X_2)$, but not about the \emph{individual} variables $X_1$ and $X_2$.}

Conveniently for our future applications, a dataset constructed by active feature acquisition is always MAR.\footnote{Interestingly, the converse is false: it is possible to construct a MAR distribution that does not correspond to any acquisition policy, but this is not obvious.  We found an example using Gurobi.}  The acquisition policy has no access to currently missing attributes, so the probability of observing attributes $\vr$ and then stopping (that is, $P^*(\vR=\vr \mid \vX=\vx)$) cannot depend on any still missing attributes.

\textbf{MCAR:} Stronger still is the \defn{missing completely at random} property, $P^*(\vR=\vr\mid\vX=
\vx) = P^*(\vR=\vr)$, which states that the missingness pattern does not depend on $\vx$ at all (not even on its observed part $\vx_\vr$).
Note that MCAR does not require i.i.d. missingness---it allows variables to be missing at different rates, and allows their missingness to be correlated, as long as the missingness pattern is selected without looking at $\vx$.

\subsection{Training Conditional Marginals on Incomplete Data}
\label{sec:training}

We now discuss possible training strategies.
In general, a conditional model $\ptheta(C \mid B)$ can be trained by
\defn{conditional maximum likelihood estimation} (CMLE):\footnote{Notice that the sum in \labelcref{eq:cmle} is a finite-sample estimate of \labelcref{eq:avgkl}, up to an affine transform.}
\begin{equation}\label{eq:cmle}
\hat{\vtheta} = \argmax_\vtheta \sum_{b,c} \log \ptheta(c \mid b)
\end{equation}
where the sum is over i.i.d.\@ draws from $P(B,C) \defeq P_1(B)\cdot P_2(C \mid B)$, known as \defn{training examples}.
This is essentially a way of fitting $\ptheta$ to $P_2$.  Even when $\ptheta$ is misspecified, meaning that no $\vtheta$ yields the true $P_2$ exactly, CMLE will still asymptotically minimize the following weighted average of divergences:\footnote{%
This follows from regarding \cref{eq:cmle} as maximum-likelihood estimation (MLE) of the \localdefn{compound model} $p_1(B)\cdot\ptheta(C\mid B)$, for any fixed distribution $p_1$ with
at least the support of $P_1$.
Thus, CMLE inherits all the usual guarantees of MLE.  In particular, under regularity conditions, MLE still converges at rate $O_p(1/\sqrt{n})$ to the minimum of $\KL{P}{p_1\cdot\ptheta}$---even in the ``quasi-MLE'' setting where the compound model is misspecified \citep{white-1982}.  It follows that CMLE converges to the minimum of \labelcref{eq:avgkl}.
}\looseness=-1
\begin{equation}\label{eq:avgkl}
\E[b\sim {P_1}]{\KL{P_2(\cdot \mid b)}{\ptheta(\cdot \mid b)}}
\end{equation}
In other words, CMLE asymptotically fits $\ptheta(C\mid b)$ to $P_2(C\mid b)$ for all $b$ such that $P_1(b) > 0$.  When no $\vtheta$ can do this exactly, CMLE favors fitting the distributions whose $P_1(b)$ is larger.  It may also converge faster to those distributions, since they are better represented in the training data. For both reasons, one may wish to choose $P_1$ to favor conditions $b$ that are similar to the expected test-time queries $\ptheta(C\mid b)$.

\paragraph{CMLE for conditional marginals.} Recall from \cref{sec:why-marginals} that our goal is to fit a conditional model $\ptheta(c\mid b)$ to match $P^*(X_i=x_i \mid \vX_\vs=\vx_\vs)$.
Since CMLE will match the conditional distributions $P_2$ in the training examples, this means that the distribution of training examples needs to satisfy
$P_2(x_i \mid (\vx_\vs,i)) = P^*(X_i=x_i \mid \vX_\vs=\vx_\vs)$.  In this setup, each training or test example $(b,c)$ will have the form $((\vx_\vs, i), x_i)$.  The input $b$ is an observation of $\vX_\vs$ paired with a target attribute $i\not\in\vs$, and the target output $c$ is the true value of $X_i$.

\paragraph{Training instances vs.\@ training examples.} Many training \emph{examples} of the above form---single predictions---can be derived from a single observed training \emph{instance} $\vx_\vr$, by masking out some of the observed attributes $\vr$ and training the \textsc{Marformer} to predict them individually.  Simply choose $\vs \subseteq \vr$ and derive $|\vr|-|\vs|$ examples corresponding to $i \in \vr\setminus\vs$.  All of these examples have the same \textsc{Marformer} input $\vx_\vs \defeq (\vx_\vr)_\vs$, so conveniently, their terms in \cref{eq:cmle} can all be computed at once in a single forward pass.

\paragraph{Observability bias.} Unfortunately, in the resulting distribution of training examples, $P_2$ may not match the conditionals of $P^*(\vX)$ as required.  The problem may arise even when the data are MAR and the masking mechanism is MCAR!  Example: Let $\vi=\{1,2\}$. Say $X_1 \in \{0,1\}$ is always observed, but $X_2 \in \{0,1\}$ is much more likely to be observed when $X_1=1$ than when $X_1=0$. Like all dynamically acquired datasets, this dataset will be MAR.  In particular, given $X_1=x_1$, whether $X_2$ is then observed ($2\in\vr$) is independent of $X_2$'s value.  This lets us safely learn $P^*(X_2\mid X_1)$ by masking the observed $X_2$ values (with some fixed probability) and learning to predict them.  However, we cannot safely learn $P^*(X_1\mid X_2)$ in the same way.  Suppose the complete-data distribution assigns $(0,0),(0,1),(1,0),(1,1)$ probabilities of $0.4,0.1,0.1,0.4$ respectively.
Then $P^*(X_1=0 \mid X_2=0)=0.8$, whereas unfortunately, few training examples with $X_2=0$ have $X_1=0$.  That is because such training examples are derived from instances in which $X_2$ was observed at all, which tend to have $X_1=1$ due to the missingness mechanism.\footnote{These are the training examples with $\vr=\{1,2\}, \vs=\{2\}, i=1$. On these examples, $P_2(C=0 \mid B=(\{2 \mapsto 0\},1)) \ll 0.8$.}

\textbf{MNAR+MNAR}: In the general case of MNAR training instances and an MNAR masking mechanism, we could correct for observability bias through \localdefn{inverse propensity weighting}.  For each condition $(\vx_\vs,i)$, we would adjust the conditional distribution in masked training examples, $P^*(X_i \mid \vX_\vS = \vx_\vs)$, to match the distribution we wish to model, $P^*(X_i \mid \vX_\vs = \vx_\vs)$.  The two distributions are related just as in \cref{eq:bayes}, so weighting each masked training example by $1/P^*(\vS=\vs \mid X_i=x_i, \vX_\vs=\vx_\vs)$ corrects $P_2$ (while distorting $P_1$).  Alas, we generally do not know that weighting factor, which depends in a complex way on both the missingness and masking mechanisms and on the complete-data distribution.

But suppose that weighting factor does not depend on $x_i$.  In other words, suppose the combined missingness+masking mechanism is IMAR, yielding IMAR masked instances $\vX_\vS$.  Then the mechanism is ignorable, just as we saw beneath \cref{eq:bayes}: the two distributions already match.
In other words, we can avoid observability bias simply by obtaining IMAR masked instances.  How can we accomplish that?

\textbf{MAR+IMAR}: Given MAR training instances, we saw above that even MCAR masking suffers from observability bias.  However, we could \localdefn{impute} missing variables like $X_2$ to obtain complete instances.  We can then mask the complete instances with any IMAR mechanism, which does not know what was originally missing (i.e., masking now chooses $\vs\subseteq\vi$ and not $\vs\subseteq\vr$).  This produces IMAR instances as needed.  As we will show in a sequel paper, the imputation can be done by the \textsc{Marformer} itself, on an as-needed basis, with no further knowledge of the missingness mechanism.  This is morally the same approach as Monte Carlo EM \citep{wei-tanner-1990}, but it trains a collection of conditional distributions rather than a single generative model.

\textbf{MCAR+IMAR}: As the present paper focuses on proving out the \textsc{Marformer} architecture, we can be less concerned with its source of training examples.  We accordingly simplify by requiring the training instances to be MCAR.  In this special case, neither weighting nor imputation is necessary.  Any IMAR masking mechanism for choosing $\vs\subseteq\vr$ will produce IMAR masked instances, which yield training examples with the correct conditional distributions $P_2$.\looseness=-1

To formalize this point: A \defn{masking mechanism} is a conditional distribution $P^*(\vS \mid \vX_\vR=\vx_\vr)$ that is used to generate masked data from incomplete data.  It is MAR if $P^*(\vS=\vs \mid \vX_\vR=\vx_\vr) = P^*(\vS=\vs \mid \vR=\vr, \vX_\vs=\vx_\vs)$.  More weakly, it is IMAR if $P^*(\vS=\vs \mid \vR=\vr, \vX_\vs=\vx_\vs, X_i=x_i) = P^*(\vS=\vs \mid \vR=\vr, \vX_\vs=\vx_\vs)$ (for all $\vr \subseteq \vi, \vs \subseteq \vr, \vx_\vs, i \in \vr\setminus\vs, x_i$), meaning that the masking pattern does not depend on any \emph{individual} masked value $x_i$.\footnote{This is slightly stronger (closer to MAR) than the IMAR condition for the missingness mechanism (\cref{sec:missingdefs}).  We still marginalize over other masked attributes $X_j$ as before.  However, we do not marginalize over the observation pattern $\vR$, but rather require the independence to hold separately for each $\vr$, even though $\vr$ will also be masked (i.e., it is not seen as part of $\vx_\vs$).}  The combined mechanism---MCAR missingness followed by IMAR masking---produces IMAR data as desired:  $P^*(\vS=\vs \mid \vX_\vs=\vx_\vs, X_i=x_i) = P^*(\vS=\vs \mid \vX_\vs=\vx_\vs)$.\footnote{Proof: multiply the left- and right-hand sides of the defining IMAR equality by $P^*(\vR=\vr \mid \vX_\vs=\vx_\vs, X_i=x_i)$ and $P^*(\vR=\vr \mid \vX_\vs=\vx_\vs)$ respectively.  Then sum over $\vr$ to obtain the desired result.  The multiplication step preserves equality because both multipliers equal $P^*(\vR=\vr)$ due to the MCAR missingness mechanism.  To see this for the second multiplier, $P^*(\vR=\vr \mid \vX_\vs=\vx_\vs) = \sum_\vx P^*(\vR=\vr,\vX=\vx \mid \vX_\vs=\vx_\vs) = \sum_\vx P^*(\vR=\vr \mid \vX=\vx) \cdot P^*(\vX=\vx \mid \vX_\vs=\vx_\vs) = \sum_\vx P^*(\vR=\vr) \cdot P^*(\vX=\vx \mid \vX_\vs=\vx_\vs) = P^*(\vR=\vr) \sum_\vx P^*(\vX=\vx \mid \vX_\vs=\vx_\vs) = P^*(\vR=\vr)$.  The same argument applies to the first multiplier.}

In short, provided that the original incomplete training instances are MCAR, we can simply use IMAR masking to produce new IMAR instances, for which the supervision provided by the masked values is unbiased.

In terms of the discussion after \cref{eq:avgkl}, we have now satisfied the core requirement to match the data distribution's conditionals $P_2$ to the desired conditionals $P_2^*$, obtaining \emph{conditionally} unbiased data.
We do not \emph{need} to also match $P_1$ to the distribution $P_1^*$ of test inputs---but doing so would give better results when $\ptheta$ has insufficient capacity
to fit $P_2$ perfectly.  Both $P_1$ and $P_1^*$ are over prediction queries $b=(\vx_\vs,i)$.  We again leave this problem to future work on deriving training examples.\footnote{\label{fn:matchinputs}Several techniques seem promising.
Given a dataset of training inputs $\sim P_1$ and a sample of test inputs $\sim P_1^*$ (all of the form $b=(\vx_\vs,i)$ in our setting), \localdefn{covariate shift} methods show how to reweight the training examples to focus training on test-like inputs \citep{bickel2009covariate}, effectively replacing $P_1$ with $P_1^*$ in the weighted average \labelcref{eq:avgkl}.
\localdefn{Gradient alignment} goes further: it attempts to dynamically upweight the parts of the $P_1$ that benefit training \citep{ren2018reweight,shu2019meta}.  For example, training on $b$ may improve performance on held-out training examples with test-like inputs, even if $b$ is not itself a test-like input (i.e., $P_1^*(b)$ is low) but merely has some similar properties.  Although reweighting methods reduce the effective sample size, we are free to tune $P_1$ so as to mitigate this (that is, tune the IMAR masking function to generate examples that need less reweighting).
We can further improve effective sample size by sampling training examples from an (arbitrarily biased) proposal distribution $Q(b,c)$ instead of $P_1(b)P_2(c\mid b)$, and correcting for this by multiplying the weight of example $(b,c)$ by an importance weight $P_1(b)P_2(c\mid b)/Q(b,c)$, provided that $Q$ covers the support of the target distribution.  \localdefn{Adaptive importance sampling} \citep{bugallo2017ais} trains $Q$ so as to maintain effective sample size: the highly weighted examples will now be proposed more often so that they do not have to be weighted as highly.  In our setting, $Q$ must select a training instance and also mask it; we can freely learn any selection distribution and any masking mechanism (perhaps sharing parameters with the \textsc{Marformer}), as long as we can compute the importance weight efficiently.} In the experiments
we describe below,
$P_1$ is already reasonably close to $P_1^*$.

\paragraph{Remark.} The presentation in \crefrange{sec:prediction}{sec:training} has assumed that both training and test instances are drawn from the same $P^*$.  Using the same complete-data distribution $P^*(\vX)$ is what ensures that the trained model $\ptheta$ can be applied to test data, as in most machine learning.  However, the missingness mechanisms $P^*(\vR\mid\vX)$ do not really have to be the same---for example, training might include more complete instances (\cref{sec:data-real}). The missingness mechanism on a training instance informs the masked-data training strategy (\cref{sec:training}), while the missingness mechanism on a test instance plays the different role of informing prediction (\cref{sec:prediction,eq:bayes}).\footnote{\label{fn:source}A more general presentation allows several \localdefn{data sources}, where data from a given source $x_0$ is distributed as $P^*(\vX,\vR\mid x_0)$.  It is this conditional distribution that may be MAR, MCAR, etc.  The special source variable $X_0$ is observed on every instance and thus can inform the masking or prediction strategy on that instance.  Training data may be drawn from one mixture of sources, and test data from a different mixture.  Notice that if a test source has probability 0 in training, then the generalization of the trained \textsc{Marformer} $\ptheta$ to this novel $x_0$ value depends entirely on the inductive bias of the \textsc{Marformer}.  Omitting $x_0$ from the \textsc{Marformer} input imposes a strong inductive bias, forcing it to use the same conditional model for all sources.}

\subsection{Experimental Setup}\label{sec:experimental-setup}

Our experimental setup for this paper is diagrammed in \cref{fig:train-eval}.
The training examples and test examples that could be derived from a given instance $\vx_\vr$ are distinct.  At training time, we would mask some of $\vx_\vr$'s observed attributes and train to predict them (predicting $x_i$ from $\vx_\vs$ for all $i \in \vr\setminus\vs$).  At test time, however, we would predict the actually missing attributes (predicting $x_i$ from $\vx_\vr$ for all $i \in \vi\setminus\vr$).

Actually \emph{evaluating} these test-time predictions requires knowing the complete instance $\vx$.   Thus, the test set $\Test$ consists of $(\vx,\vr)$ pairs drawn i.i.d.\@ from $P^*(\vX,\vR)$.
We evaluate with log-loss, a \defn{proper scoring rule} that rewards $\ptheta$ for matching the true conditional distributions in test data \citep{gneiting2007strictly}.
Our specific evaluation loss
in this paper is
\begin{equation}\label{eq:eval}
\sum_{(\vx,\vr)\in\Test} \sum_{i\in\vi\setminus\vr} -\log\ptheta(x_i\mid (\vx_\vr,i))
\end{equation}
Note that \cref{eq:eval} evaluates $\ptheta$ only on its marginal predictions (not on chained joint predictions as in \cref{eq:chain}).  It assumes that the test instances are IMAR, so that it can ignore the missingness mechanism in \cref{eq:bayes} and simply make the predictions directly from $\ptheta$.

In effect, \cref{eq:eval} converts the test instance $(\vx,\vr)$ into $|\vi|-|\vr|$ test examples $(b,c)=((\vx_\vr, i), x_i)$ that will be individually evaluated by log-loss.
It also weights all test examples equally.  (An applied setting should evaluate instead on the test predictions that are useful for the application, changing the weighting $P_1^*$.)  Test instances with smaller $|\vr|$ will generate more test examples and thus contribute more terms to \cref{eq:eval}.
We report the result in units of nats per example,
dividing the total loss \labelcref{eq:eval} by the total number of examples $\sum_{(\vx,\vr)\in\Test} (|\vi|-|\vr|)$.

In settings where we know the true distribution $P^*(\vX)$ that governs $X_i$, we can ignore $x_i$ and more directly evaluate the marginal predictions by measuring their divergence from the true marginals, replacing \labelcref{eq:eval} with
\begin{equation}\label{eq:evalkl}
\sum_{(\vx,\vr)\in\Test} \sum_{i\in\vi\setminus\vr} \underbrace{\KL{P^*(X_i={\cdot} \mid \vX_\vr=\vx_\vr)}{\ptheta(\cdot \mid \vx_\vr)}}_{\text{expectation of }(-\log \ptheta) - (-\log P^*) \geq 0}
\end{equation}

We train the model $\ptheta$ using a training set $\Train$ of incomplete MCAR \defn{training instances} $\vx_\vr$ together with an IMAR masking mechanism $P^*(\vS \mid \vx_\vr)$.  Following the discussion in \cref{sec:training}, we may locally minimize the training objective
\begin{equation}\label{eq:trainsimple}
\E[\vx_\vr\in\Train]{\E[\vs \sim P^*(\vS\mid\vx_\vr)]{\sum_{i\in\vr\setminus\vs} -\log \ptheta(x_i\mid\vx_\vs)}}
\end{equation}
which stochastically converts a training instance $\vx_\vr$ into $|\vr-\vs|$ equally weighted training examples $(b,c)=((\vx_\vs, i), x_i)$ for CMLE.  We use a stochastic gradient method so that the instances $\vx_\vr$ along with their masking patterns $\vs\subseteq\vr$ can be sampled randomly during training, according to the expectations in \cref{eq:trainsimple}.  \Cref{eq:trainreg} below adds regularization to this objective, and \cref{fn:matchinputs} discussed possibilities for reweighting its various summands.

In our present experiments, we use a very simple masking mechanism: each $i\in\vr$ has a 15\% independent probability of being masked.  As in BERT \citep{devlin-etal-2019-bert}, the 15\% number is intended to be small enough that the masked training inputs $\vx_\vs$ are not distributed too differently from the unmasked test inputs $\vx_\vr$, yet large enough that we obtain several training examples each time we sample an instance $\vx_\vr\in\Train$.

We will compare to classical generative modeling as discussed in \cref{sec:intro}.  These methods have access to the same $\Train$ and $\Test$ sets and evaluate with the same \cref{eq:eval}.  However, they do not convert the training instances $\vx_\vr$ into conditional training examples as in \cref{eq:trainsimple}.  Instead they fit a joint model $\qphi$ to maximize the joint log-posterior, which---when $\Train$ is MAR---is
\begin{equation}\label{eq:map}
    \left(\sum_{\vx_\vr\in\Train} \log \qphi(\vx_\vr)\right) + \log \pprior(\vphi)
\end{equation}
plus an ignorable constant.  $\Train$ is MCAR and therefore MAR in our present experiments.

Some of our experiments make the test inputs available at training time, which is known as \defn{transductive} learning \citep{Vapnik98,ouchi-etal-2019-transductive}.
More precisely, $\Train$ includes all incomplete test instances $\{\vx_\vr: (\vx,\vr)\in\Test\}$.
These are only test \emph{inputs}; of course, the desired test \emph{outputs} $\vx_{\vi\setminus\vr}$ are not seen during training!  Transductive \textsc{Marformer} training will mask the test inputs to obtain additional training examples. With this paper's simple masking strategy, that works only if $\Test$ is MCAR (like the rest of $\Train$), not just MAR.

The reason we train transductively is to compare fairly to Bayesian MCMC (\cref{sec:joint-mg}), which is naturally transductive.  The MCMC sampler operates on training and test data together, sampling from the joint posterior over parameters and missing values; making it non-transductive would be difficult and expensive.\footnote{One would have to run MCMC without the test inputs to obtain samples of the generative model parameters, and then for \emph{each} sample, run MCMC again with the test inputs given and the parameters frozen to obtain samples of the test outputs.}

Many users of the \textsc{Marformer} might prefer to apply it in the usual \defn{non-transductive} manner: train $\ptheta$ with no advance knowledge of the test inputs, then run it on the stream of test inputs.  Skipping the training on the test inputs ignores potentially useful information, but is much faster at test time than any transductive method.  We evaluate this case as well and show that it has comparable accuracy at large training sizes.

\section{A Neural Architecture for Conditional Marginal Prediction}
\label{sec:methods}

\subsection{\mbox{\textsc{Marformer}} Architecture}\label{sec:architecture}\label{sec:types}

We now present our neural model $\ptheta(X_i \mid \vX_\vs = \vx_\vs)$.  Given the observation $\vx_\vs$ (where $\vs$ may be any subset of $\vi$), the \textsc{Marformer} predicts a conditional marginal distribution over $X_i$, for all attributes $i\in\vi$ in parallel.

\begin{figure}[h!]
  \centering
      \begin{subfigure}[t]{0.55\linewidth}
    \centering
    \includegraphics[width=\linewidth]{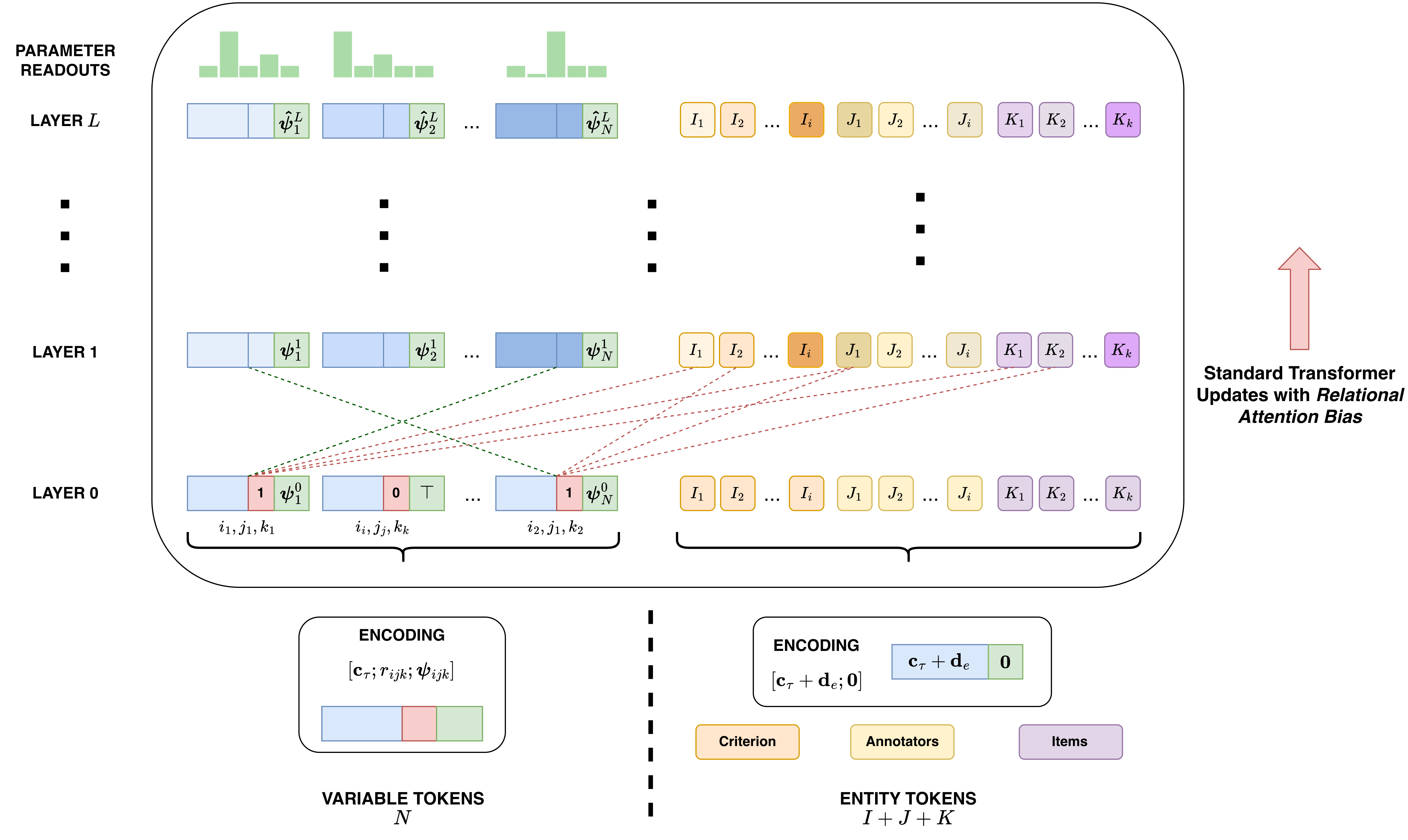}
    \caption{Full architecture. Variable tokens (left, $N$ per instance) carry attribute observations; entity tokens (right) carry relational identity. Parameter distributions $\vpsihat$ are read from variable tokens at layer $L$. Not all inter-layer connections are shown.}
    \label{fig:marformer-arch}
  \end{subfigure}
  \hfill
      \begin{subfigure}[t]{0.42\linewidth}
    \centering
                \includegraphics[width=\linewidth]{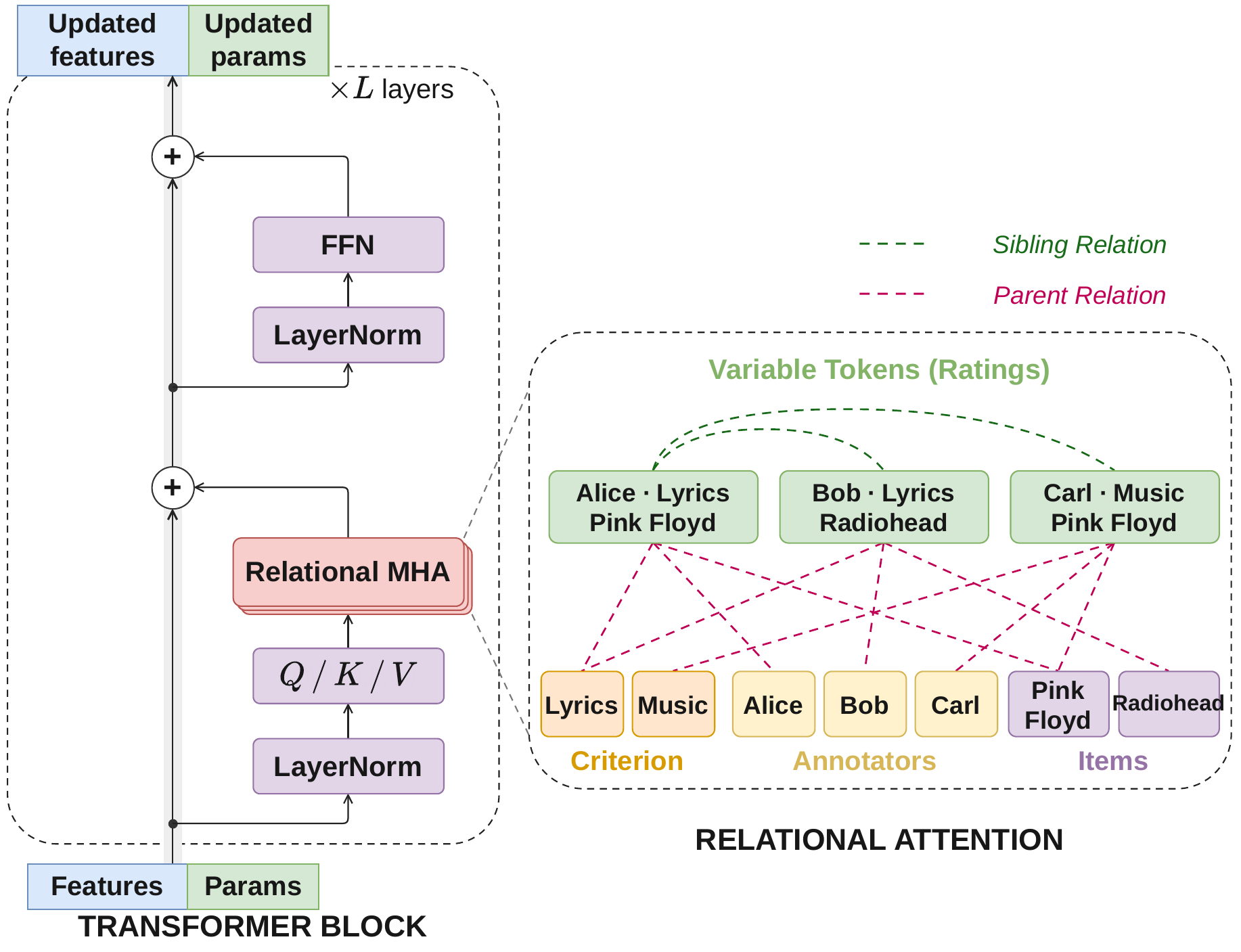}
    \caption{Transformer block internals (left) and relational attention graph (right), illustrated for the annotation domain. Dashed red = parent edges; dashed green = sibling edges.}
    \label{fig:marformer-block}
  \end{subfigure}
  \caption{\textsc{Marformer} architecture. The annotation domain (items, judges, criteria) is used as the running example, but the architecture supports any typed entity structure.}
  \label{fig:marformer}
\end{figure}

\begin{enumerate}

\item The first step of the forward pass is to encode the observed input $\vx_\vs$ as a collection of $|\vi|$ vectors.  This is analogous to how a text Transformer encodes a length-$|\vi|$ input sentence \citep{vaswani2017attentionneed} that may have missing tokens \citep{devlin-etal-2019-bert} into $|\vi|$ vectors.  For each $i\in\vs$, $x_i$ denotes the observed value of $X_i$.  For each $i\in\vi\setminus\vs$, let $x_i=\wildcard$ to indicate that the value of $X_i$ is missing.  In either case, the \textsc{Marformer} will encode the input fact $X_i=x_i$ into a vector $\vh_i\psup{0} \in \mathbb{R}^d$.

\item For each attribute $i\in\vi$, the \textsc{Marformer} now constructs a \defn{residual stream} of vectors $\vh_i\psup{1}, \dots, \vh_i\psup{L} \in \mathbb{R}^d$.  Each $\vh_i\psup{\ell}$ in this stream is computed from the entire preceding \defn{layer}, meaning the collection of representations $\vh\psup{\ell-1} = \{\vh_j\psup{\ell-1}: j\in\vi\}$.  The overall architecture is illustrated in \cref{fig:marformer-arch}. Again we follow the classical Transformer architecture (\cref{fig:marformer-block}): $\vh_i\psup{\ell-1}$ attends to all vectors in $\vh\psup{\ell-1}$ to compute a \emph{residual update}, which is added to $\vh_i\psup{\ell-1}$ to yield the \localdefn{transformed} representation $\vh_i\psup{\ell}$.  The exact formulas are given in \cref{eq:attention,eq:feedforward}.

\item For each $i\in\vi$, the top-layer representation $\vh_i\psup{L}$ now contains the parameters for the predicted conditional marginal $\ptheta(X_i \mid \vX_\vs = \vx_\vs)$. If $i\in\vs$, this distribution should simply place high probability on the original observation $X_i=x_i$.  Otherwise it should accurately predict the \emph{missing} value.
\end{enumerate}

Hyperparameters of the \textsc{Marformer} include the number of dimensions $d$ and the number of layers $L$, as well as others described below.  The training objective and optimization algorithm also have hyperparameters (\cref{sec:marformer-training}), as does the masking mechanism (\cref{sec:experimental-setup}).

\paragraph{Parametric marginal distributions.} To create a \textsc{Marformer}, one must also specify the \defn{type} $\tau_i$ of each random variable $X_i$.  A type can be reused across many attributes $i$.  A type $\tau$ specifies
\begin{itemize}[nosep]
\item a domain $\Dom$, which is shared by all random variables of type $\tau$;
\item a parametric family $p_\vpsi$ of probability distributions over $\Dom$, parameterized by $\vpsi \in \mathbb{R}^{d'}$ for some $d' \leq d$;
\item a layer-0 encoding function $e: (\Dom\cup\{\wildcard\}) \rightarrow \mathbb{R}^{d'}$, such that when $\vpsi=e(x)$, the distribution $p_\vpsi$ places most of its mass on the value $x$ if $x\in\Dom$, or is essentially uniform if $x=\wildcard$;
\item optionally, a loss function that can be used to extract point predictions via \cref{eq:point-prediction}.
\end{itemize}

The type $\tau$ has a learnable embedding $\ve_\tau\in\mathbb{R}^{d-d'-1}$, which is included in the \textsc{Marformer}'s parameters $\vtheta$.\footnote{If the \textsc{Marformer} uses only a single type $\tau$ as in Domains 1--2 below, we fix its embedding to $\vec{0}$ without loss of generality.}  To distinguish among multiple attributes of type $\tau$, each such attribute $i$ also has a learnable embedding $\ve_i \in \mathbb{R}^{d-d'-1}$.

Fix $i$ and suppose $\tau_i = \tau$, the type that was just described.  (For example, $d'$ denotes its parameter dimensionality.)
Then the layer-0 encoding of the input fact $X_i=x_i$
is constructed as the concatenation
\begin{equation}\label{eq:layerzero}
\vh_i\psup{0}=[b_i; (\ve_{\tau}+\ve_i)\;;\;\vpsi_i\psup{0}] \in \mathbb{R}^d \text{ \ where \ } b_i = [x_i=\wildcard] \text{ and } \vpsi_i\psup{0} = e(x_i) \in \mathbb{R}^{d'}
\end{equation}
The bit $b_i$ indicates whether $X_i$ is missing, and the subvector $\vpsi_i\psup{0}$ represents its value $x_i \in \Dom \cup \{\wildcard\}$. Which variable is being bound to this value?  That question is answered by the subvector $\ve_{\tau}+\ve_i$, which serves as an embedding of $i$, precisely analogous to a learned positional embedding when encoding text \citep{gehring2017convolutional,vaswani2017attentionneed}.  Notice that the position dimensions are kept separate from the value dimensions, because the value dimensions are meant to be directly interpretable as parameters without any unembedding function (this seems to work better in practice).

In general, for \emph{each} layer $\ell \in [0,L]$, again let $\vpsi_i\psup{\ell} \in \mathbb{R}^{d'}$ represent the final $d'$ elements of $\vh_i\psup{\ell}$.   We regard this as specifying parameters for $p_\vpsi$ to yield a distribution $p_{\vpsi_i\psup{\ell}}$ over the value of $X_i$.
At the final layer $L$, the \textsc{Marformer} returns the distribution $p_{\vpsi_i\psup{L}}$ over $\Dom_i$ as its output marginal $\ptheta(X_i\mid\vX_\vs=\vx_\vs)$, for every $i\in\vi$.

For an observed attribute $i\in\vs$, we expect
that the residual stream will preserve the
observed value, so $\vpsi_i\psup{L} \approx
\vpsi_i\psup{0}$.  For a missing attribute
$i\notin\vs$, we expect that as $\ell$ increases,
the distribution will gradually sharpen from the
uniform distribution at the input layer $\ell=0$
to a useful predictive distribution at the output
layer $\ell=L$ (as \cref{sec:logit-lens} bears out
empirically).  The training objective
\labelcref{eq:trainreg} will encourage both of
these behaviors.

\paragraph{Constructing the residual stream.}
At each layer $\ell \in [1,L]$, each attribute has a single residual stream $\vh_i\psup{\ell} \in \mathbb{R}^d$. The token representations are updated by a standard pre-norm Transformer block \citep{vaswani2017attentionneed}, consisting of multi-head self-attention followed by a ReLU feedforward module:
\begin{align}
 \vec{h'}_i\psup{\ell} &= \vec{h}_i\psup{\ell-1} + \mathrm{Attention}\!\left(\mathrm{LayerNorm}(\vec{h}\psup{\ell-1})\right)_{\!i} \label{eq:attention} \\
 \vec{h}_i\psup{\ell} &= \vec{h'}_i\psup{\ell} + \mathrm{FeedForward}\!\left(\mathrm{LayerNorm}(\vec{h'}_i\psup{\ell})\right) \label{eq:feedforward}
\end{align}
Layer normalization is applied to the full token representation, including its parameter subvector. Attention combines information across attributes, while the feedforward module acts separately on each attribute. Each sublayer adds its output to its input through the residual connection shown above. The resulting $\vh_i\psup{\ell}$ is passed directly to the next layer. The feature and parameter dimensions are subvectors of this single residual stream, and both can be updated by each sublayer. At the final layer, the predicted parameters $\vpsihat_i = \vpsi_i\psup{L}$ are read directly from the final $d'$ dimensions of $\vh_i\psup{L}$.

Each layer has its own attention, feedforward, and layer-normalization parameters. Hyperparameters include the number and dimensionality of attention heads, and the number and dimensionality of intermediate layers in the feedforward network. The attention operator uses the standard multi-head procedure, with the optional relational modification described next.

\paragraph{Relational attention.} Sometimes one needs to predict an attribute $i$ that was rarely or never observed.  For example, the annotation task below (\cref{sec:domain-annot}) has only a single instance with thousands of attributes $\vi$. Each variable $X_i$ is observed at most once in this single instance---and the missing variables that we need to predict at test time were observed zero times.  How do we generalize to $X_i$ from other variables that we predicted during masked training?  Clearly there was not enough training data to learn a useful attribute embedding $\ve_i$.

In the annotation task, $X_i \in \{1,2,\ldots,C\}$ represents a judge's rating of an item, so $i$ might be a structure like \texttt{\{item: doc53, judge: alice, criterion: humor\}} that describes the annotation request.
A reasonable first idea is to construct $\ve_i$ compositionally.  We could augment the \textsc{Marformer} with a neural network that encodes the structure $i$ into a vector $\ve_i$, so that related structures have related encodings.  However, that would require designing this encoder.  The approach might also fail.  We would like the model to learn that \texttt{alice} and \texttt{bob}'s ratings of the same document are correlated, but it may be difficult to determine whether $\ve_i$ and $\ve_j$ encode ratings of the \emph{same} document: the reason is that if \texttt{doc53} and \texttt{doc54} both appear only in test data, then the encoding network will not have had an opportunity to learn distinct parameters for them.

Instead, we provide a \defn{relational attention} mechanism.  When creating a \textsc{Marformer}, one may optionally specify relationships of the form $i \edge{\rho} j$ among the attributes, where $i,j\in\vi$ and $\rho \in \Rho$ denotes the type of relation.  Following \citet[\S3.1]{shaw2018self}, we \emph{modify the attention mechanism} so that $i$ can learn to attend specifically to the attributes $j$ such that $i \edge{\rho} j$.  Its residual stream can extract information from their residual streams.\footnote{\cite{shaw2018self} gave a general design for relational attention before applying it to relative positions in a sentence.}

For the annotation experiments, we use learned relational key and value embeddings, following
\citet[\S3.1]{shaw2018self} and extending its formulation to multiple relational types. Let $\mathcal{R}_{ij}$ be the set of relation types linking $i$ to $j$.
For attention head $h$, the score and output are
\begin{align}
 e_{ij}^{h} &= \frac{(\vec q_i^{h})^\top
 (\vec k_j^{h}+\sum_{\rho\in\mathcal{R}_{ij}}\vec a_{\rho}^{K,h})}
 {\sqrt{d_h}}, \\
 \alpha_{ij}^{h} &= \operatorname{softmax}_{j}(e_{ij}^{h}), \\
 \vec z_i^{h} &= \sum_j\alpha_{ij}^{h}
 (\vec v_j^{h}+\sum_{\rho\in\mathcal{R}_{ij}}\vec a_{\rho}^{V,h}).
\end{align}
Here $d_h$ is the head dimension, and the query, key and value vectors are learned linear
functions of the normalized residual streams. Each head uses its own slice of the relation
embeddings $\vec a_{\rho}^{V,h}$ and $\vec a_{\rho}^{K,h}$; multiple relations between two tokens contribute additively. The key term modifies
attention weights, while the value term supplies relation-specific information to the output.
These are learned embeddings rather than fixed relation indicators. The key modification alone
is not a claim of greater expressiveness than a linear relation readout; the value modification
also changes the information aggregated by attention.

Relational attention allows prediction by analogy.  To predict $X_i$ at test time---even if it was never predicted at training time---the \textsc{Marformer} will know to consult $X_j$ if the supervised training examples taught it to predict attributes $X_{i'}$ by consulting the variables $X_{j'}$ such that $i' \edge{\rho} j'$.  For example, if $\edge{\texttt{samedoc}}$ connects pairs of ratings on the same document, then the \textsc{Marformer} may have learned that such ratings are positively correlated.

\paragraph{Relational attention to and from entities.} Relational attention also allows the \textsc{Marformer} to build up representations in its residual stream, without the use of a separate encoder.  In the annotation example, one can encode the structural description of $i$ through relations like $i \edge{\texttt{judge}} \texttt{alice}$ and $i \edge{\texttt{item}} \texttt{doc53}$.  This requires the \defn{entities} such as \texttt{alice} and \texttt{doc53} to be added as new attributes with their own residual streams.  Technically, an entity is simply an attribute $j$ whose domain $\Dom_j$ consists of a single dummy value (which is considered to be observed and is never masked, i.e., $b_j=0$).
Predicting this value is trivial, so the entity's residual stream contains no parameters ($d'=0$).  Instead the residual stream serves solely to build up a useful representation of the entity.  For example, \texttt{doc53}'s representation might indicate that it is highly rated by children but adult judges have a wider range of opinions.
Constructing a good representation of \texttt{doc53} (at lower layers of the \textsc{Marformer}) lets attributes such as $i$ usefully attend to \texttt{doc53} (at higher layers) in order to predict how \texttt{alice} will rate it.
Constructing such a representation requires the relation graph to also include \localdefn{inverse relations} such as $\texttt{doc53} \edge{\texttt{item}^{-1}} i$, so that the residual stream of \texttt{doc53} can attend to all of its annotations.

Notice that the layer-0 embedding of $j=\texttt{doc53}$ is determined by the type embedding $\ve_{\tau_j}$ and the entity embedding $\ve_j$.  In this example, the type $\tau_j$ would presumably record that $j$ is a \emph{document entity} (as distinct from a \emph{judge entity} like \texttt{alice}).  If this particular document entity $j=\texttt{doc53}$ was frequently encountered in training data, then the entity embedding $\ve_j$ might learn idiosyncratic features of it.  If $j$ was rarely or never encountered, then $\ve_j \approx \vec{0}$, and its residual stream will be built ``from scratch'' by examining its related attributes (e.g., ratings).

\paragraph{Pruning.} When $\vi$ is very large or infinite, it is possible to reduce computation by omitting some of the residual streams.  In general, the streams to retain may depend on the input: the observation $\vx_\vs$ and the target attributes $i$ that one wishes to predict.  They could be chosen by a policy that has been hand-designed or trained to achieve a good speed-accuracy tradeoff \citep{stoyanov-eisner-2012-icmlw,vieira-eisner-2017-learning,NEURIPS2021_747d3443}.  The policy should retain residual streams for the target attributes; for all of the observed attributes $i\in\vs$ that might provide relevant information about them; and for any latent attributes $i\in\vi\setminus\vs$ that might also provide relevant information if reconstructed.

\subsection{\textsc{Marformer} Training}
\label{sec:marformer-training}

Training the parameters $\vtheta$ was sketched in \cref{sec:experimental-setup}.
In practice, we augment \labelcref{eq:trainsimple} with a copy regularizer and parameter
regularization. Writing the latter as an $L_2$ penalty gives the following objective, with
coefficients $\lambda_{\mathrm{copy}}\geq0$ and $\lambda_{L_2}\geq0$:

\begin{equation}\label{eq:trainreg}
\E[\vx_\vr\in\Train]{\E[\vs \sim P^*(\vS\mid\vx_\vr)]{\sum_{i\in\vr\setminus\vs} -\log \ptheta(x_i\mid\vx_\vs) + \overbrace{\lambda_{\mathrm{copy}} \sum_{i\in\vs} -\log \ptheta(x_i\mid\vx_\vs)}^\text{copy regularizer}}} +\overbrace{\lambda_{L_2} \,\norm{\vtheta}^2}^{\text{$L_2$ regularizer}}
\end{equation}

\paragraph{Regularization.} The optional \defn{$L_2$ regularizer} penalizes large parameter
norms. In the annotation experiments, we instead
use the decoupled weight decay of the standard AdamW optimizer.

The \defn{copy regularizer} says that the \textsc{Marformer} should correctly predict observed values, not just masked values.  BERT \citep{devlin-etal-2019-bert} includes this trivial task to ensure that an observed word token's encoding retains information about the word's identity. Similarly, our \cref{eq:trainreg} trains the residual stream for attribute $i$ to provide a conditional distribution over $X_i$ regardless of whether $X_i$ was \emph{observed} in the input $\vx_\vs$ or merely \emph{inferred} from it.  This homogeneous representation means that other residual streams have a standard location from which to obtain information about $X_i$.\footnote{Presumably, the \textsc{Marformer} can learn very early in training to satisfy the copy
regularizer, as this only requires retaining the observation $x_i$ (for $i\in\vs$) by preserving the layer-0 parameters throughout the residual stream: $\vpsi_i\psup{\ell} \approx \vpsi_i\psup{0} = e(x_i)$.    The residual streams of correlated variables will then learn to attend to $\vpsi_i\psup{\ell}$---initially to retrieve $x_i$ when $i\in\vs$, but then more generally to inspect layer $\ell$'s predictive distribution over $X_i$  regardless of whether $i\in\vs$.}

Potentially, one could also regularize through
\defn{deep supervision}
\citep{lee2015deeply,alrfou2019character}.  Recall
that $\ptheta$ in \cref{eq:trainreg} makes use of
marginal distributions $p_{\vpsi_i\psup{L}}$ from
the top layer $L$ of the \textsc{Marformer}.  Deep
supervision asks to also do well on this objective
when $p_{\vpsi_i\psup{\ell}}$ is used instead, for
lower layers $\ell$. \Cref{sec:logit-lens} shows that learned probes can recover useful predictions
from intermediate layers, although the raw parameter readout still benefits substantially
from the final layers. Whether deep supervision improves training remains an open question.

\paragraph{Optimization algorithm.} We optimize the conditional log-loss and copy term in
\cref{eq:trainreg} using AdamW \citep{loshchilov2019decoupledweightdecayregularization}.
In the annotation experiments, regularization uses
decoupled weight decay of $0.01$.

\paragraph{Validation and early stopping.}  As usual, early stopping on a validation set is one way to prevent overfitting to the training examples.  In our experiments, we found that this was only necessary in Domain 3 (\cref{sec:domain-annot}), and we describe there how we carve off the validation set for that domain (\cref{sec:marformer-annot}).

\paragraph{Dropout.} We use one more regularization trick, inspired roughly by \cite{srivastava2014dropout}.  Recall that each attribute $i$ has an embedding $\ve_i$.  If the number of distinct attributes grows unboundedly with the size of \Train, so does the number of parameters---making the \textsc{Marformer} nonparametric and at risk of overfitting.  Specifically, attributes that are common in training data can overcome the $L_2$ regularizer to learn useful idiosyncratic embeddings. Their very usefulness may sap the \textsc{Marformer}'s incentive to learn to handle \emph{uncommon} attributes that lack such embeddings, such as entities that may be important in test data but appear rarely or never in training data.  \cite{smith-cohn-osborne:2005:ACL} call this kind of problem \localdefn{weight undertraining}.

Thus, for each attribute type $\tau$, we define a \defn{dropout rate}
$\delta_\tau\in[0,1]$ for omitting the attribute-specific embedding $\ve_i$ during training.
On such an update, prediction relies on the type embedding and relational evidence rather than
the identity-specific deviation. In the annotation experiments, entity deviations are
initialized to $\vec 0$, while type embeddings are initialized from $\mathcal{N}(0,0.02^2)$.
An entity whose deviation has never been trained is therefore represented by its type and
relations. We use a fixed entity-dropout rate of $0.7$ on the generalized axis and $0$ on
the other axis. This heuristic is shared by the synthetic and real annotation experiments.
 This encourages the model to use
relational evidence even for entities encountered during training.

\section{Domain 1: Graphical Models}
\label{sec:domain-bn}

\begin{figure}[h!]
\centering
\includegraphics[width=\linewidth]{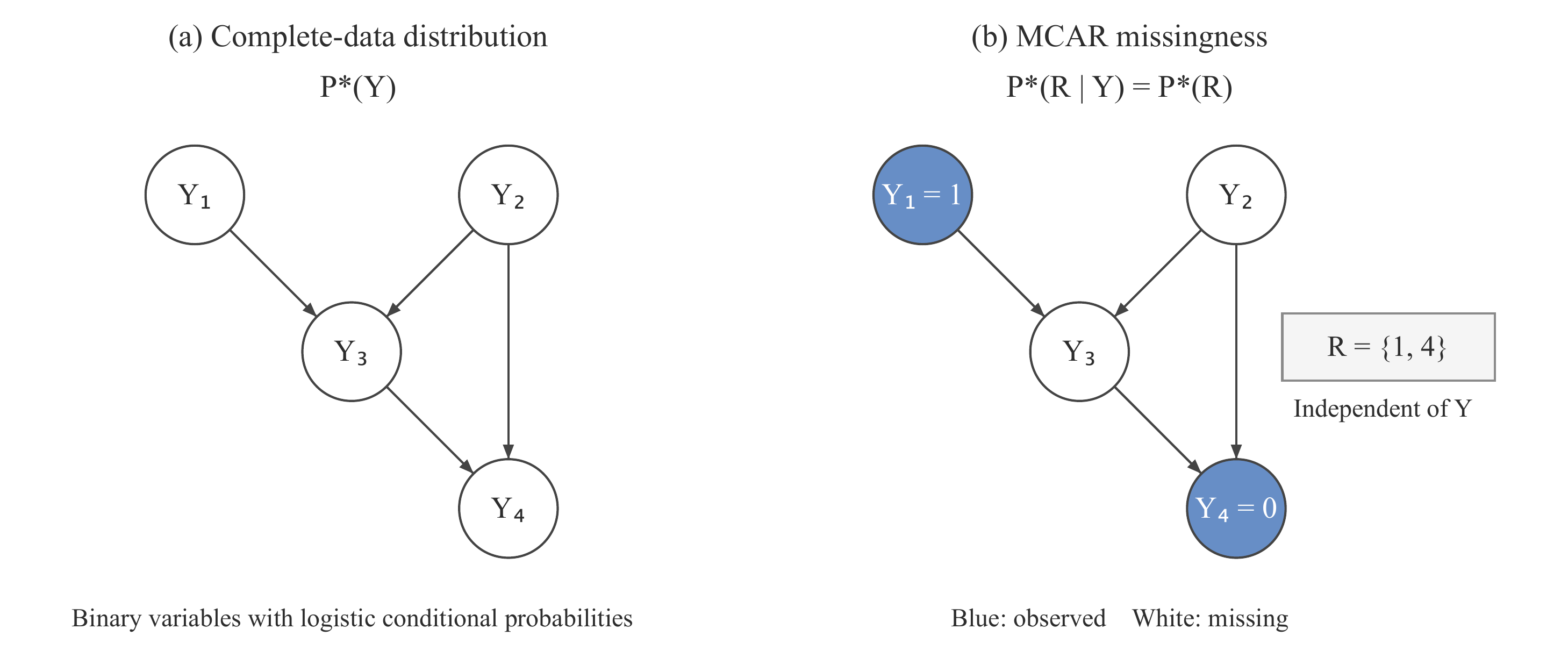}
\caption{Illustration of the complete-data distribution
and MCAR missingness in Domain 1, using a
four-variable example. (a) A Bayesian network over
binary variables, with logistic conditional
probabilities at each node. (b) An incomplete
instance under MCAR: the observed index set $\vR$
is independent of the complete data. Blue nodes
show observed values; white nodes are missing.
The variables labeled $Y_i$ in the figure correspond
to $X_i$ in the text. The experiments use randomly
generated networks with 10 variables.}
\label{fig:bn}
\end{figure}

To study the capabilities of the \textsc{Marformer}, we will evaluate its performance in three \emph{synthetic domains}---all MCAR, but with different qualitative properties.  In a synthetic domain, the data are generated from a known distribution $P^*(\vX,\vR)$.  This facilitates controlled evaluation:
\begin{itemize}[nosep]
\item We know the complete test-time instances $\vx$.  Hence we can evaluate test-time log-loss and measure calibration.
\item We know the true conditional marginals $P^*(X_i \mid \vX_\vR=\vx_\vr)$.  Hence we can measure their divergence from the conditional marginals predicted by $\ptheta$.
\item We know how $P^*$ was constructed.  Hence we can compare to classical methods for estimating $P^*$, in their favorable setting when they are estimating the parameters of the correct distribution family, with full knowledge of the prior from which those parameters were drawn to obtain $P^*$.
\end{itemize}

In this first section, $P^*$ is a \defn{Bayesian network} (BN).  The challenge here for the \textsc{Marformer}
is to predict missing variables without knowing
the conditional independence structure (\cref{fig:bn}).  In contrast, the
classical methods do know that structure.  The
full domain specification is in \cref{alg:bn}.

\subsection{True Generative Distribution}
\label{sec:true-bn}

The sample space has $10$ binary attributes $\vi=\{1,2,\dots,10\}$ where $(\forall i\in\vi)\,\Dom_i=\{0,1\}$.  The complete-data distribution $P^*(\vX)$ is given by a Bayesian network $\qphi(\vX) = \qphi(X_1,\dots,X_{10})$ with known parameters $\vphi=\vphi^*$.
The structure of the distribution is specified by a directed acyclic graph (DAG) $G$ with vertices $\vi$, and the conditional probability table (CPT) for each $X_i$ has the form of a logistic regression model:
\begin{align}\label{eq:bn}
\qphi(\vX=\vx)
&= \prod_{i\in\vi} \qphi(X_i=x_i \mid \vX_{\pa(i)} = \vx_{\pa(i)}) \\
&= \prod_{i\in\vi} \sigma((2x_i-1)\cdot\vw_i^T\,[1;\vx_{\pa(i)}])
\end{align}
where $\pa(i) \subsetneq \vi$ denotes the set of parents of node $i$ in $G$, and $\vw_i$ is the parameter vector (part of $\vphi$) for the conditional distribution of $X_i$ given the values of just those parents.

\defn{Selecting a particular distribution.} To generate the \emph{structure} of $\qphi$---the DAG $G$---we include the edge $j \to i$ (for $1 \leq j < i \leq 10$) with independent probability $\min(1,c/(i-1))$.  We take $c=5$.  Thus, each of $X_1,\dots,X_5$ has all previous variables as parents, and each of $X_6,\dots,X_{10}$ has an expected 5 of the previous variables as parents.

We then draw the \emph{parameters} $\phi^*$ i.i.d.\@ from $\mathcal{N}(0,1.5)$.  These logistic regression parameters $\vphi$ consist of one bias term per vertex plus one weight per edge, for a total of $\sum_{i\in\vi} (1+|\pa(i)|)$ parameters.

\paragraph{MCAR missingness mechanism.} Our missingness mechanism $P^*(\vR\mid\vX)=P^*(\vR)$ reveals each variable $X_i$ with i.i.d.\@ probability of $1/2$, in both training and test instances.  We use this mechanism to produce both training and test instances in all experiments.  Because it is MCAR, it is ignorable for all methods.

\paragraph{Inference under the true model.} In our experiments, we compare the models' predictions on test examples to the true conditional marginal distributions $\qphistar(X_i \mid \vx_\vr)$, which we compute with an exact variable elimination algorithm \citep{lazy-prop}.

\subsection{Datasets}\label{sec:data-bn}

We follow the experimental design in \cref{sec:experimental-setup}.  Given $P^*$:
\begin{itemize}
    \item We generate the $\Test$ dataset by
drawing $M=250$ test instances $(\vx,\vr)$ i.i.d.\@
from $P^*$.
    \item The corresponding $\Train$ dataset consists of $n$ incomplete training instances $\vx_\vr$ drawn i.i.d.\@ from the same $P^*$.  Our experiments in this domain are non-transductive, so $\Train$ does not include the $\Test$ instances.  We experiment with $\Train$ datasets of different sizes $n \in \{10, 250, 500, 750, 1000, 1250, 1500, 1750, 2000\}$.  Each $\Train$ dataset is a superset of the preceding one, for better comparability across sizes, but its model is trained independently.
\end{itemize}

We use these datasets to train and test the {\textsc{Marformer}} $\ptheta$ and also the full joint model $\qphi$.

\newcommand{\bnruns}{10\xspace}

We repeat our experiments \bnruns times, selecting a different distribution $(G,\vphi^*)$ each time.  This allows us to display 95\% bootstrap confidence intervals for all lines on our graphs.

\subsection{\textsc{Marformer} Configuration}\label{sec:marformer-bn}

In the terms of \cref{sec:architecture}, we create random variables $X_i$ for 10 attributes $i$, all with the same boolean type $\tau$, whose distribution family $p_\psi$ over $\Dom=\{0,1\}$ is parameterized by a single log-odds ratio (logit) $\psi\in\mathbb{R}$, so that $d'=1$ and $p_\psi(1) = \sigma(\psi)$.  The input encoding uses three-dimensional vectors: $e(\wildcard)=(1,0,0)$, $e(0)=(0,1,0)$, and $e(1)=(0,0,1)$, where the first coordinate indicates missingness. We use no relational edges among the 10 attributes, but each attribute $i$ has its own learnable embedding $\ve_i$.

We compare three \textsc{Marformer} variants to assess architectural scaling: Tiny ($L=2$ layers, $d=32$ dimensions), Small ($L=4, d=64$), and Large ($L=6, d=128$). The Tiny, Small, and Large variants use 2, 4, and 8 attention heads, respectively.

To stochastically create training examples from an instance in  $\Train$, we use a simple masking mechanism (\cref{sec:training}) that masks each observed variable with independent 0.15 probability.
We train each \textsc{Marformer} as described in \cref{sec:marformer-training} to obtain a parameter estimate $\hat{\vtheta}$.  We use AdamW with \texttt{weight\_decay=$0.01$}.  Since all test attributes will be very familiar from training data, we do not use dropout, setting $\delta_\tau=0$.

Each epoch is one shuffled pass over $\Train$ in minibatches of 32.
We clip gradients to a maximum norm of 1.0 and use
\texttt{ReduceLROnPlateau} to halve the learning rate after 22 epochs
without improvement in validation loss.

We run AdamW with learning rate $0.0001$ for up to 100 epochs, diagnosing early convergence when all of the 10 most recent epochs failed to decrease training loss by at least 0.1\%.

\begin{figure}[h!]
 \centering
 \begin{subfigure}{0.48\linewidth}
   \centering
   \includegraphics[width=\linewidth]{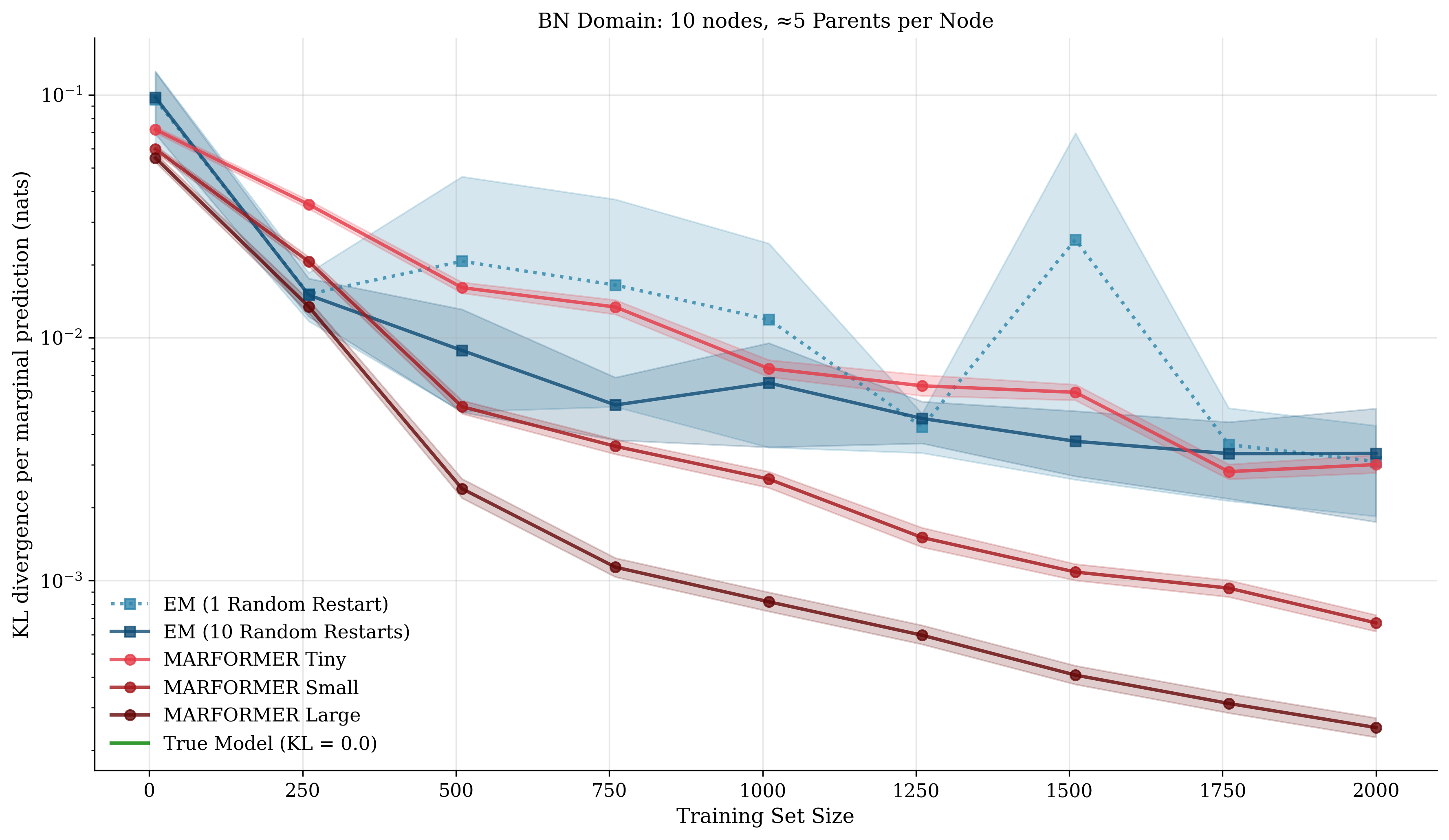}
   \caption{Domain 1 (BN)}
   \label{fig:kl-bn}
 \end{subfigure}
 \begin{subfigure}{0.48\linewidth}
   \centering
   \includegraphics[width=\linewidth]{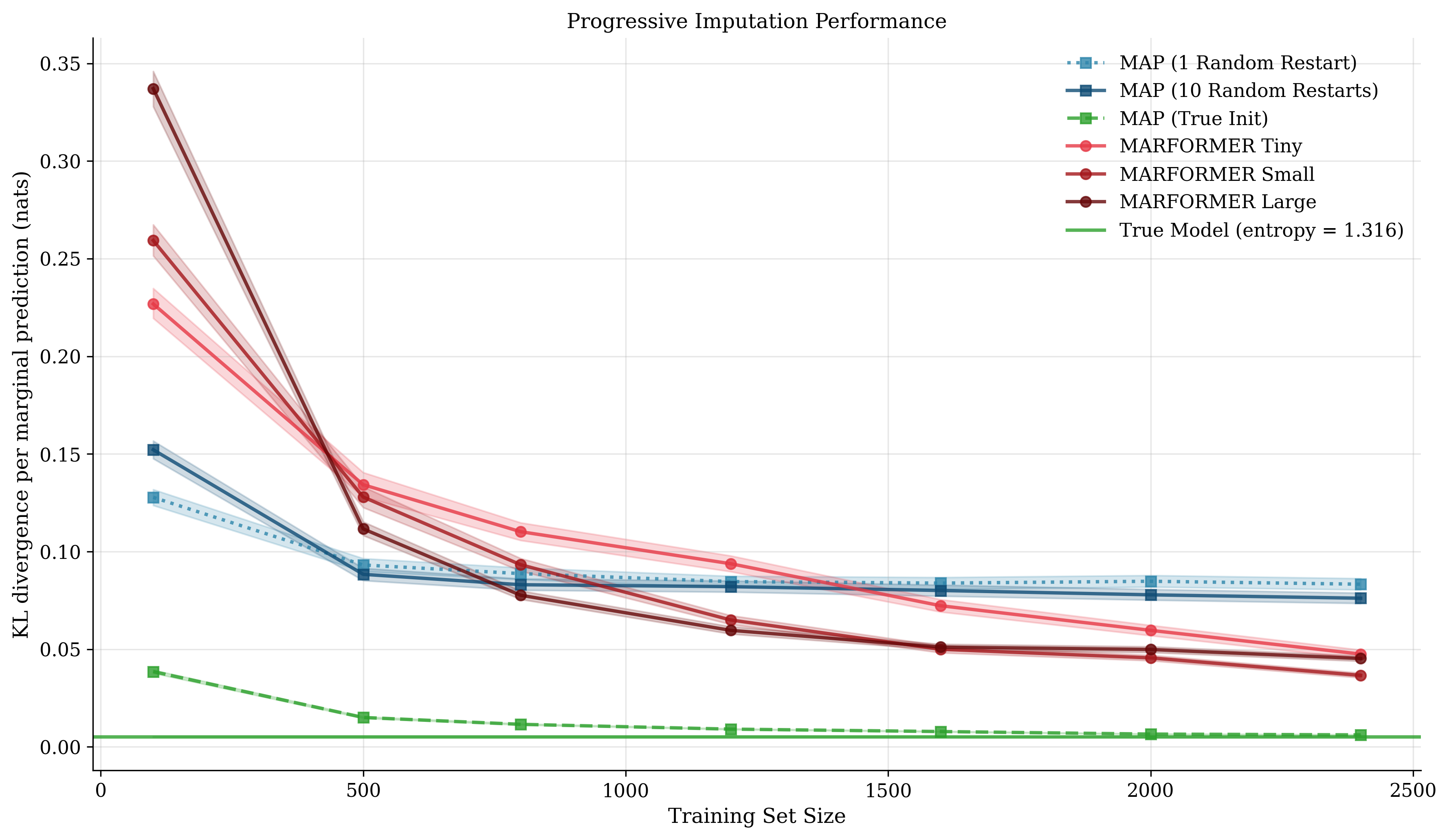}
   \caption{Domain 2 (Gaussian)}
   \label{fig:kl-mg}
  \end{subfigure}
 \caption{Learning curves for Domains 1 and 2, showing KL divergence of the true marginals from the predicted marginals (\cref{eq:evalkl}), averaged over \bnruns distributions $P^*$.  Shaded regions show 95\% bootstrap confidence intervals.  In Domain 1, the KL divergences are so small that they are displayed on a log scale.  For measuring KL, the true probabilities are computed exactly in Domain 1, but are estimated in Domain 2 using Monte Carlo integration.  That is why in Domain 2, the KL divergence between two copies of the true model $q_{\phi^*}$ is measured as slightly positive rather than zero (horizontal green line): the estimated probabilities are slightly different in the two copies.}\label{fig:kl}
\end{figure}

\begin{figure}[h!]
 \centering
 \begin{subfigure}{0.48\linewidth}
   \centering
   \includegraphics[width=\linewidth]{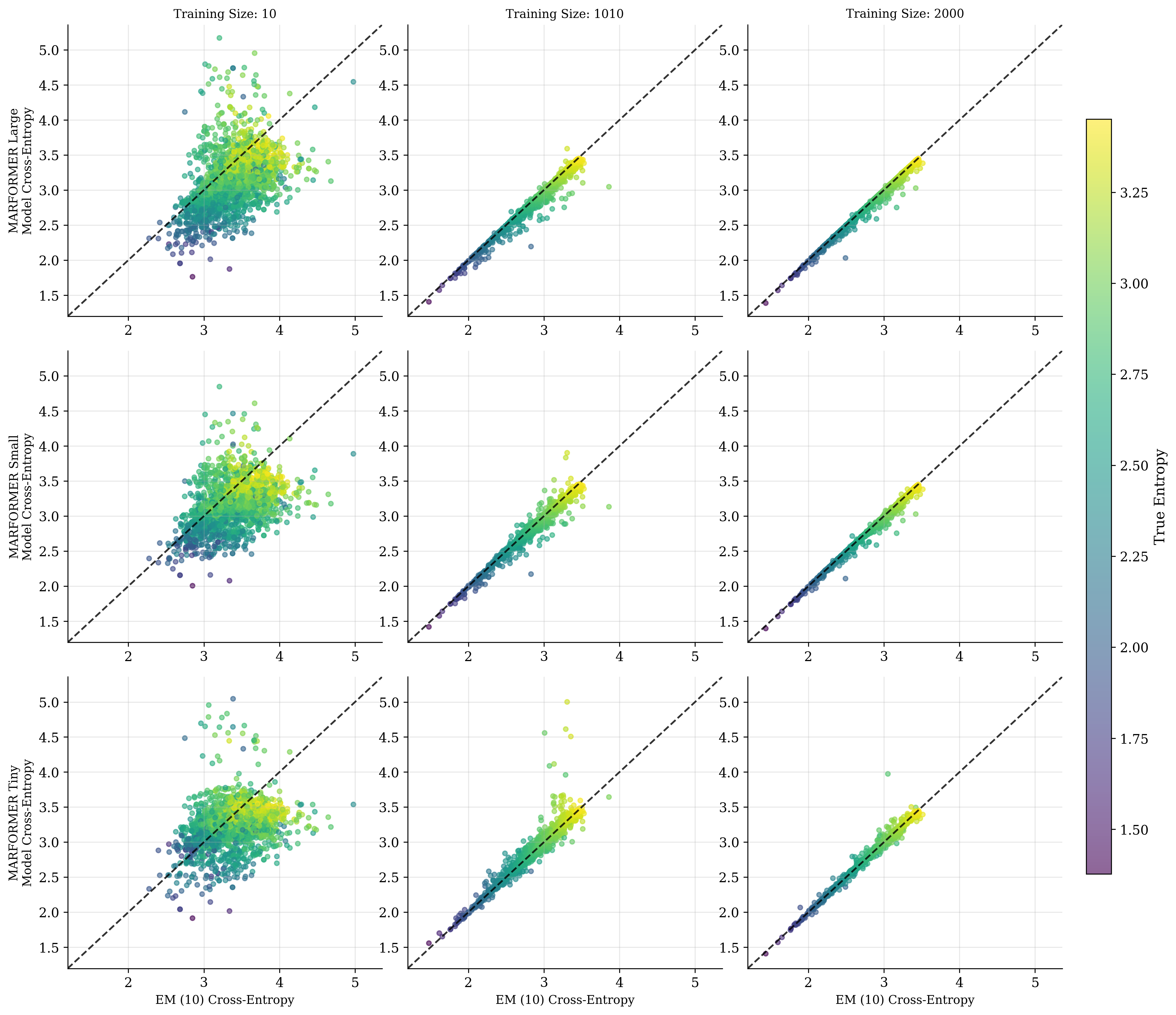}
   \caption{Domain 1}
   \label{fig:scatter-bn}
 \end{subfigure}
 \hfill
 \begin{subfigure}{0.48\linewidth}
   \centering
   \includegraphics[width=\linewidth]{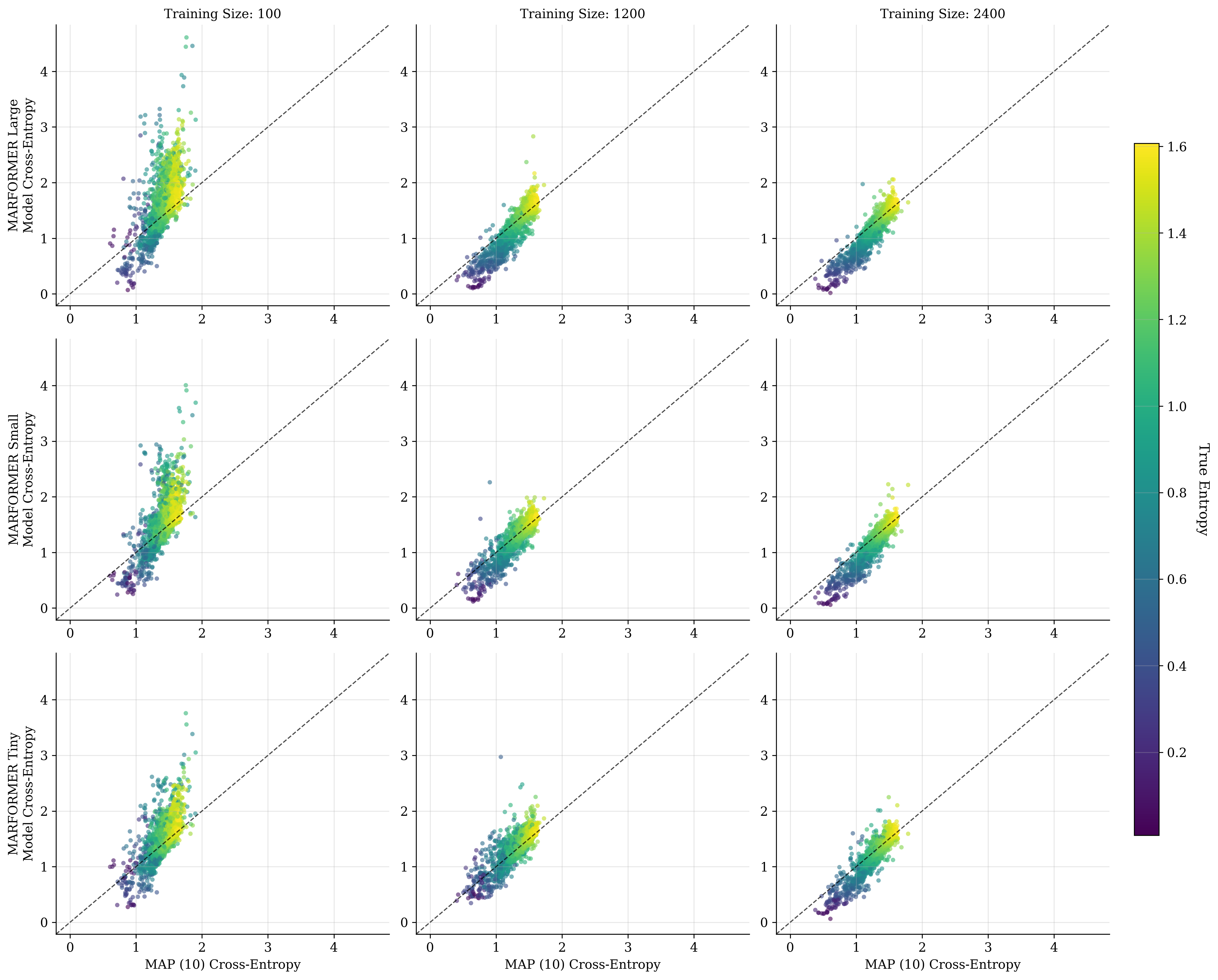}
   \caption{Domain 2}
   \label{fig:scatter-mg}
 \end{subfigure}
 \caption{Horizontal axis: Cross-entropy $\H{P^*,\qphihat}$ achieved by the estimated joint model (MAP-EM in Domain 1, \cref{sec:joint-bn}; direct MAP optimization in Domain 2, \cref{sec:joint-mg}).  Vertical axis: Cross-entropy $\H{P^*,\pthetahat}$ achieved by the estimated \textsc{Marformer}.  Different rows evaluate different \textsc{Marformer} sizes; different columns evaluate different $|\Train|$ sizes.  Each point corresponds to a single marginal prediction $P^*(X_i\mid\vx_\vr)$ on a test instance with $i\notin\vr\psup{m}$, color-coded by the entropy $\H{P^*}$ of the true marginal distribution (aleatoric uncertainty). Points below the diagonal are predictions on which \textsc{Marformer} is more accurate than the joint model.  Brighter points have higher entropy, due to less informative conditions $\vx_\vr$, so they tend to have higher cross-entropy under both models.  All graphs for the same domain use the same true generative distribution $P^*$.}
 \label{fig:scatter}
\end{figure}

\subsection{Joint Model Estimation}
\label{sec:joint-bn}

We compare to the classical approach of fitting a joint model of all the variables $X_i$.  Joint modeling is given the true model family $\qphi$ (which includes $P^* = \qphistar$) and only needs to estimate $\vphi$.  Specifically, joint modeling knows the true DAG $G$ and the logistic regression parameterization of its CPTs---as well as the true prior $\mathcal{N}(0,1.5)$ over $\vphi$.  The \textsc{Marformer} knows none of this.

As discussed in \cref{sec:intro}, we use the MAP-EM algorithm, which finds a parameter estimate $\hat{\vphi}$ that locally maximizes \cref{eq:map}.  We halt when the objective changes by less than 0.1\%, similar to the \textsc{Marformer}.  MAP-EM is sensitive to initialization because the objective is non-convex.  Thus, we allow 10 \defn{random restarts} and take the $\hat{\vphi}$ that achieves the best value of \labelcref{eq:map}.

We then make exact marginal predictions from $\qphihat$, and compare them to the true marginals from $\qphistar$.

\subsection{Experimental Results}\label{sec:results-bn}

Aggregate results are shown in \cref{fig:kl-bn}.  Larger {\textsc{Marformer}}s reliably achieve lower KL divergence, which continues to decrease as $|\Train|$ increases.  Traditional joint modeling does worse than all but the Tiny \textsc{Marformer}, despite its knowledge of the true model family and prior. Given that we use exact inference, the reason is presumably that MAP-EM suffers from the local maximum problem, as can be seen from the fact that 10 random restarts improves its average KL divergence to the truth.  Even with 10 random restarts, it presumably does not find the actual MAP estimate of $\vphi$.

\Cref{fig:scatter-bn} breaks down these KL divergences at three specific training sizes.
We saw in \cref{fig:kl-bn} that with only 10 training instances, neither approach is clearly better, but \cref{fig:scatter-bn} (column 1) further shows that they make different errors---the two methods can predict quite different probability distributions---and that there is a minority of outlier cases where the \textsc{Marformer} predictions are quite wrong.  By 1010 training instances (column 2),
when both methods achieve low KL (\cref{fig:kl-bn}), the Large \textsc{Marformer} is more accurate than the joint model at most predictions.  By 2000 training instances (column 3), the Small \textsc{Marformer} is as well.

\paragraph{Variant.} Recall that the naive masking method of this paper is valid only when the training instances are MCAR (\cref{sec:training}).\footnote{The problem is only with training (and specifically the naive method), not with testing.  A properly trained \textsc{Marformer} can directly predict missing values on test instances that are MAR (\cref{sec:prediction}).}
Out of curiosity, we also try the naive method in a setting that is MAR but not MCAR, even though the masked training examples will be biased, and find qualitatively similar results (\cref{sec:exp-mar}).

\subsection{Runtime}\label{sec:runtime-bn}

\begin{figure}[h!]
 \centering
 \begin{subfigure}{0.48\linewidth}
   \centering
   \includegraphics[width=\linewidth]{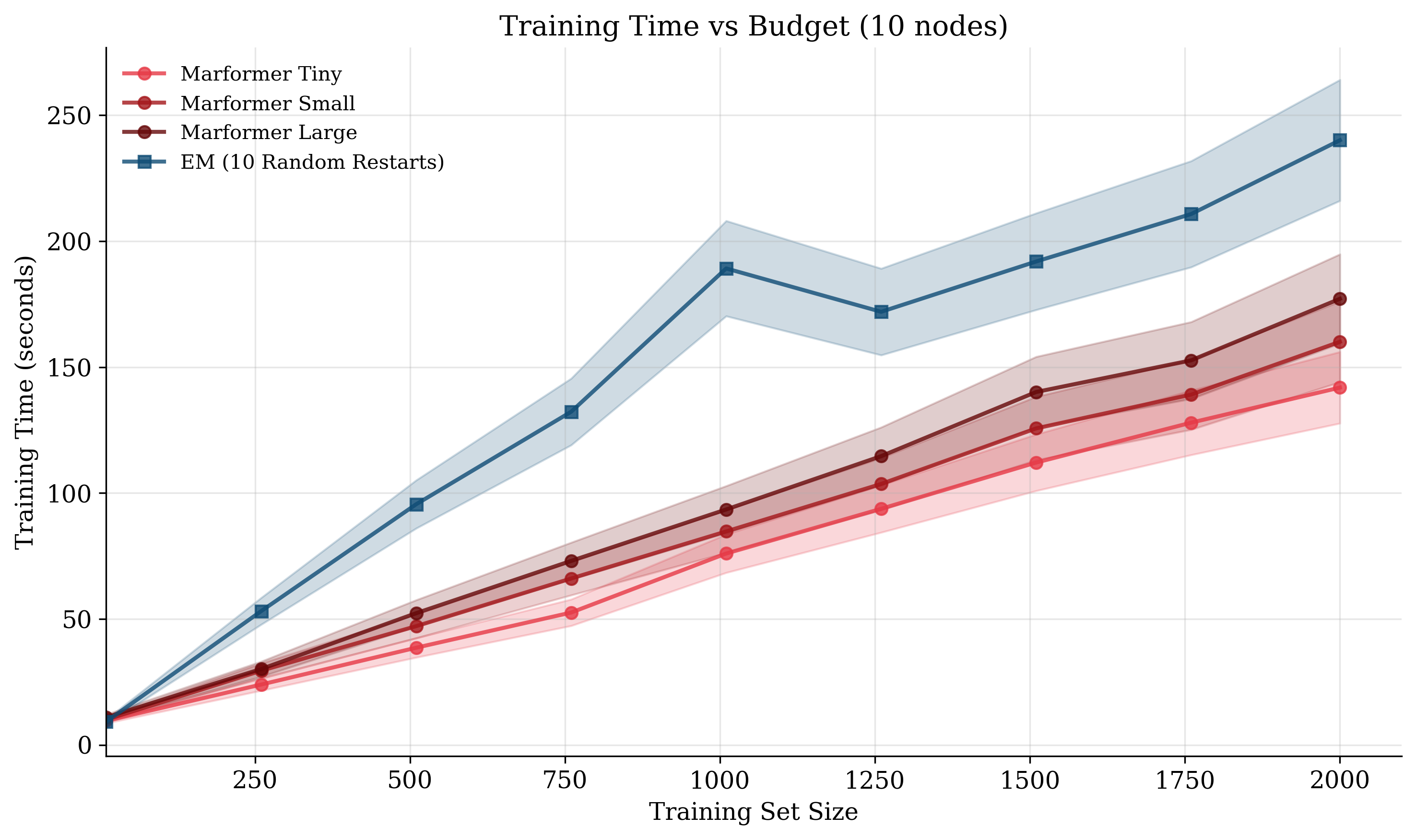}
   \caption{Training Time: Domain 1}
   \label{fig:runtime-bn}
 \end{subfigure}
 \hfill
 \begin{subfigure}{0.48\linewidth}
   \centering
   \includegraphics[width=\linewidth]{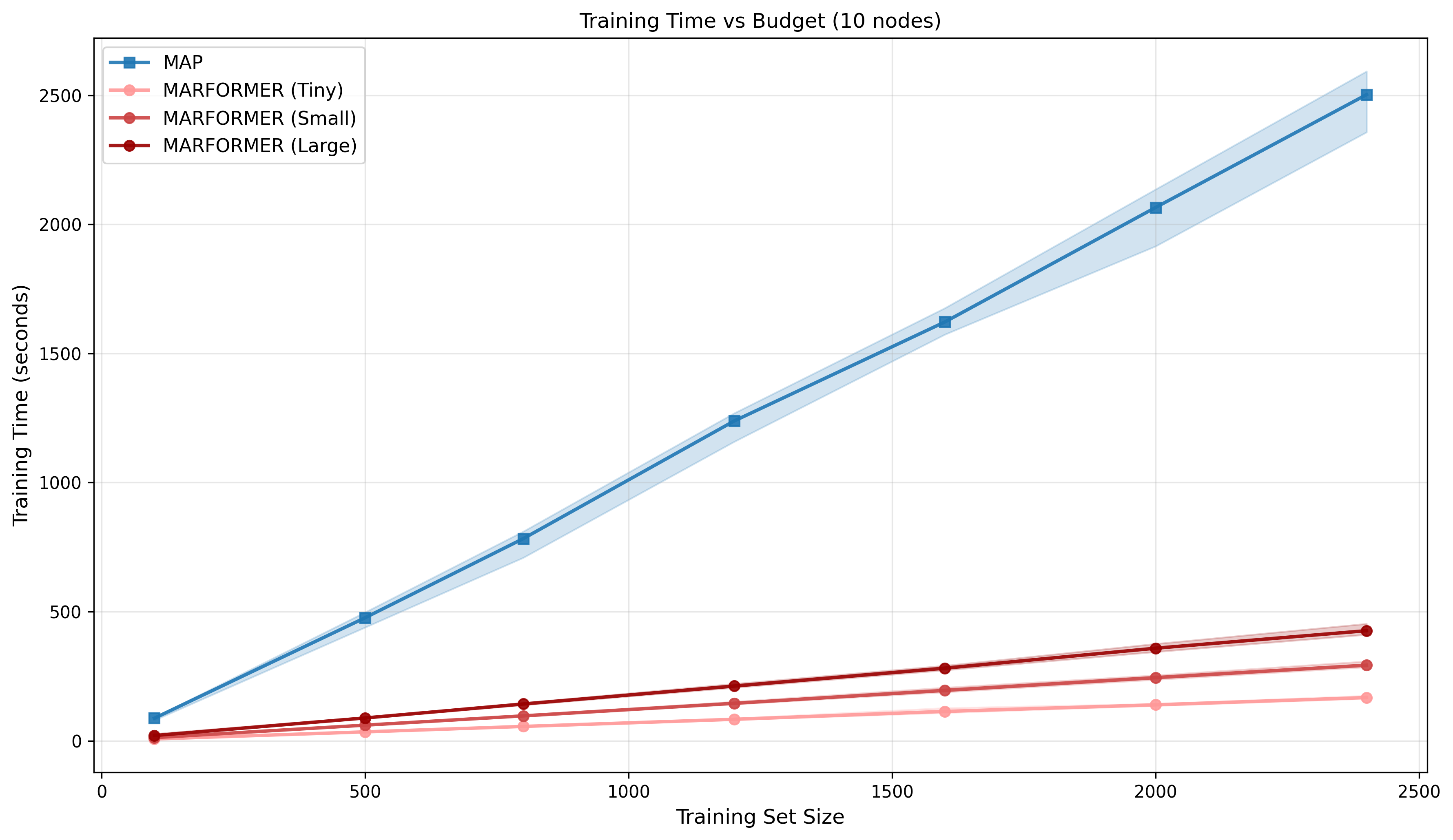}
   \caption{Training Time: Domain 2}
   \label{fig:runtime-mg}
 \end{subfigure}
 \caption{Training time (wall-clock seconds) on Domains 1 and 2 is essentially proportional to the training set size.  Even for these small problems, \textsc{Marformer} is substantially faster to train.  It is also faster at test time (not shown here), since inference requires only a single forward pass per test instance.} \label{fig:runtime}
\end{figure}

The classical approach of \cref{sec:joint-bn} is not always practical.  Extracting the marginal distributions from $\qphihat$, as well as performing the E step of MAP-EM, must marginalize over the missing variables $\vX_{\vi\setminus\vr}$ given $\vx_\vr$.
This task is computationally intractable except for low-treewidth BNs \citep{cooper1990computational,chandrasekaran2008complexity}.  We were able to run these experiments only because the BNs are quite small.

\Cref{fig:runtime-bn} shows that MAP-EM is slower than \textsc{Marformer} training even for these small Bayesian networks $G$, due to its need for random restarts.   Furthermore, the runtime of known MAP-EM algorithms will blow up exponentially as $G$ grows, since at the E-step they must solve the \#P-complete problem of marginal inference \citep{roth1996hardness}.  In practice, large BNs would be treated instead with approximate inference algorithms---introducing errors beyond those shown in \cref{fig:kl-bn}.\footnote{For sufficiently large training samples, the existing errors can be attributed to the fact that even exact MAP-EM is only an approximation to MAP---it finds a local optimum and not necessarily the true MAP estimate.}

The \textsc{Marformer} is a fast feed-forward computation
that can be \emph{trained} to approximate marginal
inference in BNs or other graphical models, as we
originally proposed in
\citet{pmlr-v15-stoyanov11a}.   Its runtime grows only
quadratically with $G$.  Previously,
\citet{stuhlmuller-et-al-2013} and
\citet{walecki2019universal} also proposed
training a single feed-forward network to predict
many conditional marginals.  However, those papers
were attempting to amortize inference in a known
model: they assumed that \emph{complete} training
instances could be sampled from $P^*$, whereas we
aim to train on naturally occurring incomplete
instances without access to $P^*$.  They also did
not use our Transformer architecture, which allows
cross-attention among unboundedly many random
variables (see
\cref{sec:domain-annot,sec:logit-lens}).

We caution that neither MAP-EM with approximate inference (using traditional polytime algorithms), nor the \textsc{Marformer}'s amortized approximate inference (using a polynomial-sized neural network), can come with any worst-case guarantees when $P^*$ is a BN, since marginal inference on general BNs is inapproximable \citep{roth1996hardness}.

\section{Domain 2: Discretized Multivariate Gaussians}
\label{sec:domain-mg}

\begin{figure}[h!]
\centering
\includegraphics[width=0.75\textwidth]{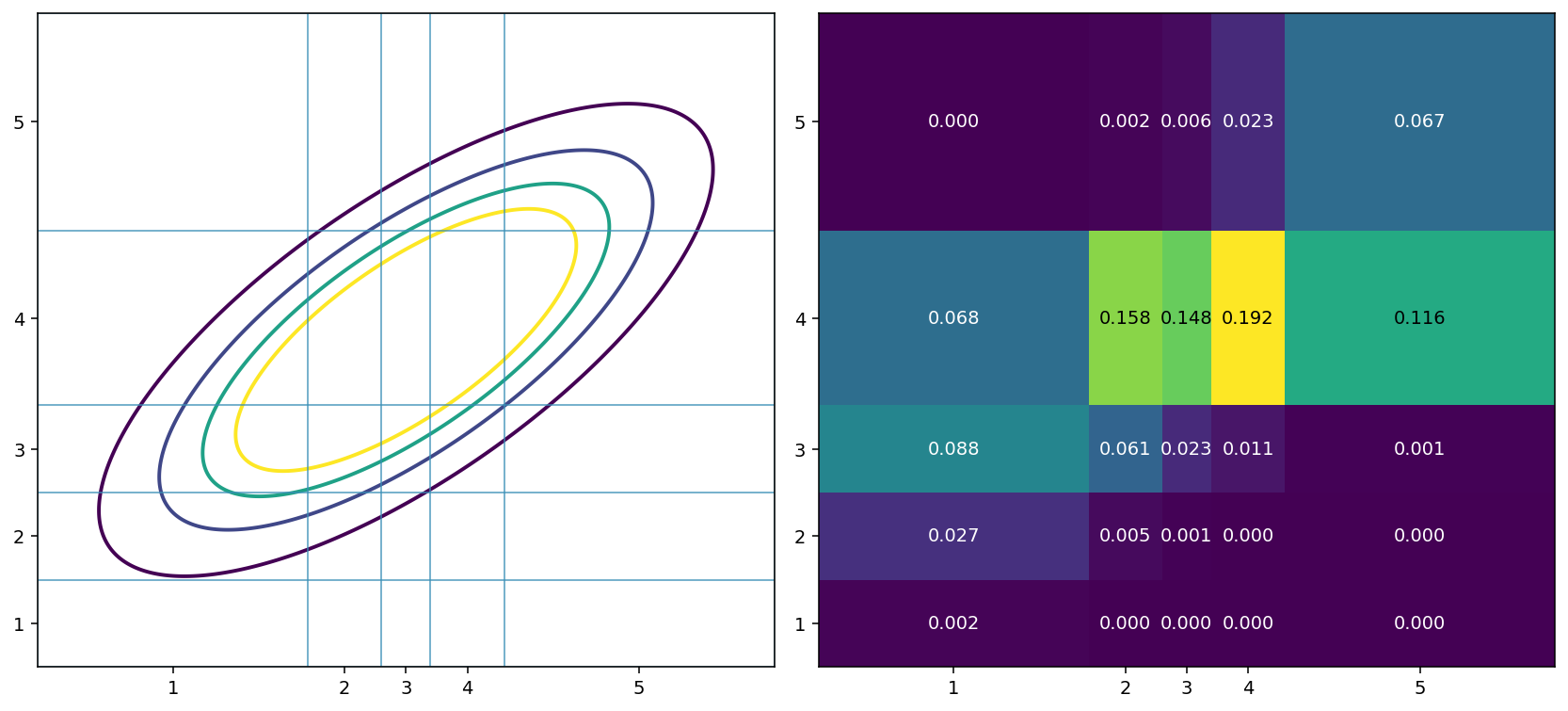}
\caption{Domain 2 begins with a joint Gaussian distribution over latent continuous variables $\vZ$ (left), and discretizes their values into fixed but irregular bins, yielding a joint distribution over observable discrete variables $\vX$ (right).  The positive correlation of $(Z_1,Z_2)$ carries over to $(X_1,X_2)$, but the probabilities in the grid at right are also affected by the bin widths.  Our experiments use 10 variables of each type, not 2 as illustrated here.}\label{fig:mg}
\end{figure}

The next domain is governed by correlated continuous random variables, which are discretized into ordinal ratings (\cref{fig:mg}).  This is a \localdefn{multivariate ordered probit model} (not conditioned on any other regressors).
Similar models are widely used for multidimensional graded (i.e., ordinal) responses in psychometrics.

The underlying continuous variables are never observed.  The classical baseline method (\cref{sec:joint-mg}) knows the entire mechanism: it explicitly reconstructs the latent values of the continuous variables and estimates their covariances and the discretization thresholds \citep{bock-aitkin-1981}.  In contrast, our \textsc{Marformer} configuration (\cref{sec:marformer-mg}) does not explicitly model the continuous variables, and does not even know the order of the discrete categories (e.g., that a \texttt{2} rating falls between \texttt{1} and \texttt{3} ratings).  The \textsc{Marformer} must discover the underlying ordinal and correlational structure, despite the irregular discretization and the fact that some of the discrete values are missing.

\subsection{True Generative Distribution}
\label{sec:true-mg}

The sample space has 10 attributes $\vi=\{1,2,\dots,10\}$ where $(\forall i\in\vi)\,\Dom_i=\{1,2,\dots,5\}$.  The complete-data distribution $P^*(\vX)$ is given by a model $\qphi(\vX)$ with a latent variable $\vZ \in \mathbb{R}^{10}$ and known parameters $\vphi=\vphi^*$:
\begin{align}
     \qphi(\vX=\vx)
     &= \int_{\vz} \qphi(\vZ=\vz,\vX=\vx)\,d\vz \\
     &= \int_{\vz} \qphi(\vZ=\vz)\prod_{i\in\vi} \qphi(X_i=x_i \mid Z_i=z_i)\,d\vz \\
     &= \int_{\vz} \qphi(\vZ=\vz)\prod_{i\in\vi} [t_{i,x_i-1} \leq z_i < t_{i,x_i}]\,d\vz
\end{align}
where $\qphi(\vZ)$ is a multivariate normal distribution $\mathcal{N}(0,\Sigma)$, and each threshold vector $\vt_i$ has elements $-\infty = t_{i,0} < t_{i,1} < \cdots < t_{i,5} = \infty$.  This construction of $\qphi(\vX)$ is illustrated in \cref{fig:mg}.

We construct the threshold vectors $\vt_i$ from underlying parameters $\vec{p}_i$.  Note that the marginal distribution of $Z_i$ is $\mathcal{N}(0,\Sigma_{ii})$.  To partition it into roughly equal bins, we take $\vec{p}_i$ to be a roughly uniform distribution over $\Dom_i=\{1,2,\dots,5\}$ and set $t_{i,k} = \sqrt{\Sigma_{ii}}\,\Phi^{-1}(p_i(1)+\dots+p_i(k))$ where $\Phi$ is the standard normal CDF.

Thus, the model parameters $\phi$ specify the positive definite covariance matrix $\Sigma$ and the distributions $\vec{p}_i$.

\paragraph{Selecting a particular distribution.}  To choose $\varphi^*$, we sample $\vec{p}_i \sim \mathrm{Dirichlet}(\frac{\alpha}{5},\dots,\frac{\alpha}{5})$, taking $\alpha=2.0$.
How about $\Sigma$?  To impose a substantive conditional independence structure (as in \cref{sec:domain-bn}), we make $\Sigma^{-1}$ sparse, corresponding to a simple path or cycle on the 10 variables.\footnote{A Gaussian distribution can be viewed as an undirected graphical model (Markov random field) over $X_1,\dots,X_{10}$, in which the zero entries of the inverse covariance matrix $\Sigma^{-1}$ correspond to missing edges.}  For example, the path case yields a tridiagonal $\Sigma^{-1}$.  The joint modeling method is given the true sparsity pattern.  The \textsc{Marformer} is not---it is not even told that there is a covariance matrix.

Specifically, we flip a fair coin to choose between the two sparsity patterns (path and cycle).  We would then like to draw the symmetric positive definite $\Sigma^{-1}$ from the Wishart prior $W_{10}(I,\nu)$, with scale matrix $I$ and degrees of freedom $\nu$, conditioned on this sparsity pattern.  For computational convenience, we impose the sparsity constraint only softly, downweighting the Wishart prior probability of $\Sigma^{-1}$ by $\prod_{ij} \exp(\beta_{ij} |\Sigma^{-1}_{ij}|)$.  We take $\beta_{ij}=-100$ for entries $ij$ that are supposed to be 0 and $\beta_{ij}=0$ otherwise. The mean absolute value of the entries intended to be zero is 0.0181. We sample $\Sigma^{-1}$ from this modified prior via Metropolis-Hastings \citep{robert2016metropolishastingsalgorithm}. We initialize the chain at an unconstrained draw from the Wishart prior, and use the proposal $\widetilde{\Sigma^{-1}} = \Sigma^{-1} + \tfrac{1}{2}(Z+Z^\top)$, where $Z$ has i.i.d.\ $\mathcal{N}(0, 0.05^2)$ entries. Proposals that are not positive definite (checked via a Cholesky decomposition) are rejected outright; remaining proposals are accepted with the usual Metropolis probability $\min(1, \exp(\log p(\widetilde{\Sigma^{-1}}) - \log p(\Sigma^{-1})))$ under the penalized log-density above. 

\paragraph{MCAR missingness mechanism.}  We apply the same simple mechanism as in \cref{sec:true-bn}.

\paragraph{Inference under the true model.}
To evaluate the accuracy of our models' predictions, we compare to the true conditional marginal distributions $\qphistar(X_i \mid \vx_\vr)$, which we estimate using Gibbs sampling with a burn-in period of 250 and sample size 1000.

\subsection{Datasets}\label{sec:data-gm}

We generate 100 instances for $\Test$.  Similar to \cref{sec:data-bn}, each $\Train$ dataset consists of $n$ incomplete training instances, with $n \in \{100, 500, 800, 1200, 1600, 2000, 2400\}$.

We repeat our experiments \mgruns times, selecting a different distribution $P^*$ each time.

\subsection{\textsc{Marformer} Configuration}\label{sec:marformer-mg}

We use the same set of hyperparameters, training procedures, and metrics as in \cref{sec:marformer-bn}.  The one difference is that our variables are no longer boolean, but range over $\{1,\dots,5\}$.  We give them all the same type $\tau$, which is parameterized by a logit vector $\vpsi \in \mathbb{R}^5$, so that $p_\psi = \mathrm{softmax}(\vpsi)$.  Its encoding function $e: x \mapsto \psi$ maps $\wildcard \mapsto (0,0,0,0,0)$, $1 \mapsto (1,0,0,0,0)$, $2 \mapsto (0,1,0,0,0)$, etc.

\subsection{Joint Model Estimation}
\label{sec:joint-mg}

We train a MAP model as a domain-specific baseline to maximize the observed data likelihood plus prior probabilities of $\Sigma^{-1}, p$. The domain model is parameterized by a Cholesky factor matrix $L$ and bin probabilities $p$, where we use $LL^T$ to obtain $\Sigma^{-1}$ and directly use $p$ to compute cumulative probability sums and real-valued bin boundaries.\footnote{To obtain real-valued boundaries, we apply the inverse CDF function to the $q_{ac}$ and multiply the result by $\sqrt{\Sigma_{aa}}$.}

For the parameters' prior probability, the model uses their prior distributions as defined in \cref{sec:true-mg}, with a Wishart prior plus soft sparsity constraints for $\Sigma^{-1}$ and a Dirichlet prior for $p$. The likelihood given an observed instance is the probability of a rectangular region in \cref{fig:mg}: a single row, column or cell, depending on which variables $X_i$ are observed.  To calculate this integral given the current parameters, we implement Genz's algorithm~\citep{83355a14-2760-31d3-afd2-16d5bbbc1183}. Genz's algorithm reparameterizes the multivariate Gaussian model into independent standard normals using Cholesky factors (transforming the integration domain to a parallelotope), further reparameterizes to transform the integration domain to a unit hypercube (changing the integrand), and then approximates the integral by Monte Carlo sampling from the hypercube.  To obtain the gradient of the log-likelihood, we hold the samples fixed and use PyTorch to backpropagate through the reparameterization to the parameters.

We use stochastic gradient ascent to locally maximize the MAP objective \cref{eq:map}, specifically using the AdamW optimizer with learning rate $0.01$, until convergence ($\epsilon < 10^{-3}$). Again we allow 10 random restarts.  We then use Gibbs sampling as in \cref{sec:true-mg} to predict missing values on the test instances. To verify our implementation, we also try initializing training at the ground-truth parameters $\vphi^*$.\footnote{Ground-truth initialization typically finds an estimate close to $\vphi^*$---perhaps even better than using infinite restarts to get the exact MAP estimate.}

\subsection{Experimental Results}
\label{sec:results-mg}

\Cref{fig:kl-mg} shows our main experimental results.  With enough training data, all \textsc{Marformer} variants achieve significantly lower KL divergence than a joint model trained with MAP, particularly when the \textsc{Marformer} is large.
With less training data, the MAP approach outperforms \textsc{Marformer} due to its incorporation of more domain knowledge. MAP again seems to get stuck in a local optimum, as demonstrated by the fact that it does much better when initialized with ground-truth parameters.

Per-prediction cross-entropy scatterplots in \cref{fig:scatter-mg} confirm that with enough training data, the \textsc{Marformer}--particularly when large---achieves lower cross-entropy than the joint model for the majority of marginal predictions, particularly on the lower-entropy marginals where there is a clearer right answer.

\textbf{Variant.}  As in \cref{sec:results-bn}, running on MAR data gets qualitatively similar results (\cref{sec:exp-mar}).

\section{Annotation as a Domain}
\newcommand{\ncrit}{K}\newcommand{\critidx}{k}\newcommand{\setcrit}{\mathcal{K}}
\newcommand{\njudge}{J}\newcommand{\judgeidx}{j}\newcommand{\setjudge}{\mathcal{J}}
\newcommand{\nitem}{I}\newcommand{\itemidx}{i}\newcommand{\setitem}{\mathcal{I}}
\newcommand{\ratesubof}[3]{#3#2#1}
\newcommand{\ratetupleof}[3]{(#3,#2,#1)}
\newcommand{\ratesub}{\ratesubof{\critidx}{\judgeidx}{\itemidx}}
\newcommand{\ratetuple}{\ratetupleof{\critidx}{\judgeidx}{\itemidx}}
\newcommand{\tensordims}{\nitem\times\njudge\times\ncrit}

\label{sec:domain-annot}

Annotation workflows exemplify the full challenge of conditional marginal prediction in a structured setting. A dataset of items must be evaluated by multiple judges on multiple criteria, but collecting every judge's ratings of every item would be prohibitively expensive.\footnote{The question of which annotations to collect next---and when enough have been gathered---is precisely the active feature acquisition problem discussed in \cref{sec:why-marginals}. At each step, one can collect the annotation with the highest VOI, and stop when further annotation would not reduce expected loss enough to justify the cost. Computing the VOI requires a model that produces calibrated marginals over all missing annotations, given the annotations observed so far.} This section evaluates the \textsc{Marformer} in this setting.

\subsection{True Generative Distribution}
\label{sec:true-annot}

An annotation dataset involves $\nitem$ items. $\njudge$ judges, and $\ncrit$ evaluation criteria.

We extend the variable indices introduced in \cref{sec:notation} to allow \defn{structured variable names}, where an index $a\in\vi$ may itself be a tuple.  In this domain, $\vi = \{1,2,\dots,\nitem\} \times \{1,2,\dots,\njudge\} \times \{1,2,\dots,\ncrit\} $, where each $\ratetuple\in\vi$ indexes a rating variable $X_{\ratesub}$ with $\Dom_{\ratesub}=\{1,\ldots,C\}$. That variable $X_{\ratesub}$ is the Likert score assigned to item $\itemidx$ by judge $\judgeidx$ on criterion $\critidx$.

The complete rating collection $\vX$ forms a complete instance, which can equivalently be viewed as a tensor in $\{1,\ldots,C\}^{\tensordims}$.
In practice only a sparse subset of entries is observed, since collecting every annotation is prohibitively expensive.

This is naturally framed as 3D matrix completion~\citep{candes2008exactmatrixcompletionconvex, keshavan2009matrixcompletionentries}.  Classical collaborative filtering recovers a  2D user–item rating matrix. The additional criterion axis here introduces richer dependencies among ratings: for a fixed $(\itemidx,\judgeidx)$ pair, ratings across criteria are jointly distributed, and for a fixed $(\itemidx,\critidx)$, 
ratings from different judges may exhibit systematic dependence.%
\footnote{Real annotation settings add a further challenge: new items may be added to a collection or new judges may join mid-study, requiring the model to generalize to entities unseen during training.}

The complete-data distribution $P^*(\vX)=\qphistar(\vX)$ belongs to a latent-variable model family $\qphi$ that we call the \defn{Compositional Projection Model} (CPM).  Recall that $\vi=\{1,2,\dots,\nitem\} \times \{1,2,\dots,\njudge\} \times \{1,2,\dots,\ncrit\} $, where $X_{\ratesub}$ denotes the rating variable indexed by $\ratetuple\in\vi$ and $\Dom_{\ratesub}=\{1,\ldots,C\}$.  The CPM encodes three structural assumptions about annotation data: items can be represented in a latent semantic space, criteria correspond to different directions of evaluation in that space, and judges exhibit systematic but partially shared preferences.

Each item $\itemidx$ has a latent embedding $\vec{e}_\itemidx\in\mathbb{R}^D$.  Rather than assigning every judge an unrelated parameter vector for every criterion, the CPM represents each judge as a mixture over $T$ latent judge prototypes.

For criterion $\critidx$ and prototype $t$, the prototype-conditioned preference vector is
$$
\sigma\!\left(
\vec{u}_\critidx + \vec{v}_t + \vec{u}_{\critidx t}
\right),
$$
where $\sigma(\cdot)$ is the sigmoid function applied elementwise.\footnote{We initially experimented with $\exp(\cdot)$ and $\mathrm{softplus}(\cdot)$ nonlinearities, but these produced more difficult posterior geometries and poorer HMC mixing. We therefore use the sigmoid nonlinearity, which bounds the preference coordinates and led to more stable inference in our Stan implementation. This was not an issue for \textsc{Marformer}; we use the sigmoid here to keep the comparison to Stan fair.}

The three terms separate shared criterion effects, shared judge-prototype effects, and criterion--prototype interactions.  The vectors $\vec{u}_\critidx,\vec{v}_t,\vec{u}_{\critidx t}\in\mathbb{R}^D$ represent, respectively, the direction associated with criterion $\critidx$, the general style of judgment associated with prototype $t$, and the deviation of prototype $t$ on criterion $\critidx$.

Each judge $\judgeidx$ has mixture weights $\vec{\alpha}_\judgeidx$ over the $T$ judge prototypes.  The effective preference vector for judge $\judgeidx$ on criterion $\critidx$ is
\[
\vec{\phi}_{\judgeidx\critidx}
=
\sum_{t=1}^{T}
\alpha_{\judgeidx t}
\sigma\!\left(
\vec{u}_\critidx + \vec{v}_t + \vec{u}_{\critidx t}
\right).
\]
Given the item embedding $\vec{e}_\itemidx$, the latent score for $X_{\ratesub}$ is
\[
z_{\ratesub}
=
\vec{\phi}_{\judgeidx\critidx}^{\top} \vec{e}_\itemidx.
\]
Thus, the score increases when the item embedding is aligned with the judge's criterion-specific preference direction.

\paragraph{Selecting a particular distribution.}
To choose the particular parameters $\vphi^*$ that define $P^*=\qphistar$, we sample the CPM parameters from an upper-level distribution. We use latent dimension $D=32$ and $T=3$ judge prototypes.  Each item embedding is sampled as
\[
\vec{e}_\itemidx \sim \mathcal{N}(\mathbf{0},\mathbf{I}_D),
\]
and the vectors $\vec{u}_\critidx$, $\vec{v}_t$, and $\vec{u}_{\critidx t}$ are sampled from zero-mean Gaussian distributions.  The interaction term $\vec{u}_{\critidx t}$ is sampled at a smaller scale than the prototype term $\vec{v}_t$, so that judge prototypes retain shared structure across criteria while still allowing criterion-specific variation.  The corresponding scale parameters are drawn from half-normal distributions.  Judge mixture weights are sampled as
\[
\vec{\alpha}_\judgeidx \sim \mathrm{Dirichlet}(\mathbf{1}_T).
\]
We set the hyperparameters of these upper-level distributions to posterior means obtained by fitting the CPM to LLM-Rubric data described in \cref{sec:data-real}; the resulting numerical values are given in \cref{tab:llmrubric-hyperparams}.  The remaining parameters used to select $P^*$ are introduced below.

The observed variable $X_{\ratesub}$ is generated from the latent score using a random-threshold ordinal model.  For each judge--criterion pair $(\judgeidx,\critidx)$, the CPM draws
\[
\vec{\pi}_{\judgeidx\critidx}
\sim
\mathrm{Dirichlet}
\left(
\frac{\kappa}{C}, \ldots, \frac{\kappa}{C}
\right),
\]
where $C=4$.  The vector $\vec{\pi}_{\judgeidx\critidx}$ controls how judge $\judgeidx$ uses the $C$ rating categories for criterion $\critidx$. Some judges may use high ratings more frequently on a given criterion, while others may use the scale more conservatively.
We choose ordinal thresholds so that, marginalizing over item embeddings and measurement noise, the category frequencies for each $(\judgeidx,\critidx)$ follow $\vec{\pi}_{\judgeidx\critidx}$.  Since $\vec{e}_\itemidx\sim\mathcal{N}(\mathbf{0},\mathbf{I}_D)$,
\[
z_{\ratesub}=\vec{e}_\itemidx^\top \vec{\phi}_{\judgeidx\critidx}
\]
is marginally Gaussian with variance $\norm{\vec{\phi}_{\judgeidx\critidx}}_2^2$.  After adding measurement noise $\epsilon_{\ratesub}\sim\mathcal{N}(0,\sigma_{\mathrm{meas}}^2)$, let
\[
s_{\judgeidx\critidx}
=
\sqrt{
\norm{\vec{\phi}_{\judgeidx\critidx}}_2^2
+
\sigma_{\mathrm{meas}}^2
}.
\]
Then $(z_{\ratesub}+\epsilon_{\ratesub})/s_{\judgeidx\critidx}$ is marginally standard normal over random items.  We therefore set
\[
\tau_{\judgeidx\critidx,1}=-\infty,\qquad
\tau_{\judgeidx\critidx,c}
=
\Phi^{-1}
\left(
\sum_{r=1}^{c-1}\pi_{\judgeidx\critidx,r}
\right)
\quad (c=2,\ldots,C),\qquad
\tau_{\judgeidx\critidx,C+1}=+\infty.
\]
Finally,
\[
X_{\ratesub}=c
\quad\Longleftrightarrow\quad
\tau_{\judgeidx\critidx,c}
\leq
\frac{z_{\ratesub}+\epsilon_{\ratesub}}{s_{\judgeidx\critidx}}
<
\tau_{\judgeidx\critidx,c+1}.
\]

The CPM is closely related to recent work on pluralistic reward modeling.  In particular, \citet{chen2024pal} represent each user as a mixture of prototypical users.  Their Model B similarly constructs a context-dependent user representation and projects an output representation onto it to obtain a scalar reward.  CPM uses this compositional structure for ordinal ratings rather than pairwise rankings: PAL's prompts correspond roughly to our criteria, its users to our judges, and its outputs to our items.\footnote{More specifically, PAL's user mixture weights correspond to $\vec{\alpha}_\judgeidx$, while its context-dependent prototype representations and output representations play roles analogous to our prototype-conditioned preference vectors and item embeddings $\vec e_\itemidx$, respectively.}  More broadly, latent representations of people and alternatives have a long history in models of human judgment; for example, \citet{coombs1950psychological} used one-dimensional latent positions to explain rankings of opinions by their similarity to a person's own position.

\subsection{Dataset Construction and Missingness}
\label{sec:data-annot}

We generate synthetic annotation data from the complete-data distribution $P^*(\vX)$ defined in \cref{sec:true-annot}. In our experiments, there are up to $\nitem=5000$ items, with $\njudge=50$ and $\ncrit=9$.

\paragraph{MCAR missingness mechanism.}
We first form a sparse annotation design by selecting, independently for each item
$\itemidx$, four judges uniformly at random without replacement, with all $\ncrit$
criteria represented for each selected item--judge pair. A further MCAR mask
partitions ratings in this design into observed context and hidden prediction targets.
Ratings outside the design are not scored. Both stages select ratings independently
of any values, so the resulting missingness mechanism is MCAR. As discussed in
\cref{sec:training}, this allows the \textsc{Marformer} to construct conditionally unbiased
training examples by masking observed ratings.

\paragraph{Test data.}
Rather than evaluating on all missing variables as the number of items grows, we designate a fixed \defn{test corner}. Let $\setitem_{\mathrm{test}}\subseteq[\nitem]$ contain $\nitem_{\mathrm{test}}=100$ items, and let $\setjudge_{\mathrm{test}}\subseteq[\njudge]$ contain $\njudge_{\mathrm{test}}=10$ judges. The test corner consists of the attributes
\[
\vi_{\mathrm{test}}
=
\setitem_{\mathrm{test}}\times\setjudge_{\mathrm{test}}\times[\ncrit]
\subseteq \vi.
\]
The dense test corner contains $100\times10\times9=9000$ potential ratings.
The sparse annotation design contains 639 of these ratings, covering 71 item--judge
pairs on 56 of the 100 test items, with all 9 criteria per pair.
Of these 639 ratings, 323 are observed context and 316 are hidden and scored.
We evaluate only these 316 targets, keeping the set of test prediction problems
fixed as the amount of training data increases.

\paragraph{Generalization settings.}
We vary the amount of available data along either the item or judge axis. In
\defn{item generalization}, predictions concern the fixed test items as more training items
become available; in \defn{judge generalization}, the analogous sweep is over judges.
Most experiments are transductive: observed ratings of test entities are available during
training as context, while their missing target ratings remain hidden.
The non-transductive variant, denoted \textsc{Marformer} NT, excludes the test entities from training
and uses their observed ratings only at prediction time. Test items or judges are therefore
new entities in this variant.

The nominal sizes $\nitem_{\mathrm{train}}$ and $\njudge_{\mathrm{train}}$ count all available
entities in the transductive setting, including test-block entities with observed context.
In the non-transductive setting, they count training entities disjoint from the test block.
The same nominal size therefore denotes different training partitions in the two protocols.
The sweeps are described in \cref{sec:results-annot}.

\subsection{\textsc{Marformer} Configuration for the Annotation Domain}
\label{sec:marformer-annot}

We instantiate the \textsc{Marformer} on the structured index set $\vi=\{1,2,\dots,\nitem\} \times \{1,2,\dots,\njudge\} \times \{1,2,\dots,\ncrit\} $ introduced above.  As usual, there is one variable token for each $a\in\vi$.  For $a=\ratetuple$, we additionally introduce entity tokens for item $\itemidx$, judge $\judgeidx$, and criterion $\critidx$, with relations
\[
a \edge{\texttt{item}} \itemidx,
\qquad
a \edge{\texttt{judge}} \judgeidx,
\qquad
a \edge{\texttt{criterion}} \critidx,
\]
together with their inverse relations as described in \cref{sec:architecture}.  Between rating variables, we use the relations
\[
a \edge{\texttt{same-item}} a',
\qquad
a \edge{\texttt{same-judge}} a',
\qquad
a \edge{\texttt{same-criterion}} a',
\]
when the corresponding components of $a=\ratetuple$ and $a'=\ratetupleof{\critidx'}{\judgeidx'}{\itemidx'}$ are equal.  These relations are symmetric and therefore do not require separate inverse relations.

In the terms of \cref{sec:architecture}, all rating variables have the same categorical type $\tau$, whose distribution family $p_\vpsi$ over $\Dom=\{1,\ldots,C\}$ is parameterized by logits $\vpsi\in\mathbb{R}^C$, so that $p_\vpsi(c)=\text{softmax}(\vpsi)_c$.  Its encoding function maps an observed rating $c$ to $e(c)=20\cdot\mathbf{1}_c$ and $\wildcard$ to $e(\wildcard)=\mathbf{0}$.  We use $C=4$.  As in \cref{sec:architecture}, the parameter stream additionally contains the wildcard bit, giving parameter-stream dimension $C+1=5$.

The feature-stream dimension is $75$, giving total token dimension $d=80$. We use $L=8$
Transformer layers, $H=4$ attention heads, and a ReLU feedforward module of width
$d_{\mathrm{ff}}=128$ with one hidden layer.
Relational attention uses
learned key and value embeddings as described in
\cref{sec:architecture}.

We train with AdamW, learning rate $2\times10^{-4}$ and weight decay $0.01$, with standard
Transformer dropout $0.1$. Each artificial mask selects $\lfloor0.15n_{\mathrm{obs}}\rfloor$
observed ratings uniformly without replacement, where $n_{\mathrm{obs}}$ is the number of
observed ratings in the training input. We draw five refreshed masks per epoch. Masked and
unmasked observed targets receive loss weights $15$ and $1$, respectively, corresponding to
$\lambda_{\mathrm{copy}}=1/15$ under \cref{eq:trainreg}. Naturally missing ratings do not enter
the training loss.

For synthetic annotation, we hold out a block containing $15\%$ of the training entities on the
generalized axis as the validation set.
We search for the epoch budget
over 2000 epochs, selecting the minimum of the
five-point-smoothed validation-loss curve. We then merge the validation block back, refit on the
full training partition for the selected number of epochs, and report the last checkpoint.
The validation block is included in the stated training size. The smallest transductive
settings have no training entities outside the test block from which to reserve validation
data. They use the axis-median selected epoch budgets: 142 epochs at
$\nitem_{\mathrm{train}}=100$ and 126 at $\njudge_{\mathrm{train}}=10$.
This exception does not apply to the non-transductive models at the same nominal sizes,
which have separate training entities and their own validation blocks.

Entity-embedding dropout is $0.7$ on the generalized axis and $0$ on the other axis. Each forward
pass uses at most $500$ items during both training and evaluation. This cap stabilizes
training and lets us scale to larger datasets within the same GPU memory budget.

\subsection{Joint-Model Estimation with Stan}
\label{sec:joint-annot}

We compare the \textsc{Marformer} to a Bayesian joint-model estimator implemented in Stan.  The correctly specified baseline is given the true CPM model family $q_\vphi(\vX)$ from \cref{sec:true-annot}, together with the same prior families and hyperparameters used to generate $P^*$.  It does not observe the realized parameters $\vphi^*$, which must instead be inferred from the observed ratings.  We refer to this baseline as \defn{Stan Oracle}: ``oracle'' means that the model family and priors are correctly specified, not that the true latent parameters are known.

Stan performs posterior inference over the CPM parameters and latent variables using Hamiltonian Monte Carlo (HMC).  In particular, it infers the item embeddings, criterion and prototype vectors, criterion--prototype interactions, judge mixture weights, rating-threshold parameters, and global scale parameters from the incomplete annotation tensor.  Because the missingness mechanism is MCAR (\cref{sec:data-annot}), it is ignorable for this inference.

For each posterior sample $\vphi^{(m)}$, the fitted CPM induces a categorical predictive distribution over the $C$ Likert categories for each missing rating $X_{\ratesub}$ using the ordinal-probit observation model from \cref{sec:true-annot}.  We average these distributions over the $M$ post-warmup HMC samples to approximate the posterior predictive marginal:
\[
q_{\mathrm{Stan}}(X_{\ratesub}=c\mid\obs{\vx})
\approx
\frac{1}{M}
\sum_{m=1}^{M}
q_{\vphi^{(m)}}(X_{\ratesub}=c\mid\obs{\vx}).
\]
The explicit calculation of the category probabilities for each posterior sample is given in \cref{sec:stan-posterior-predictive}.

Thus, unlike a plug-in MAP estimator, Stan integrates over posterior uncertainty in the latent parameters when predicting a missing rating.

We run Stan with one chain, 300 warmup iterations and 500 post-warmup sampling iterations,
\texttt{adapt\_delta}=0.85, \texttt{max\_treedepth}=12, and random initialization. The sampler
configuration is summarized in
\cref{sec:stan-convergence}.

\paragraph{Misspecified CPM variants.}
The correctly specified Stan Oracle gives the joint-model approach privileged knowledge of the data-generating family.  To study the effect of model misspecification, we also consider three restricted CPM variants.

\defn{Stan Misspec Proj} removes judge-specific projection behavior.  Rather than assigning each judge $\judgeidx$ separate mixture weights $\vec{\alpha}_\judgeidx$ over the latent prototypes, the model uses a shared prototype mixture across judges.  Judges may still differ in how they use the rating scale.

\defn{Stan Misspec Bin} retains judge-specific projection behavior but removes judge-specific rating thresholds.  For each criterion $\critidx$, all judges share the same ordinal threshold parameters, rather than using a separate calibration for each $(\judgeidx,\critidx)$ pair.

\defn{Stan Misspec Both} applies both restrictions: judges share both their prototype mixture and their criterion-specific rating thresholds.  The remaining criterion- and item-level latent structure is unchanged.

\subsection{Other Baselines}
\label{sec:annot-baselines}

In addition to the CPM-based Stan models, we compare against three simpler empirical baselines.  All three estimate missing ratings directly from the observed annotation tensor, without assuming the latent structure of the CPM.

\paragraph{Pooled Unigram.}
The \defn{Pooled Unigram} baseline predicts missing ratings using smoothed empirical
conditional frequencies. A pooling rule $S\subseteq\{\itemidx,\judgeidx,\critidx\}$ specifies
which tensor indices distinguish the distributions: $S=\emptyset$ uses global rating
frequencies, while $S=\{\judgeidx,\critidx\}$ estimates a separate distribution for each
criterion--judge pair, pooling across items. The reported baseline uses the latter fixed
rule, with Laplace smoothing and backoff to global frequencies for unseen judges.
It provides a simple comparison for what can be achieved by pooling ratings along the
criterion and judge axes.

\paragraph{Naive Bayes.}
The \defn{Naive Bayes} baseline additionally conditions on observed ratings neighboring the target variable.  For a missing target $X_{\ratesub}$ and candidate category $c$, it uses
\[
p_{\mathrm{NB}}(X_{\ratesub}=c\mid\obs{\vX})
\propto
p(c)\,
p_\nitem(\itemidx\mid c)\,
p_\njudge(\judgeidx\mid c)\,
p_\ncrit(\critidx\mid c)
\prod_{\substack{\critidx'\neq \critidx\\X_{\ratesubof{\critidx'}{\judgeidx}{\itemidx}}\ \mathrm{observed}}}
p_{\critidx',\critidx}(X_{\ratesubof{\critidx'}{\judgeidx}{\itemidx}}\mid c)
\prod_{\substack{\judgeidx'\neq \judgeidx\\X_{\ratesubof{\critidx}{\judgeidx'}{\itemidx}}\ \mathrm{observed}}}
p_{\mathrm{chgJ}}(X_{\ratesubof{\critidx}{\judgeidx'}{\itemidx}}\mid c).
\]
The first product captures correlations among criteria for the same judge--item pair: each ordered criterion pair $(\critidx',\critidx)$ has its own conditional distribution $p_{\critidx',\critidx}$.  The second captures agreement across judges rating the same criterion--item pair.  This factor is shared across judge identities, while judge-specific rating tendencies are represented separately by $p_\njudge(\judgeidx\mid c)$.  All categorical factors are estimated from the training data with smoothing, and the resulting scores are normalized over $c\in\{1,\ldots,C\}$.

\paragraph{Log-linear model.}
Finally, we use a discriminatively trained \defn{multiclass logistic regression} model with features analogous to the factors in Naive Bayes. For each candidate category $c$, the model includes bias features for the target criterion, judge, and item.  It also includes criterion-pair features $(\critidx',\critidx,X_{\ratesubof{\critidx'}{\judgeidx}{\itemidx}},c)$ for observed ratings on other criteria of the same judge--item pair; equivalently, each ordered criterion pair $(\critidx',\critidx)$ has a learned $C\times C$ compatibility table.  Cross-judge evidence is represented by a shared $C\times C$ compatibility table between an observed rating $X_{\ratesubof{\critidx}{\judgeidx'}{\itemidx}}$ and the candidate rating $c$.  Contributions from multiple observed neighbors add in log-space.  Unlike Naive Bayes, this model makes no conditional-independence assumption: its feature weights are estimated jointly by regularized conditional maximum likelihood.

All three empirical baselines are transductive in our experiments: observed ratings in the test corner may be used when predicting its missing entries.  We separately report a non-transductive \textsc{Marformer} variant, allowing us to evaluate whether the \textsc{Marformer} remains competitive when it is denied this transductive training advantage.

\subsection{Experiments and Results}
\label{sec:results-annot}

We evaluate conditional marginal prediction under the item- and judge-generalization settings defined in \cref{sec:data-annot}.  In both settings, predictions are evaluated on the missing ratings in the fixed test corner.

\begin{figure}[h!]
 \centering
 \begin{subfigure}{0.48\linewidth}
   \centering
   \includegraphics[width=\linewidth]{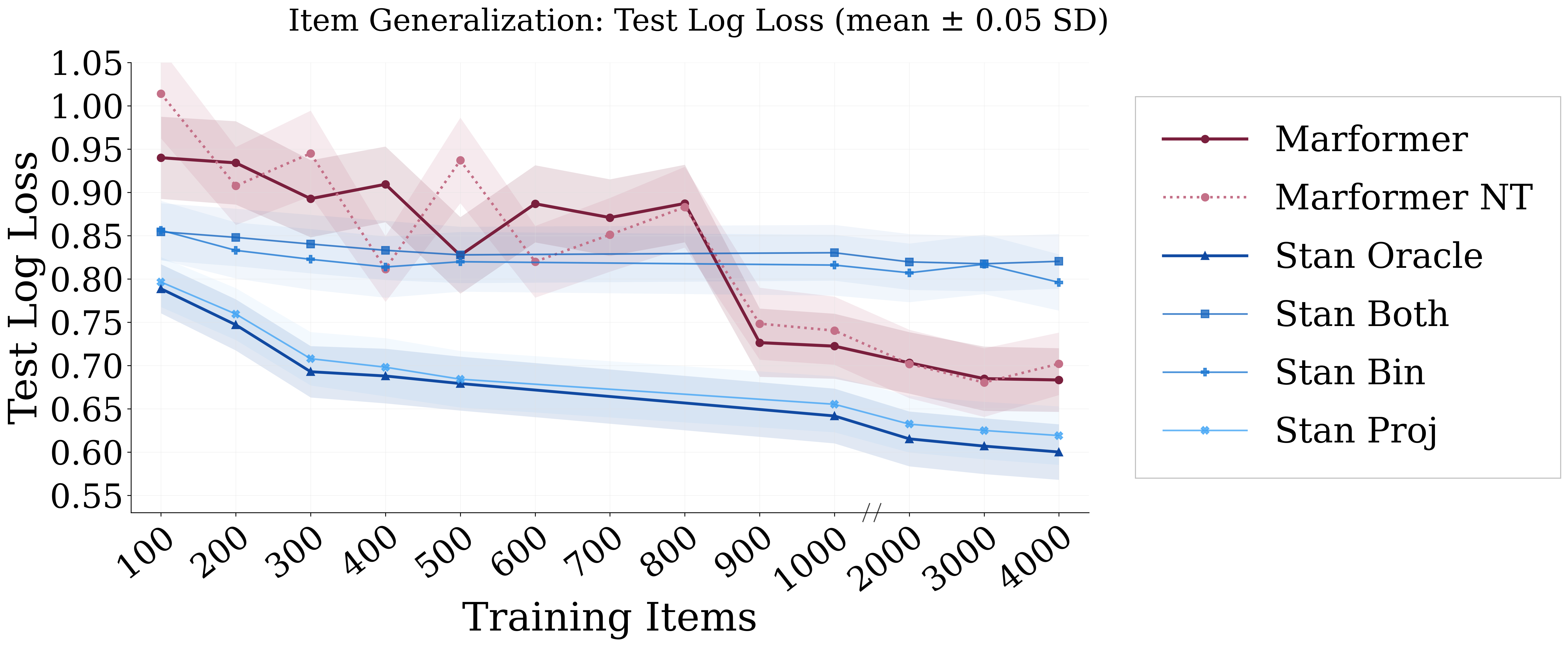}
   \caption{Log-loss: item generalization}
   \label{fig:dom3-ll-item}
 \end{subfigure}
 \hfill
 \begin{subfigure}{0.48\linewidth}
   \centering
   \includegraphics[width=\linewidth]{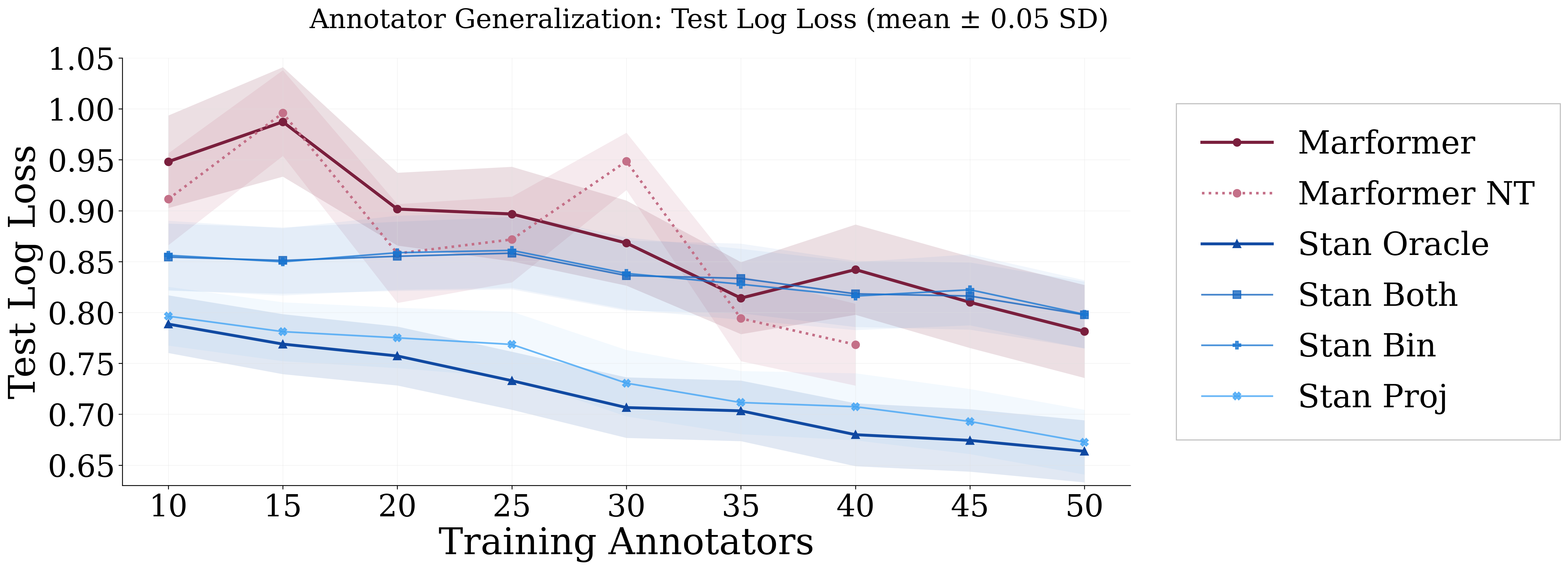}
   \caption{Log-loss: judge generalization}
   \label{fig:dom3-ll-anno}
 \end{subfigure}
 \\[0.4em]
 \begin{subfigure}{0.48\linewidth}
   \centering
   \includegraphics[width=\linewidth]{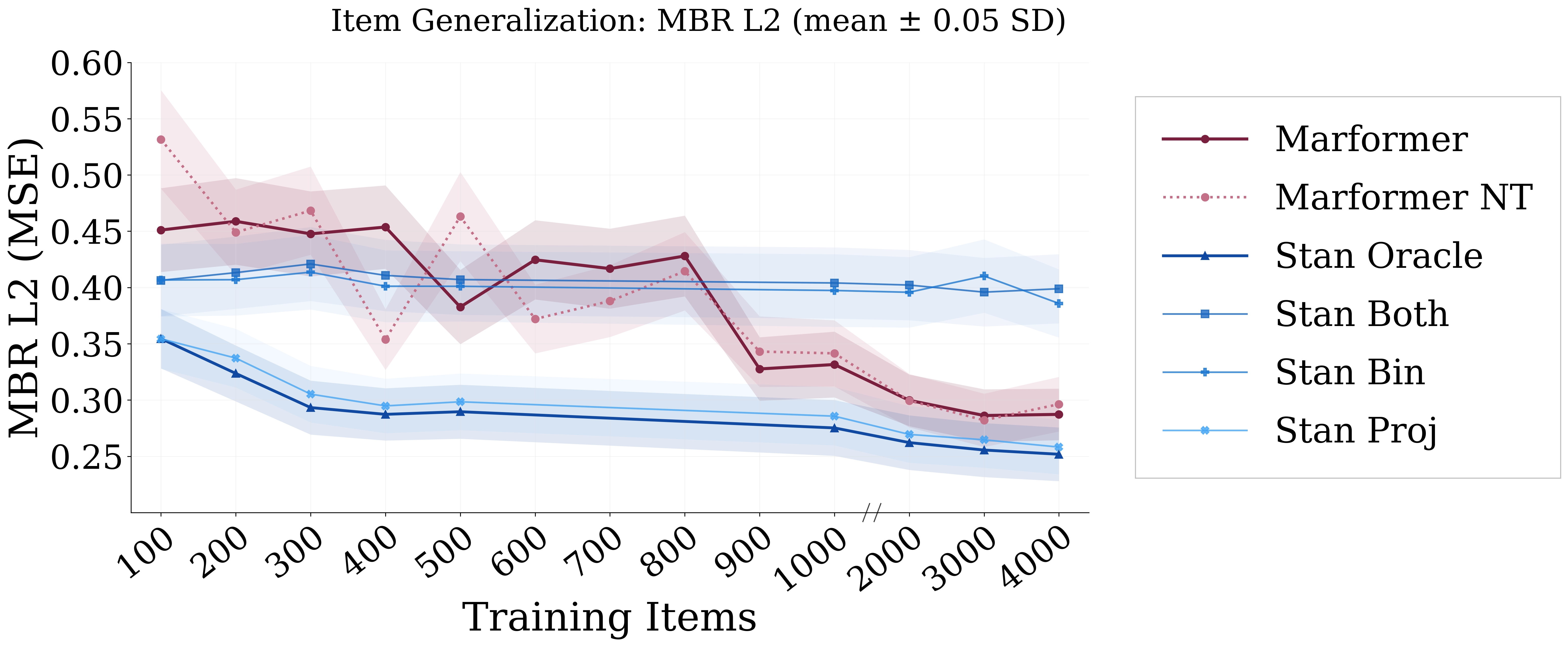}
   \caption{MBR-$L_2$: item generalization}
   \label{fig:dom3-mbrl2-item}
 \end{subfigure}
 \hfill
 \begin{subfigure}{0.48\linewidth}
   \centering
   \includegraphics[width=\linewidth]{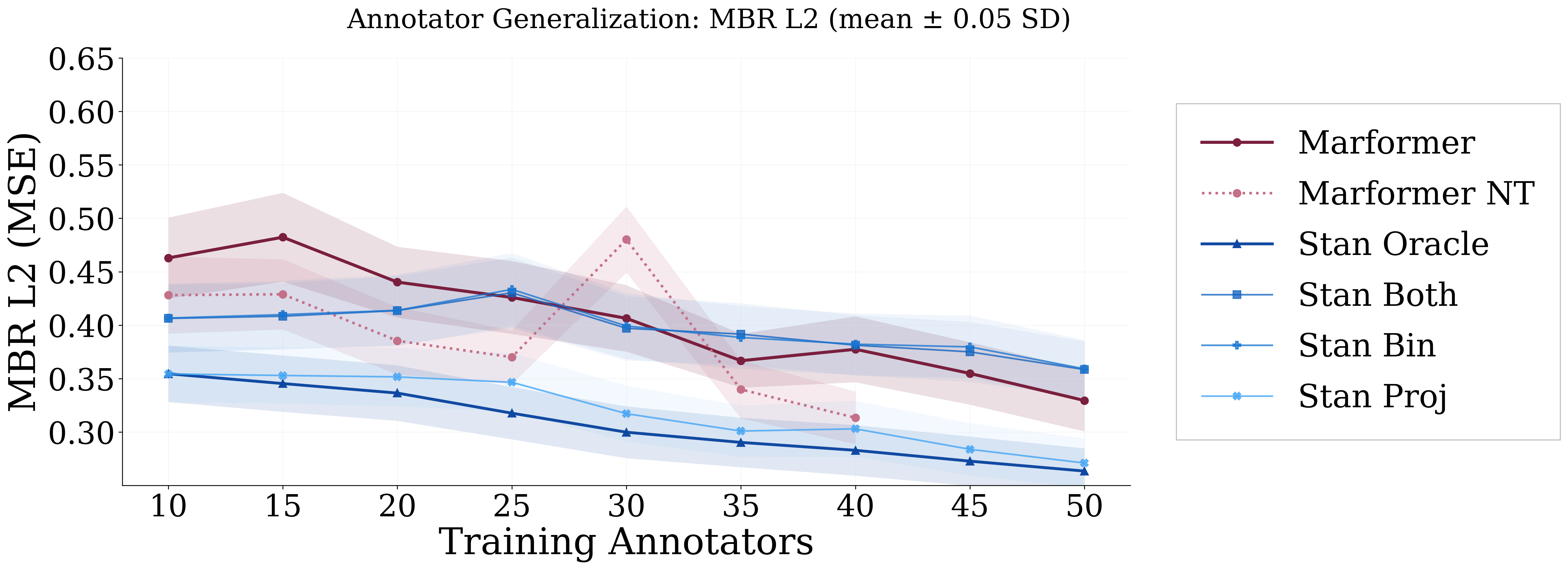}
   \caption{MBR-$L_2$: judge generalization}
   \label{fig:dom3-mbrl2-anno}
 \end{subfigure}
 \caption{Log-loss (a)--(b) and MBR-$L_2$ (c)--(d) on missing ratings in the fixed test corner as the amount of training data increases. Item-generalization results extend to $\nitem_{\mathrm{train}}=4000$ for all models; in judge generalization, \textsc{Marformer} NT stops at $\njudge_{\mathrm{train}}=40$ because test judges
are excluded from training. Shaded bands show $\pm$ one bootstrap standard error of the mean,
resampling items on the item axis and judges on
the judge axis. Full comparisons
including the empirical baselines are reported in \cref{sec:additional-domain3}.}
 \label{fig:dom3-results}
\end{figure}

\begin{figure}[h!]
 \centering
 \begin{subfigure}{0.48\linewidth}
   \centering
   \includegraphics[width=\linewidth]{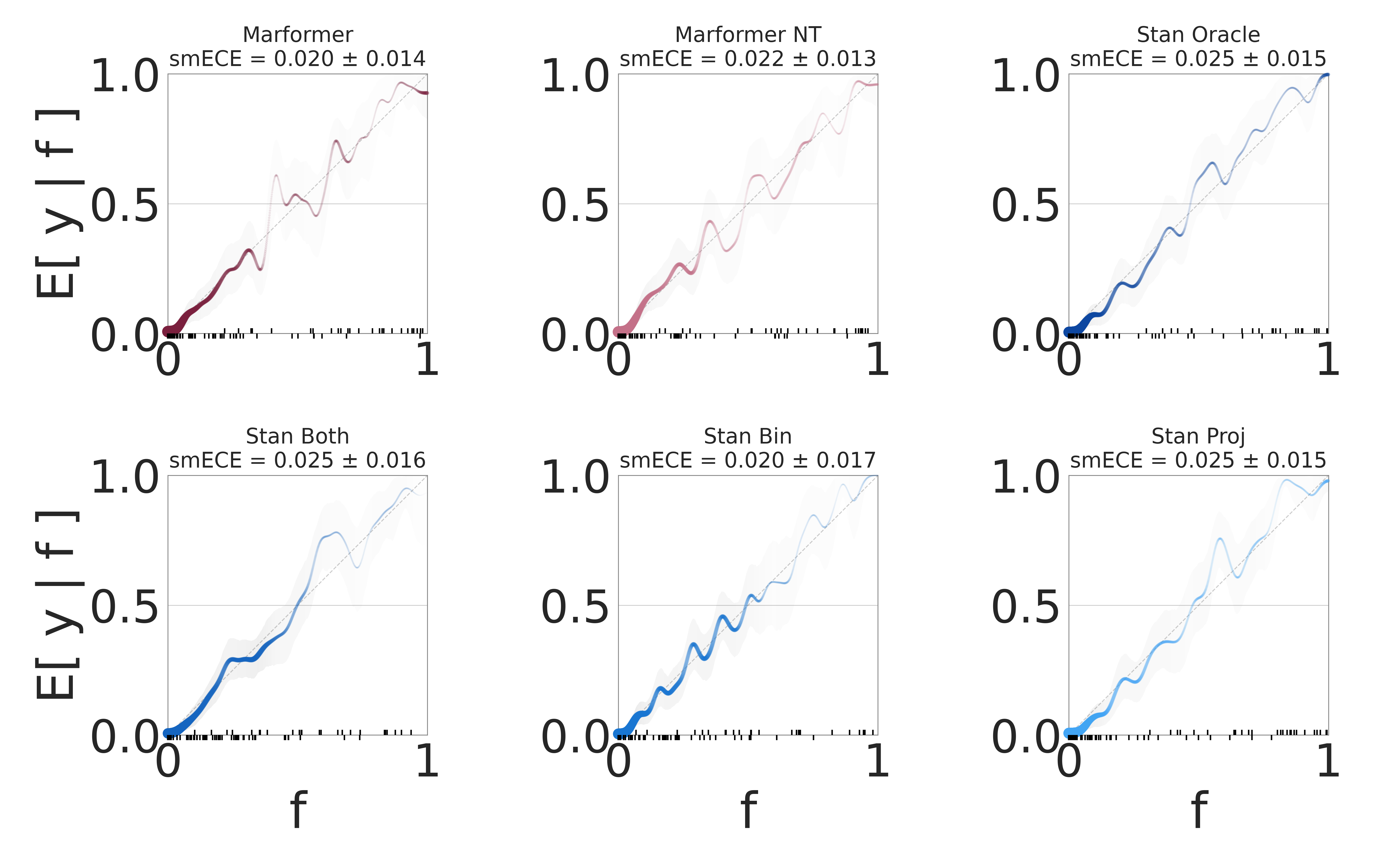}
   \caption{Item generalization ($\nitem_{\mathrm{train}}=4000$)}
   \label{fig:dom3-cal-item}
 \end{subfigure}
 \hfill
 \begin{subfigure}{0.48\linewidth}
   \centering
\includegraphics[width=\linewidth]{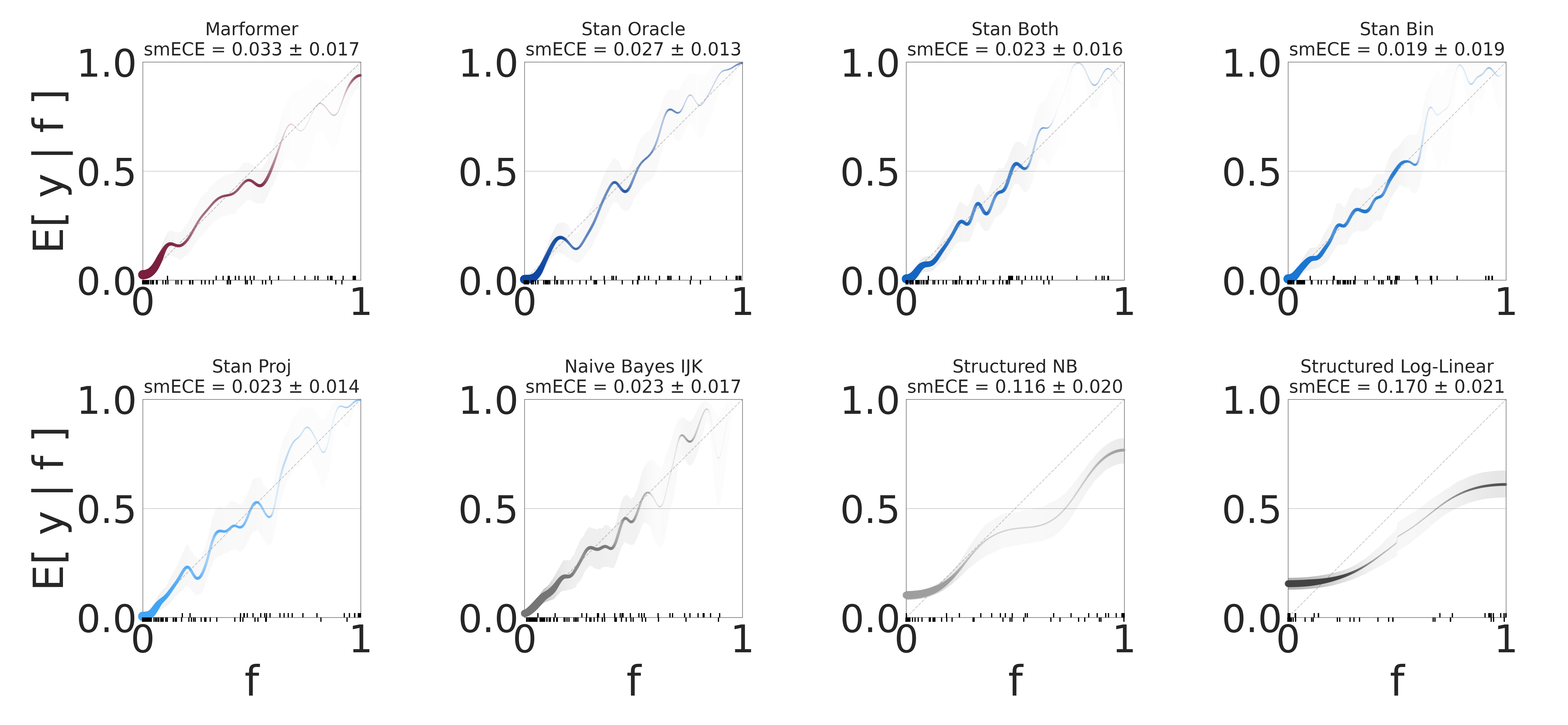}
   \caption{Judge generalization
($\njudge_{\mathrm{train}}=40$)}
   \label{fig:dom3-cal-anno}
 \end{subfigure}
 \caption{Calibration of predicted categorical
probabilities against empirical frequencies at
$\nitem_{\mathrm{train}}=4000$ and
$\njudge_{\mathrm{train}}=40$, the largest
training sizes with results for all models shown.
Probabilities for all
$C=4$ Likert categories are pooled in the reliability calculation. The diagonal indicates
perfect calibration; the displayed smECE widths are \texttt{relplot}'s reported
calibration-error widths from 500 bootstrap
resamples.}
  \label{fig:dom3-cal}
\end{figure}

\paragraph{Metrics.}
We evaluate predictive distributions using log-loss and the quality of their induced point predictions using MBR-$L_2$ (\cref{fig:dom3-results}).  For a predicted categorical distribution $p(X_{\ratesub}=c\mid\obs{\vx})$, the Bayes-optimal prediction under squared error is its conditional mean,
\[
\hat x_{\ratesub}
=
\sum_{c=1}^{C}
c\cdot p(X_{\ratesub}=c\mid\obs{\vx}),
\]
and MBR-$L_2$ measures the resulting squared
prediction error.  We additionally evaluate
probabilistic calibration using smooth expected
calibration error (smECE; \cref{fig:dom3-cal}).

\paragraph{Experimental sweep.}
For item generalization, we vary
\[
\nitem_{\mathrm{train}}
\in
\{100,200,\ldots,1000\} \cup \{2000,3000,4000\}.
\]
At $\nitem_{\mathrm{train}}=100$, the available items consist only of the 100 test-corner items.  Both Stan and the \textsc{Marformer} are evaluated up to $\nitem_{\mathrm{train}}=4000$ to compare their predictions as more data become available.

For judge generalization, we vary
\[
\njudge_{\mathrm{train}}
\in
\{10,15,20,25,30,35,40,45,50\}.
\]
We evaluate the standard transductive \textsc{Marformer} and a non-transductive variant, denoted
\textsc{Marformer} NT, which excludes whole test entities
during training.

Empirical baselines are excluded from the main-text figures for clarity; full comparisons appear in \cref{sec:additional-domain3}.

\paragraph{Predictive performance.}

In item generalization, \cref{fig:dom3-results}(a) shows that log-loss improves overall as more
training items become available, while a gap to the correctly specified Stan CPM remains.
At $\nitem_{\mathrm{train}}=4000$, log-loss is $0.683$ for the \textsc{Marformer} versus $0.600$ for Stan.
The non-transductive \textsc{Marformer} follows a similar overall trend. We do not interpret small
differences between individual training-size points as evidence of a systematic advantage for
either \textsc{Marformer} variant. Stan is given the correct generative model family, whereas the \textsc{Marformer}
learns the conditional structure from the observed ratings.

The corresponding MBR-$L_2$ results in \cref{fig:dom3-results}(c) show the same qualitative behavior for point prediction.  As the number of training items increases, the conditional means produced by the \textsc{Marformer} improve overall, but remain less accurate than those obtained from the correctly specified joint model.

In judge generalization, \cref{fig:dom3-results}(b)--(d) shows that both the \textsc{Marformer} and the correctly specified Stan CPM improve overall as more training judges become available, with the \textsc{Marformer} remaining competitive throughout.  The comparison with the misspecified Stan variants and empirical baselines is reported in full in \cref{sec:additional-domain3}.

\paragraph{Calibration.}
Because the goal is conditional marginal prediction, good point-prediction performance alone
is insufficient: predicted probabilities should also reflect empirical frequencies.
\cref{fig:dom3-cal} examines calibration at the largest available training sizes.
In item generalization at $\nitem_{\mathrm{train}}=4000$, smECE is $0.020$ for the \textsc{Marformer},
$0.022$ for \textsc{Marformer} NT, and $0.025$ for the correctly specified Stan CPM.
These estimates indicate calibration comparable to the oracle baseline at this training size.
At smaller item-training sizes
($\nitem_{\mathrm{train}}=200$--$500$), the
\textsc{Marformer} has higher smECE than Stan Oracle.
The conditional log-loss objective in \cref{eq:cmle} rewards matching conditional
distributions on artificially masked observations rather than only their most likely values.
The layer-by-layer development of these
predictions is examined in \cref{sec:logit-lens}
using logit and tuned lenses: learned readouts recover predictions close to the
final model's by layer 3, while the raw parameter coordinates continue to improve
through the final layers.

Conditional predictions also support chained sampling and VOI-based acquisition
(\cref{sec:why-marginals}).

\paragraph{Computational cost.}
For synthetic annotation, we account for both validation-based epoch selection and the
subsequent refit. The reported runs used a 2000-epoch search budget; for runtime accounting,
we give a hypothetical cost estimate for ending the search 50 epochs after the
selected epoch, then performing the measured refit.
For selected epoch $e_*$, measured full-search time $t_{\mathrm{search}}$, and measured refit
time $t_{\mathrm{refit}}$, the estimate is
\[
\widehat t_{\mathrm{total}}
= \frac{e_*+50}{2000}\,t_{\mathrm{search}} + t_{\mathrm{refit}}.
\]
This estimate uses the recorded search and refit times and assumes approximately constant
cost per epoch and that an earlier stopping rule would reach the selected epoch.
A patience-50 rule could instead stop before a later validation minimum; this estimate
therefore does not establish the cost of recovering the reported checkpoint.
For transductive \textsc{Marformer} at $\nitem_{\mathrm{train}}=4000$, the estimated total is 60.7 minutes
(including 27.7 minutes for refitting), compared with about 757 minutes for Stan Oracle,
a roughly $12\times$ difference between this hypothetical estimate and the measured Stan cost. At $\nitem_{\mathrm{train}}=500$, the
corresponding totals are 2.5 and about 52 minutes. On the judge axis, the \textsc{Marformer} totals
are 1.6 minutes at $\njudge_{\mathrm{train}}=40$ and 2.9 at 50; Stan takes about 55 minutes
at 40. Runtime curves appear in \cref{fig:dom3-runtime}.

\section{Applying \textsc{Marformer} to Real Annotation Data}
\label{sec:domain-real}

Real annotation datasets differ from the synthetic setting in two important ways. Missingness follows a structured study-design protocol rather than the i.i.d.\ Bernoulli mechanism of Domains 1--2. Applying the masking justification requires the additional assumption that the retained human ratings are MCAR; we do not verify this assumption from the processed data.\footnote{Random judge assignment would support MCAR if assignment and subsequent retention were independent of rating values. The sparse pattern alone does not establish this independence.} Second, there is no known ground-truth generative distribution and hence no oracle model for comparison. We evaluate in a fully transductive setting and focus on generalization across items, with the full judge set fixed throughout.

Real annotation datasets also motivate a broader use of conditional marginal prediction: using abundant, inexpensive measurements to predict the outcomes of measurements that are costly to obtain.  Here, for example, LLM ratings can be collected much more cheaply and densely than human ratings, so that observed LLM ratings may help predict the human ratings that were not collected.  The same pattern appears in other domains.  DataDecide~\citep{magnusson2025datadecide}, for example, studies whether inexpensive small-scale language-model training experiments can predict which pretraining choices will perform best at much more expensive scales.  Sim-to-real methods similarly exploit inexpensive simulated experience to reduce the amount of costly real-world data required to train or evaluate robotic systems~\citep{zhao2020simtoreal}. Omnimodal Encoders~\citep{harwath2026omnimodal} consider an analogous problem in multimodal modeling: developing inexpensive intrinsic evaluations of encoders that predict their downstream performance after integration into a full multimodal model, without requiring every candidate encoder to be evaluated by training the full system. In each case, inexpensive observations are valuable insofar as they provide information about the outcomes of more expensive experiments; conditional marginal prediction provides a natural way to represent both the prediction and the remaining uncertainty.

The predicted marginals can also be propagated
into approximate uncertainty about downstream
population statistics. For example, suppose we
wish to estimate the difference in mean human
rating between two populations of items when some
human ratings are missing.  We can repeatedly
impute missing ratings by chaining their predicted conditional marginals
(\cref{eq:chain}) and recompute the population difference, obtaining a distribution over the
difference that reflects uncertainty about the
unobserved ratings.

In addition to the Stan CPM and empirical
baselines described in
\cref{sec:joint-annot,sec:annot-baselines}, we
compare against two neural imputation methods
adapted for categorical prediction.
\defn{MIWAE}~\citep{mattei2019miwaedeepgenerativemodelling}
trains a deep generative model via an
importance-weighted ELBO over incomplete
observations; we replace its reconstruction head
with $C$-dimensional logits to produce Likert
distributions.
\defn{ReMasker}~\citep{du2023remaskerimputingtabulardata}
applies masked autoencoding to tabular data,
iteratively re-masking observed entries and
training a Transformer to reconstruct them; we
similarly adapt its output head to produce $C$
logits. Both methods treat the data as a flat
table without the entity-structured relational
attention used by \textsc{Marformer}.

The Stan CPM baseline is fit here with weakly informative priors, allowing the observed data to determine the posterior rather than calibrating the priors to a known data-generating distribution as in the synthetic experiments.\footnote{Unlike the CPM, \textsc{Marformer} requires no prior specification. The posterior scale hyperparameters inferred by Stan CPM from the real data are reported in \cref{sec:additional-domain3-real}. These posterior estimates also informed the scale choices used to construct the synthetic CPM in \cref{sec:true-annot}, grounding the controlled synthetic evaluation in magnitudes observed on real annotation data.}

\subsection{LLM-Rubric}
\label{sec:data-real}\label{sec:results-real}

Multi-judge datasets have been used to study subjective judgments in NLP, including disagreement among human judges \citep{chhun-etal-2022-human}. We focus on LLM-Rubric~\citep{hashemi-etal-2024-llm}, which combines sparse human judgments with dense LLM judgments on the same items. The dataset contains 225 conversations from a human--AI information-seeking task, each evaluated on $\ncrit=9$ rubric dimensions, including naturalness, conciseness, and citation quality. Our annotation tensor includes $\njudge=25$ judges: 24 human judges and one LLM judge (GPT-4o-mini). Ratings use a $C=4$ Likert scale; human ratings are hard labels, while the LLM provides a probability distribution over the four categories. Conversations have ratings from up to 3 of the 24 human judges and from the LLM. The human annotation tensor is therefore approximately $3/24=12.5\%$ dense per criterion, while the LLM slice is fully observed.\footnote{This sparsity pattern is characteristic of annotation studies in which LLM judgments are cheaper to obtain than human judgments. Because LLM judgments may be biased, an imputation method should produce predictions that are calibrated against human judgments.} We use a 175/25/25 train/validation/test split over conversations and follow the item-generalization protocol of \cref{sec:data-annot}, varying $\nitem_{\mathrm{train}} \in \{10, 20, 30, 40, 50, 75, 100, 125, 150, 175\}$.

For the 25 test conversations, we hide and score all 666 collected human ratings:
74 conversation--judge pairs across all 9 criteria. Of these conversations,
24 have ratings from three human judges and one has ratings from two.
The 225 LLM rating distributions ($25\times9$) remain visible as context and are
not evaluation targets. Uncollected human ratings are not scored.
No human rating from a test conversation is available as input or as a training target.
In transductive training, test conversations enter the graph only through their
LLM ratings; human supervision comes from the training split. The same test targets
and observed context are used for all evaluated methods.

\paragraph{\textsc{Marformer} Configuration}
\label{sec:marformer-real}

We use the architecture and optimization
hyperparameters described in
\cref{sec:marformer-annot}, with item-embedding
dropout $\delta_\itemidx=0.7$ and the following
domain-specific details. For LLM ratings, the
encoding function receives the full categorical
distribution in place of a one-hot human rating,
allowing the dense LLM slice to enter the graph as
soft evidence. We train for $300$ epochs and
select the checkpoint with the lowest loss on the
$25$-conversation validation split, retaining that checkpoint without refitting.
Each forward pass is restricted to $10$ items during both training and evaluation.

The architecture accepts any subset of human and LLM ratings as context.
The training masking mechanism determines which conditional prediction tasks receive
supervision (\cref{sec:training}). Structured masks can therefore target particular
groups of ratings---for example, human ratings conditioned on LLM ratings---while
using the same architecture. As with other masking schemes, their statistical
justification depends on the missingness assumptions described in \cref{sec:training}.

\begin{figure}[h!]
  \centering
  \begin{subfigure}{0.48\linewidth}
    \centering
    \includegraphics[width=\linewidth]{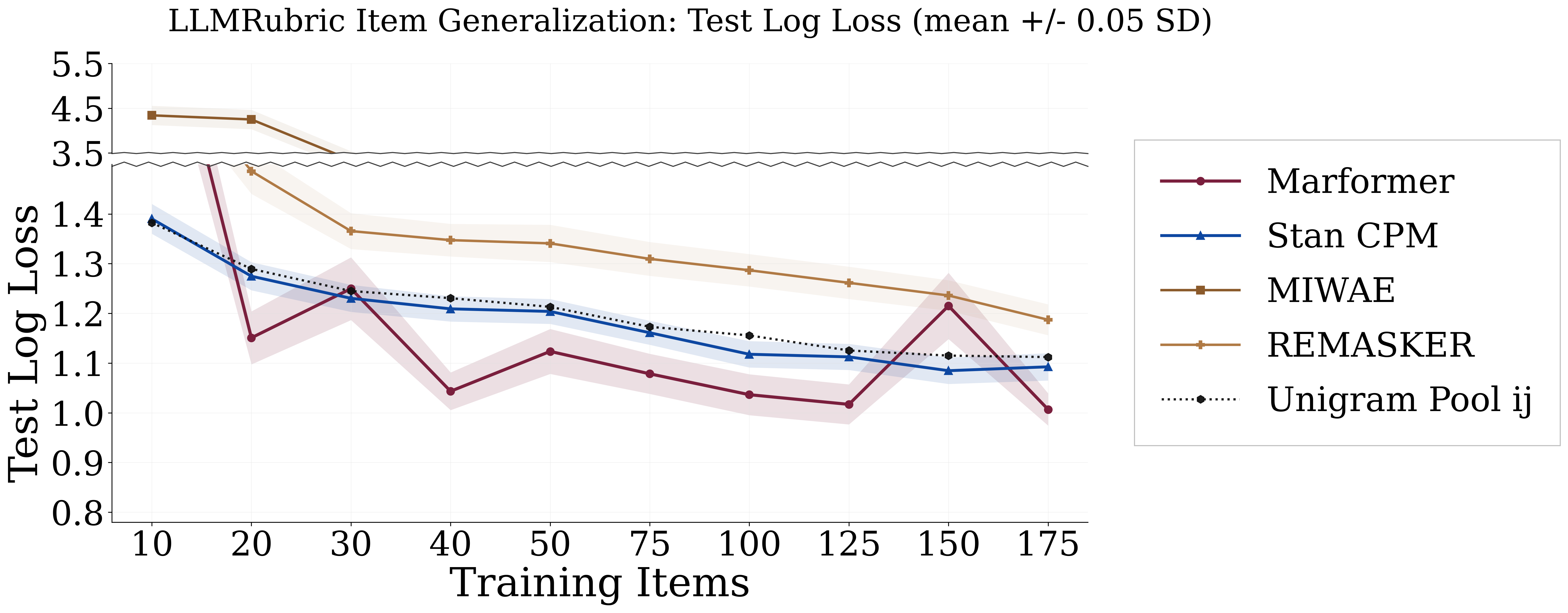}
    \caption{Log-Loss}
    \label{fig:llmrubric-ll}
  \end{subfigure}
  \hfill
  \begin{subfigure}{0.48\linewidth}
    \centering
    \includegraphics[width=\linewidth]{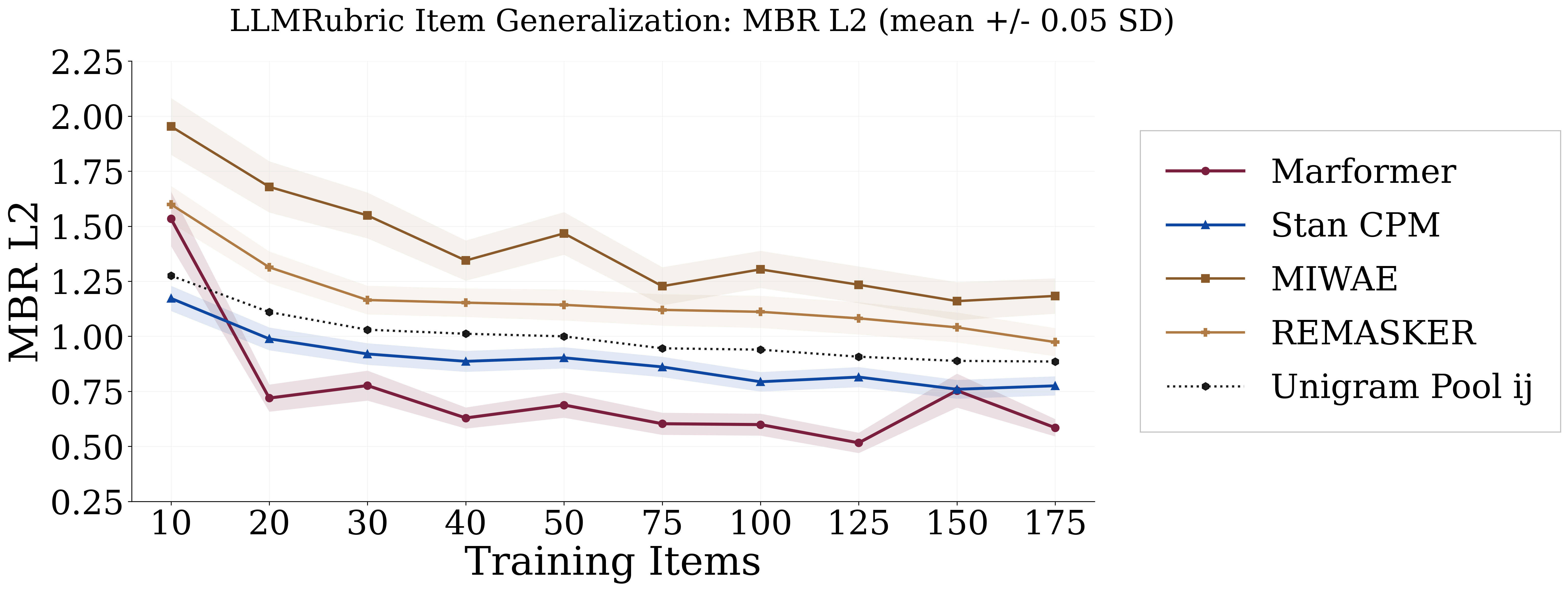}
    \caption{MBR-L2}
    \label{fig:llmrubric-mbrl2}
  \end{subfigure}
  \caption{Log-loss (a) and MBR-L2 (b) on the
LLMRubric test set as a function of training
items. Stan CPM is run with the HMC configuration
described in \cref{sec:stan-convergence}. Bands
are bootstrap standard errors of the mean, obtained by resampling conversations
conditional on the fitted model and observed data realization. Runtime curves appear in
\cref{sec:additional-domain3-real}.}
  \label{fig:llmrubric-results}
\end{figure}

\begin{figure}[h!]
  \centering
  \begin{subfigure}{0.48\linewidth}
    \centering
    \includegraphics[width=\linewidth]{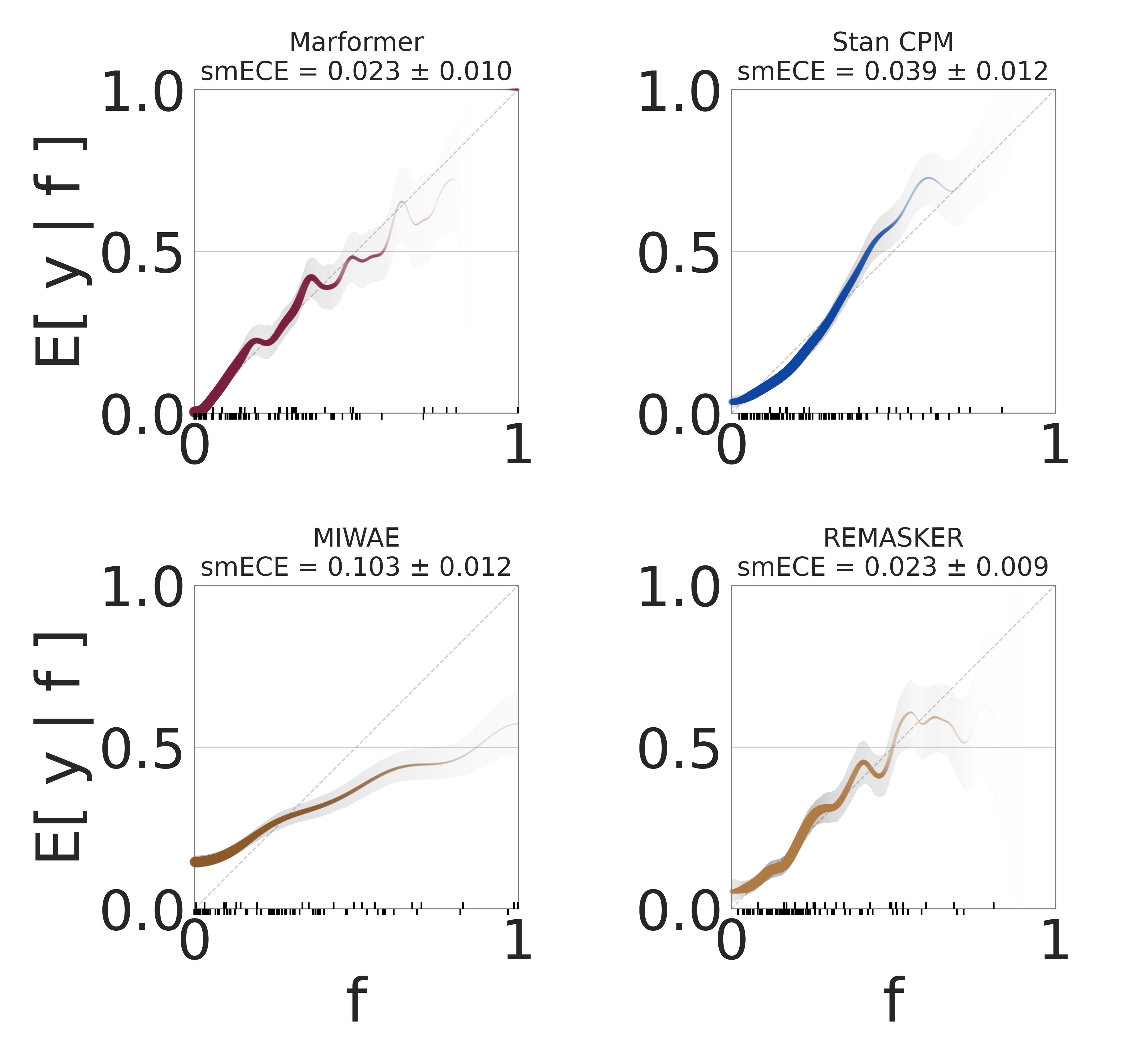}
    \caption{$\nitem_{\mathrm{train}}=175$}
    \label{fig:llmrubric-cal-175}
  \end{subfigure}
  \hfill
  \begin{subfigure}{0.48\linewidth}
    \centering
    \includegraphics[width=\linewidth]{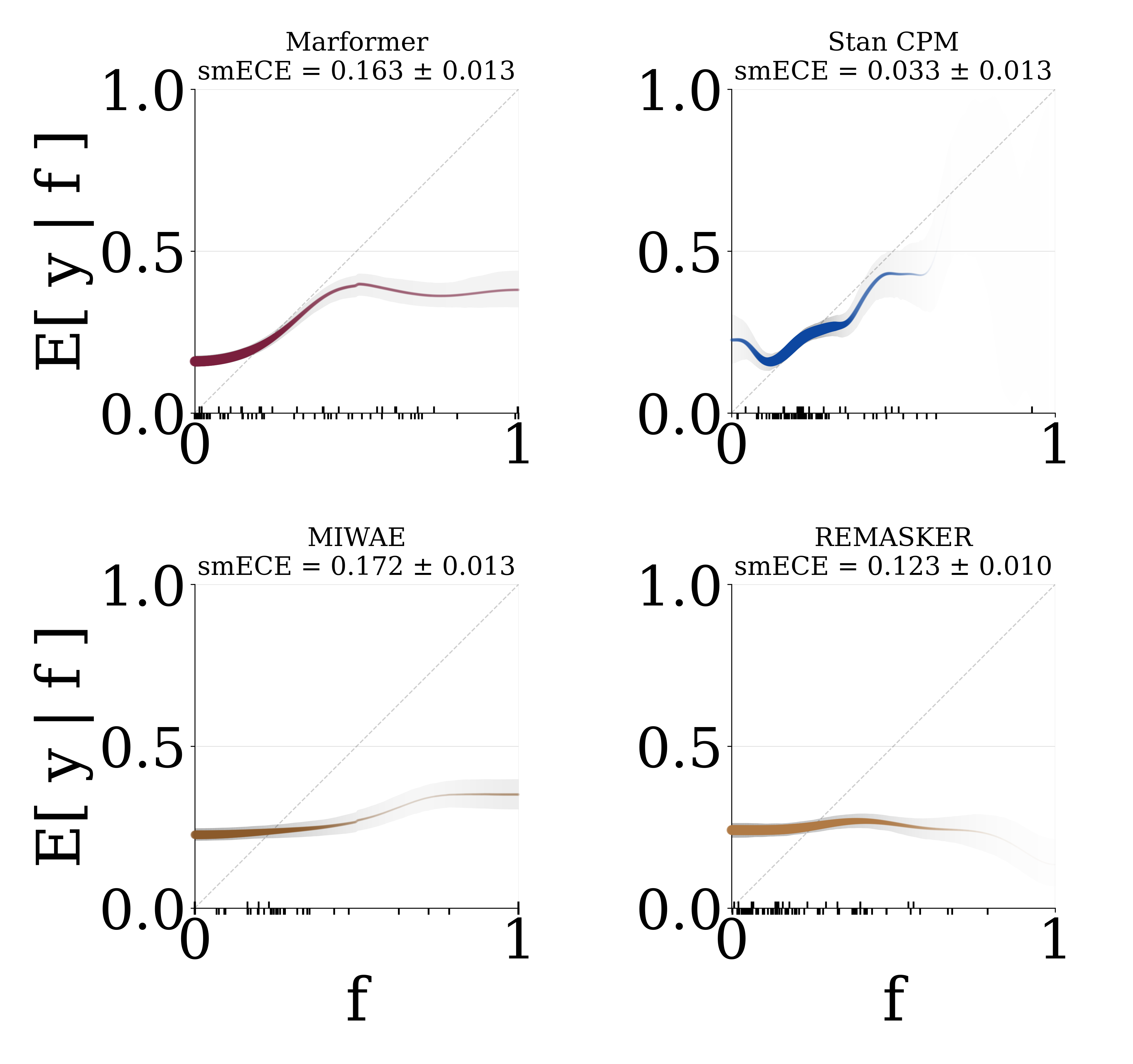}
    \caption{$\nitem_{\mathrm{train}}=10$}
    \label{fig:llmrubric-cal-10}
  \end{subfigure}
  \caption{Calibration plots (predicted
probability vs.\ empirical frequency) on LLMRubric
at full and minimal training size. At
$\nitem_{\mathrm{train}}=175$, \textsc{Marformer} is
well-calibrated; at $\nitem_{\mathrm{train}}=10$,
calibration degrades for all methods, with MIWAE
and ReMasker exhibiting overconfident predictions
at the scale extremes.}
  \label{fig:llmrubric-cal}
\end{figure}

\cref{fig:llmrubric-results}(a) shows log-loss as
a function of training size. At
$\nitem_{\mathrm{train}}=10$, Stan CPM (1.39) and
Pooled Unigram (1.38) lead \textsc{Marformer} (1.95). At
$\nitem_{\mathrm{train}}=50$, \textsc{Marformer} has lower
log-loss: 1.12 against 1.20 for Stan CPM. The
\textsc{Marformer} improves overall, though
non-monotonically, reaching $1.007$ at
$\nitem_{\mathrm{train}}=175$ versus $1.093$ for
Stan CPM and $1.112$ for Pooled Unigram. Unlike
the synthetic setting, Stan CPM here has no
privileged access to the correct generative
family, and its early-data advantage is reversed
at the largest training size.
\cref{fig:llmrubric-results}(b) also favors
\textsc{Marformer} on MBR-L2 at
$\nitem_{\mathrm{train}}=175$: $0.59$ versus
$0.78$ for Stan CPM. \cref{fig:llmrubric-cal}
shows calibration at representative training
sizes: at $\nitem_{\mathrm{train}}=175$, \textsc{Marformer}
is well-calibrated; at
$\nitem_{\mathrm{train}}=10$, it is less
well-calibrated than Stan CPM. The recorded
real-data training times at
$\nitem_{\mathrm{train}}=175$ are approximately
22 minutes for \textsc{Marformer} and 162 minutes for Stan CPM. The \textsc{Marformer} timing includes the 300-epoch training run and best-validation checkpoint selection, without refitting. Runtime curves appear in \cref{sec:additional-domain3-real}.

\section{Other Related Work}
\label{sec:related}

\textbf{Classical and generative methods.} Early work on imputation is largely statistical or generative in nature. Classical approaches include multiple imputation by chained equations (MICE)~\citep{JSSv045i03}, nuclear-norm matrix completion (\textsc{SoftImpute})~\citep{specreg}, and nonparametric methods such as random-forest imputation (\textsc{MissForest})~\citep{Stekhoven2011MissForestN}. These remain widely used but often require model selection per dataset and struggle with heterogeneous data. Generative models attempt to overcome these limitations by modeling the joint distribution of features: \textsc{GAIN}~\citep{yoon2018gainmissingdataimputation} and \textsc{MisGAN}~\citep{li2019misganlearningincompletedata} rely on adversarial training, while \textsc{HI-VAE}~\citep{nazabal2020handlingincompleteheterogeneousdata}, \textsc{VAEAC}~\citep{ivanov2018variational}, and \textsc{MIWAE}~\citep{mattei2019miwaedeepgenerativemodelling} provide likelihood-based training and probabilistic imputations, with extensions for MNAR data in the case of \cite{ipsen2021notmiwaedeepgenerativemodelling}. These methods model distributions over missing values, while our work focuses on individual conditional marginals, which are directly aligned with downstream decision tasks.\footnote{An example of such tasks is active feature acquisition, where conditional marginals allow a principled choice of the next variable to observe via VOI (\cref{sec:why-marginals}).}

\textbf{Transformers for tabular modeling.} The success of the Transformer architecture~\citep{vaswani2017attentionneed} has motivated its application to tabular data. \textsc{TabTransformer}~\citep{huang2020tabtransformertabulardatamodeling}, \textsc{FT-Transformer}~\citep{gorishniy2023revisitingdeeplearningmodels}, and \textsc{SAINT}~\citep{somepalli2021saintimprovedneuralnetworks} demonstrate that attention-based tokenization can outperform classical learners on supervised tasks. More recently, these architectures have been adapted for imputation. \textsc{NAIM}~\citep{caruso2025imputationmethodtransformerbasedmodel} introduces feature-specific embeddings with masked self-attention to model conditional structure, while \textsc{UnmaskingTrees}~\citep{mccarter2025unmaskingtreestabulardata} offers a competitive non-neural alternative via gradient-boosted ``unmasking'' steps. Our approach differs in that we work with structured variable names encoded through compositional entity embeddings, and evaluate on synthetic datasets with known distributions.

\textbf{Masked Autoencoding.} Masked autoencoding (MAE)~\citep{he2021maskedautoencodersscalablevision} has emerged as a powerful self-supervised training signal across domains, and its adaptation to imputation is natural. \textsc{ReMasker}~\citep{du2023remaskerimputingtabulardata} extends MAE to tabular data by repeatedly re-masking observed entries during training and reconstructing them, showing strong results on benchmark datasets. Our work builds on this line but differs in key respects: we frame imputation as conditional marginal prediction rather than regression to observed entries, enabling uncertainty-aware downstream decisions; we introduce compositional entity embeddings to capture structured variables; and we use an encoder-only design without positional order. While \textsc{ReMasker} validates the potential of masked autoencoding for imputation, our approach extends it to a setting that supports conditional distribution prediction over structured variables.

\textbf{Active feature acquisition.} A related
line of work trains agents to decide which
features to observe before making a
prediction~\citep{doi:10.1287/mnsc.1080.0952,
li-et-al-afa, yin2020reinforcement}. These methods
typically combine a classifier with a sequential
acquisition policy optimizing a joint objective.
Our contribution is orthogonal: we produce the
calibrated conditional marginals that are a
prerequisite for VOI computation, rather than the
acquisition policy itself. Combining the \textsc{Marformer}
with a VOI-based acquisition policy is a natural
direction for future work.

Prior generative imputation methods represent distributions over missing values
\citep{yoon2018gainmissingdataimputation,mattei2019miwaedeepgenerativemodelling,
nazabal2020handlingincompleteheterogeneousdata}, while tabular methods typically operate on
a fixed set of columns \citep{du2023remaskerimputingtabulardata,
caruso2025imputationmethodtransformerbasedmodel}. MICE uses a collection of conditional models
in an iterative imputation procedure \citep{JSSv045i03}. Our approach directly predicts
conditional marginals with shared neural parameters and relational structure among variables
and entities.

\section{Discussion and Limitations}
\label{sec:conclusions}

We introduced the \textsc{Marformer}, a Transformer that
directly predicts conditional marginal
distributions over missing values. It replaces inference in a specified generative model with a single forward pass learned
from incomplete training data. Across
three synthetic domains with known ground truth,
the \textsc{Marformer} can match or outperform methods with
access to the correct generative family, although
a gap remains to Stan Oracle in the synthetic
annotation domain. On real data, the \textsc{Marformer}
improves overall with training scale and
outperforms the evaluated baselines at the largest
training size, while running at a fraction of
Stan's cost.

The \textsc{Marformer} has three notable limitations. First,
our justification for masked autoencoding assumes
MCAR training instances. If
missingness is informative, even in the weaker MAR
sense, where observation depends on other observed
values, artificially masked training examples need no
longer represent the same conditional distribution
as naturally missing ones, and learned conditional
distributions may be systematically biased.
Second, the \textsc{Marformer} predicts conditional marginals without enforcing their compatibility
with a coherent joint distribution. Third, calibration matters for active learning:
overconfident conditional predictions can lead to overconfident VOI estimates and suboptimal
acquisition decisions. The layerwise analysis in \cref{sec:logit-lens} illustrates how
calibration develops through the model's computation.

Deep models that predict output distributions, like $\ptheta$, sometimes overfit to the training examples, leading them to predict low-entropy distributions in general \citep{wongtoi2024pathologies}.  Our experiments show that calibration can deteriorate with limited training data (see \cref{fig:dom3-cal,fig:llmrubric-cal}).  Overfitting may be reduced by our $L_2$ regularization and early stopping,
or by the fact that random masking generates exponentially many training examples.  Still, the model could generalize poorly to test examples that are unlike any training examples.  Consider MCAR bandit data, in which every instance has $\vr=\{1\}$ or $\vr=\{2\}$.  Half of the test examples ask to predict $X_2$ from $X_1$, but there are no training examples of this form, only predicting $X_2$ from $\emptyset$.  Whether $X_2$ depends on $X_1$ cannot be resolved by the training data, but is wholly determined by the inductive bias of the learner.  We would like to ensure that the model will yield a high-entropy predictive distribution on inputs where the prediction is underconstrained by the training data.
If $L_2$ regularization does not achieve this, it might help to add a second regularizer to \cref{eq:trainreg} that explicitly rewards predictive entropy on held-out real or synthetic inputs from the test distribution $P_1^*$.  Ensemble learning is another way to achieve this goal, by returning a mixture of plausible predictive distributions \cite[e.g.,][]{lakshminarayanan2017simple}.  In the example, we might be able to predict $X_2$ with low entropy if we knew whether it was positively or negatively correlated with $X_1$, but since we do not know, we take a mixture of many possibilities, which has high entropy.

\bibliography{main}
\bibliographystyle{tmlr-inlinelinks}

\newpage

\appendix

\section{Results in MAR Setting}
\label{sec:exp-mar}

\begin{figure}[h!]
\begin{subfigure}{0.48\linewidth}
   \centering
   \includegraphics[width=\linewidth]{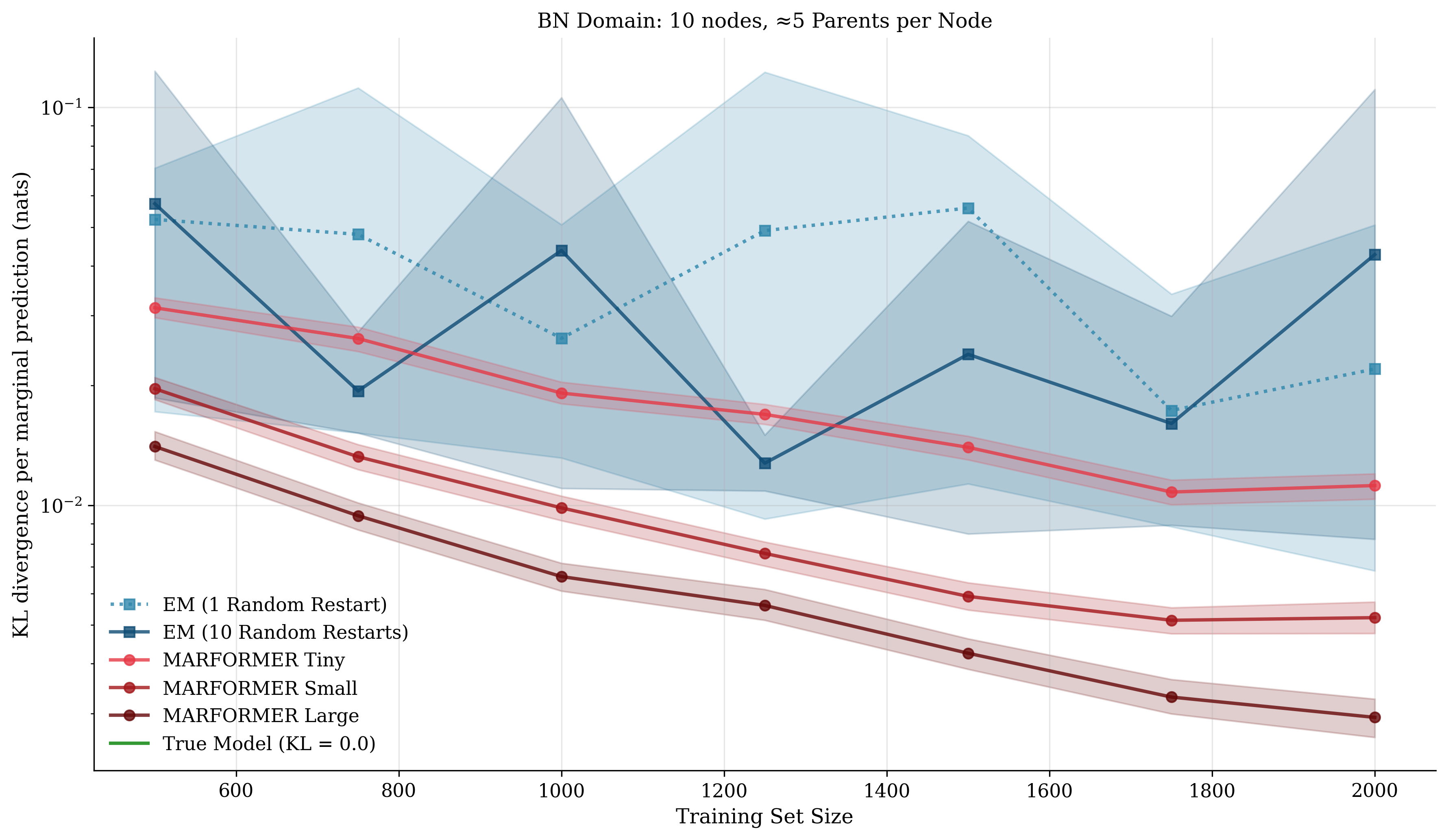}
   \caption{Domain 1 (BN): Sequential MAR}
   \label{fig:kl-bn-mar}
 \end{subfigure}
 \hfill
 \begin{subfigure}{0.48\linewidth}
   \centering
   \includegraphics[width=\linewidth]{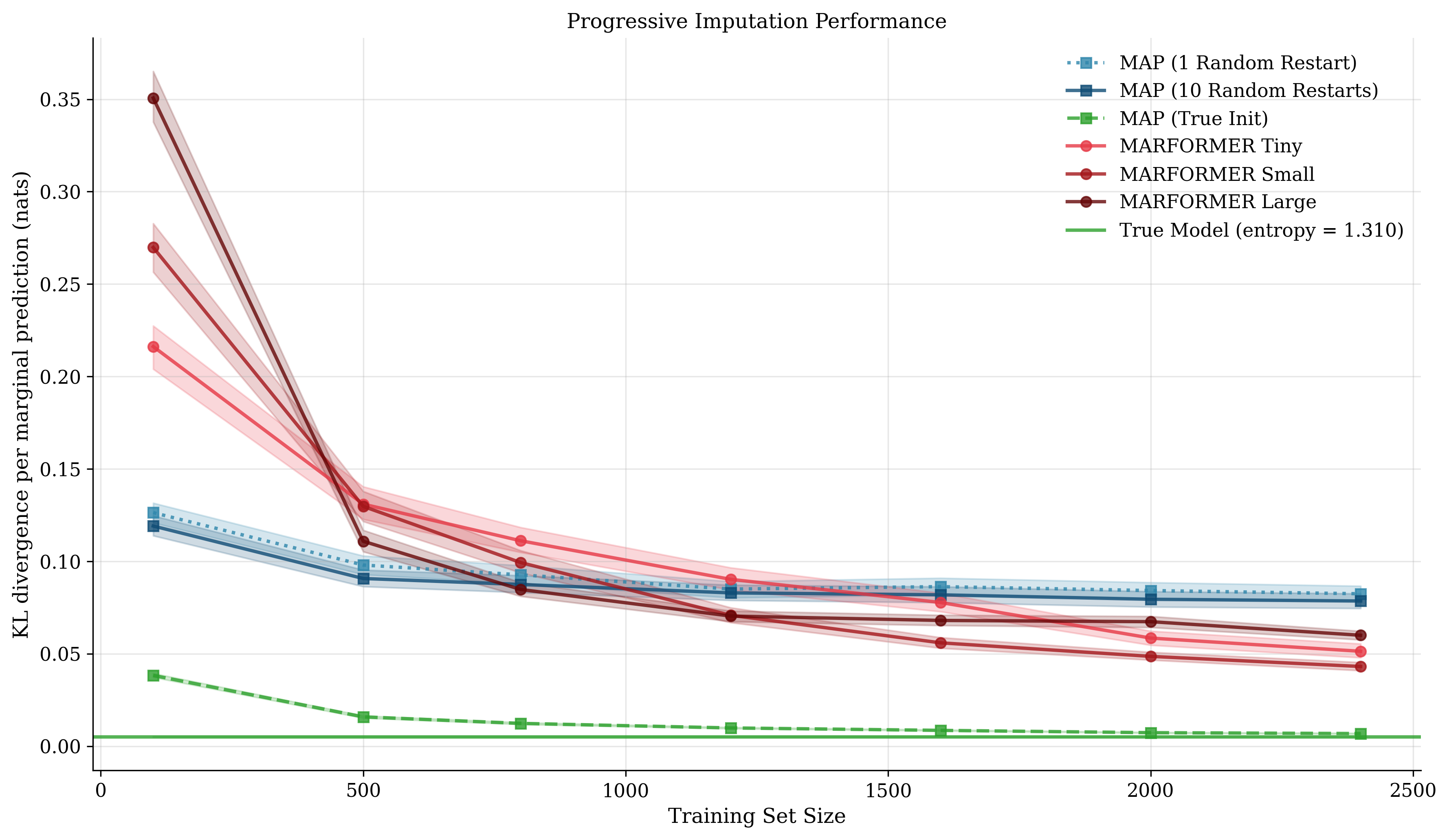}
   \caption{Domain 2 (Gaussian): Sequential MAR}
   \label{fig:kl-mg-mar}
 \end{subfigure}
 \caption{Learning curves, similar to \cref{fig:kl} but where $\Train$ is MAR rather than MCAR.  Despite the biased training examples, performance is qualitatively similar, though loss for all methods is higher in Domain 1.}\label{fig:kl-mar}
\end{figure}

\begin{figure}[h!]
  \centering
  \begin{subfigure}{0.48\linewidth}
    \centering
    \includegraphics[width=\linewidth]{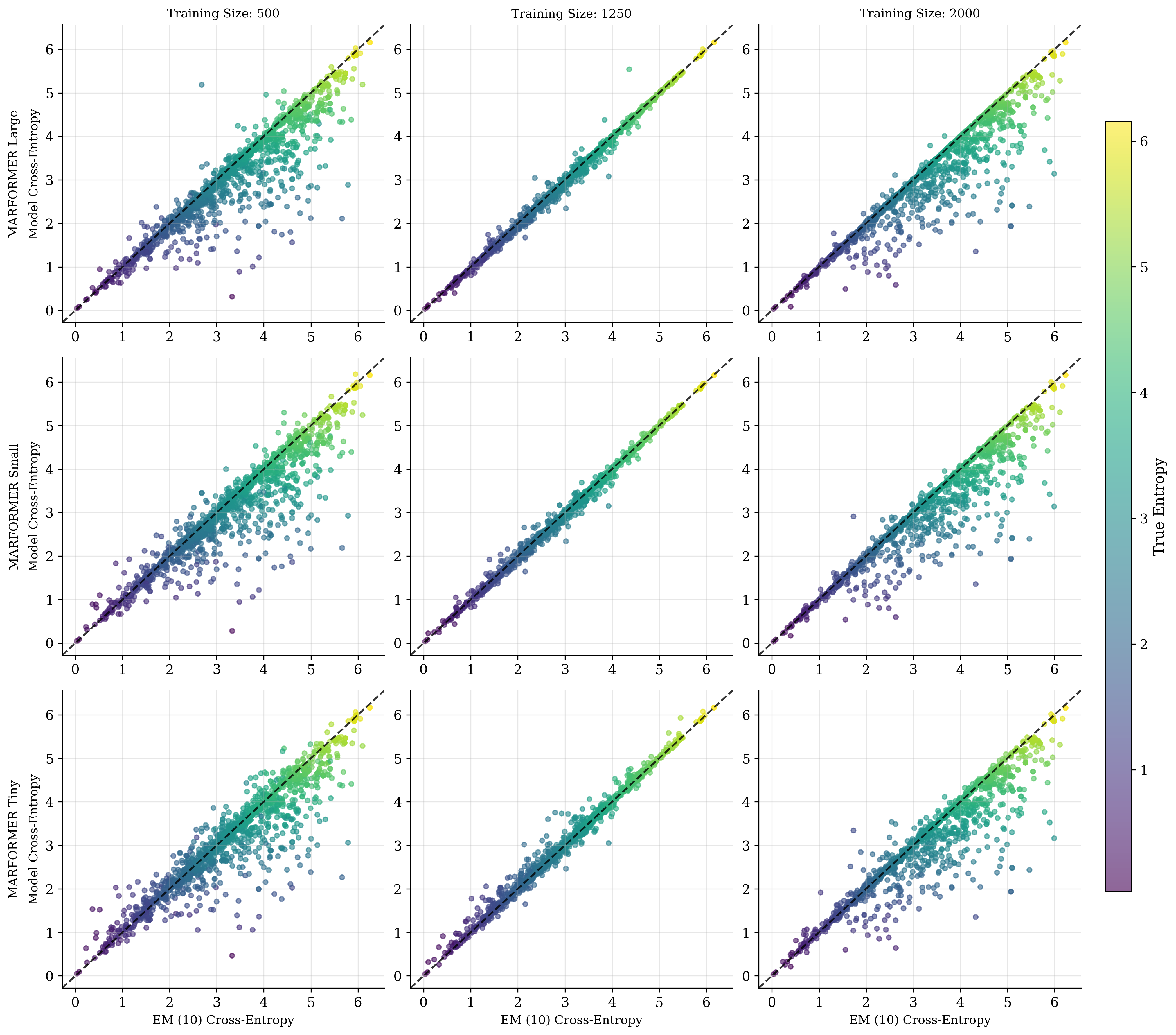}
    \caption{Domain 1 (Sequential MAR)}
    \label{fig:scatter-bn-mar}
  \end{subfigure}
  \hfill
  \begin{subfigure}{0.48\linewidth}
    \centering
    \includegraphics[width=\linewidth]{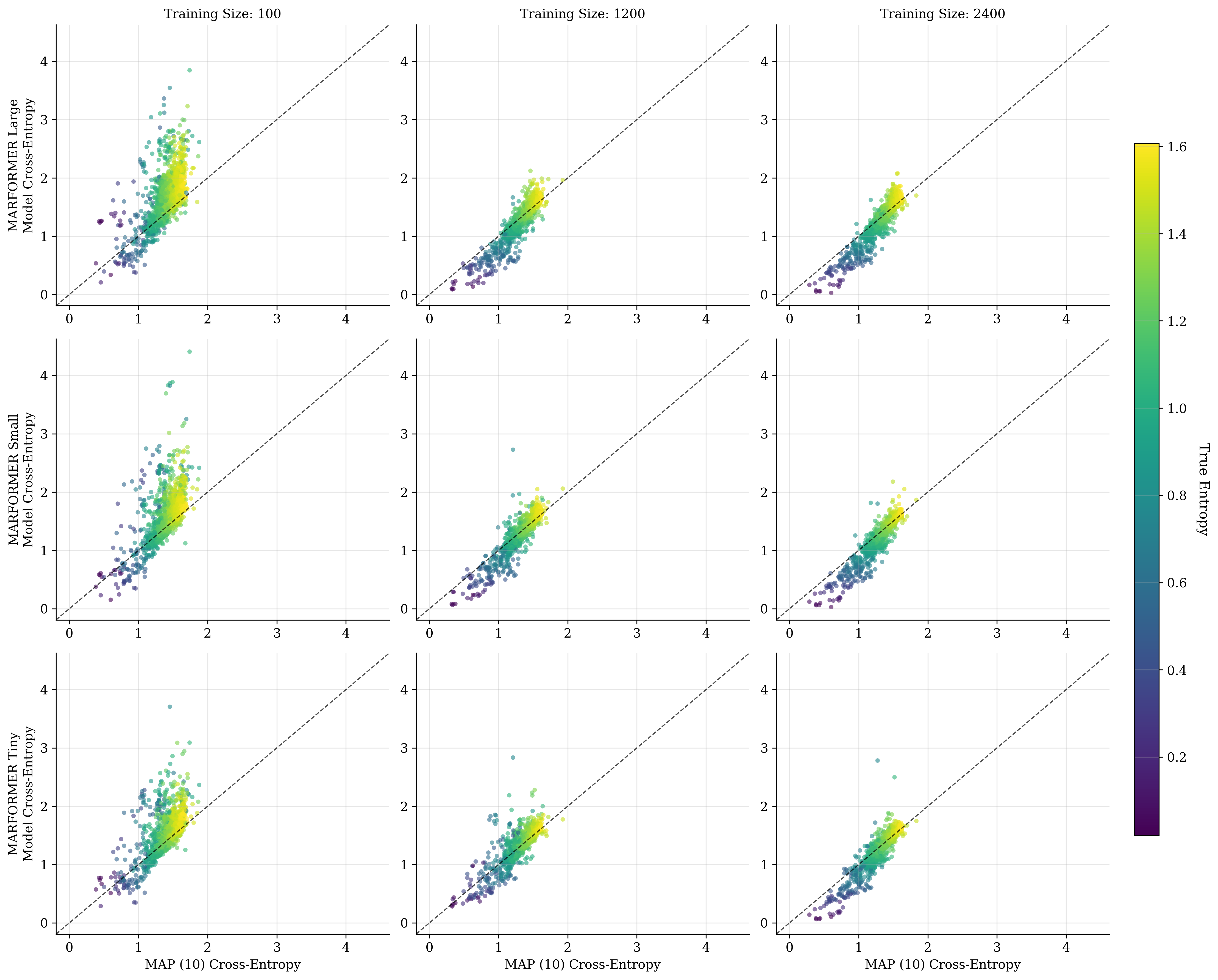}
    \caption{Domain 2 (MAR)}
    \label{fig:scatter-mg-mar}
  \end{subfigure}
  \caption{Prediction-by-prediction comparison of two methods, similar to \cref{fig:scatter} but where the models were trained on MAR rather than MCAR data.}
  \label{fig:scatter-mar}
\end{figure}

We rerun the comparisons of \cref{sec:results-bn} (Domain 1) using $\Train$ and $\Test$ datasets that are MAR but not MCAR. Training remains valid for MAP-EM, but the naive masking method that we use for \textsc{Marformer} training in this paper now produces biased examples.

Specifically, we use a \localdefn{sequential MAR} missingness mechanism, which has the form
\begin{align}
P^*(\vR=\vr \mid \vX=\vx)
&= \prod_{i\in\vi} P^*(R_i=r_i \mid \vR_{\pa(i)}=\vr_{\pa(i)}, \vX_{\pa(i)\cap\vr}=\vx_{\pa(i)\cap\vr})
\end{align}
Similarly to \cref{sec:true-bn}, each factor has the form of a logistic regression model. Here the model for factor $i$ has two binary features for each $j\in\pa(i)$, indicating whether $x_j$ is missing ($j\notin\vr$) and whether $x_j$ is observed to be 1 ($j\in\vr$ and $x_j=1$).  It also has a bias term.  The $\sum_{i\in\vi} (1+2|\pa(i)|)$ parameters are drawn i.i.d.\@ from $\mathcal{N}(0,1)$.

Owing to the symmetry of this construction, about half of the variables on any instance are missing, just as in the MCAR data of \cref{sec:data-bn}.

We similarly rerun the comparisons of \cref{sec:results-mg} (Domain 2) under MAR missingness.

For \textsc{Marformer} training, we apply the same simple masking mechanism as in \cref{sec:marformer-bn}, in which each variable is masked with independent probability of 0.15.

  Results are shown in \cref{fig:kl-mar,fig:scatter-mar}.

\section{Domain Specifications}
\label{sec:domain-specs}

The following two domain algorithms provide complete specifications of the data-generating processes for Domains 1 and 2, summarizing the generative models, missingness mechanisms, and modeling setups described in \cref{sec:true-bn,sec:true-mg}. The color coding follows the structure of the algorithms: \colorbox{lightgray!30}{grey} for given quantities (data source family), \colorbox{red!15}{red} for the data generation process, and \colorbox{blue!15}{blue} for the modeling choices.

\begin{algorithm}[htbp]
\renewcommand{\thealgorithm}{1}
\renewcommand{\algorithmname}{Domain}
\caption{BN Domain: Data Source, Generation and Modeling}
\label{alg:bn}\label{alg:domain-bn}

\colorbox{lightgray!30}{%
\begin{minipage}{0.95\textwidth}
\textbf{\textsc{Given}}
\begin{enumerate}
  \item Domain Family: DAGs with $A$ binary attributes and target parent density $c$.
  \item Generate Multiple \textit{Data Sources}: For each data source $k$, sample $G^{(k)} \sim \texttt{DAG}([1,\ldots,A], c)$ and $\Psi^{(k)} = \{\psi_{a \mid \pa(a)}^{(k)} : a \in [1,A]\}$.
\end{enumerate}
\end{minipage}%
}

\vspace{0.4em}

\colorbox{red!15}{%
\begin{minipage}{0.95\textwidth}
\textbf{\textsc{Data Generation}} (for each data source)
\begin{enumerate}
  \item $P^*(\mathbf{X}, \mathbf{R}) = P^*(\mathbf{X} \mid \Psi^{(k)}, G^{(k)}) \cdot P^*(\mathbf{R})$
  \item For $n \in [1, N+M]$: $X_a\psup{n} \mid X_{\pa(a)}\psup{n} \sim \text{Bernoulli}(\psi_{a|\pa(a)}^{(k)})$, $R_a\psup{n} \sim \text{Bernoulli}(0.5)$
  \item $\obs{\vx}\psup{n} = \{X_a\psup{n} : R_a\psup{n} = 1\}$ (Naturally Observed)
  \item Split: Training $(\obs{\vx}\psup{n}, \mathbf{r}\psup{n})$ for $n \in [1,N]$, Test $(\obs{\vx}^{(m)}, \mathbf{r}^{(m)})$ for $m \in [N+1,N+M]$
\end{enumerate}
\end{minipage}%
}

\vspace{0.4em}

\colorbox{blue!15}{%
\begin{minipage}{0.95\textwidth}
\textbf{\textsc{Modeling}}

EM Baseline:
\begin{enumerate}
  \item Input: $(\obs{\vx}\psup{n}, \mathbf{r}\psup{n})$ for $n \in [1,N]$
  \item EM algorithm with multiple restarts $\rightarrow q_\phi$
  \item Evaluate: $q_\phi(X_a \mid \obs{\vx}^{(m)})$
\end{enumerate}

\textsc{Marformer}:
\begin{enumerate}
  \item Input: $\obs{\vx}\psup{n}$ for $n \in [1,N]$
  \item Artificial masking: $\mask{\vx}\psup{n} \sim \text{Masking}(\obs{\vx}\psup{n}, 0.15)$
  \item Learn $\ptheta(\obs{\vx}\psup{n} \mid \mask{\vx}\psup{n})$
  \item Evaluate: $\ptheta(X_a \mid \obs{\vx}^{(m)})$
\end{enumerate}
\end{minipage}%
}
\end{algorithm}

\begin{algorithm}[ht]
\renewcommand{\thealgorithm}{2}
\renewcommand{\algorithmname}{Domain}
\caption{Discretized Multivariate Gaussian Domain: Data Source, Generation and Modeling}
\label{alg:mg}

\colorbox{lightgray!30}{%
\begin{minipage}{0.95\textwidth}
\textbf{\textsc{Given}}
\begin{enumerate}
  \item Domain Family: Multivariate Gaussian distribution with $A$ dimensions (attributes) and $C$ bins for each dimension.
  \item Generate Multiple \textit{Data Sources}: For each data source $k$, sample ${\Sigma^{-1}}^{(k)}$ from the softly constrained Wishart prior $W_A(I,\nu)$ described in \cref{sec:true-mg}, and sample $\vec p_a^{(k)}\sim\mathrm{Dirichlet}(\alpha/C,\ldots,\alpha/C)$ for each $a\in\{1,\ldots,A\}$. Set $q_{ac}^{(k)}=\sum_{r=1}^{c}p_{ar}^{(k)}$ for $c\in\{1,\ldots,C\}$.
\end{enumerate}
\end{minipage}%
}

\vspace{0.4em}

\colorbox{red!15}{%
\begin{minipage}{0.95\textwidth}
\textbf{\textsc{Data Generation}} (for each data source)
\begin{enumerate}
  \item $P^*(\mathbf{X}, \mathbf{R}) = P^*(\mathbf{X} \mid \Sigma^{(k)}, p_{ac}^{(k)}) \cdot P^*(\mathbf{R})$
  \item For $n \in [1, N+M]$:
    \begin{enumerate}
      \item $\vZ\psup{n} \sim \mathcal{N}(0, \Sigma^{(k)})$
      \item Set $u_a\psup{n}=\Phi(Z_a\psup{n}/\sqrt{\Sigma_{aa}^{(k)}})$, where $\Phi$ is the standard normal CDF, and $X_a\psup{n}=\min\{c\in\{1,\ldots,C\}:u_a\psup{n}<q_{ac}^{(k)}\}$.
      \item $R_a\psup{n} \sim Bernoulli(0.5)$
    \end{enumerate}
  \item $\obs{\vx}\psup{n} = \{X_a\psup{n} : R_a\psup{n} = 1\}$ (Naturally Observed)
  \item Split: Training $(\obs{\vx}\psup{n}, \mathbf{r}\psup{n})$ for $n \in [1,N]$, Test $(\obs{\vx}^{(m)}, \mathbf{r}^{(m)})$ for $m \in [N+1,N+M]$
\end{enumerate}
\end{minipage}%
}

\vspace{0.4em}

\colorbox{blue!15}{%
\begin{minipage}{0.95\textwidth}
\textbf{\textsc{Modeling}}

MAP Baseline:
\begin{enumerate}
  \item Input: $(\obs{\vx}\psup{n}, \mathbf{r}\psup{n})$ for $n \in [1,N]$
  \item MAP model with Genz's Algorithm~\citep{83355a14-2760-31d3-afd2-16d5bbbc1183} evaluating the observed likelihood $\rightarrow q_\phi$
  \item Evaluate: $q_\phi(X_a \mid \obs{\vx}^{(m)})$
\end{enumerate}

\textsc{Marformer}:
Same as \cref{alg:bn}
\end{minipage}%
}
\end{algorithm}

\section{Additional Results: Domain 3 (Synthetic)}
\label{sec:additional-domain3}

\begin{figure}[h!]
 \centering
 \begin{subfigure}{0.48\linewidth}
   \centering
   \includegraphics[width=\linewidth]{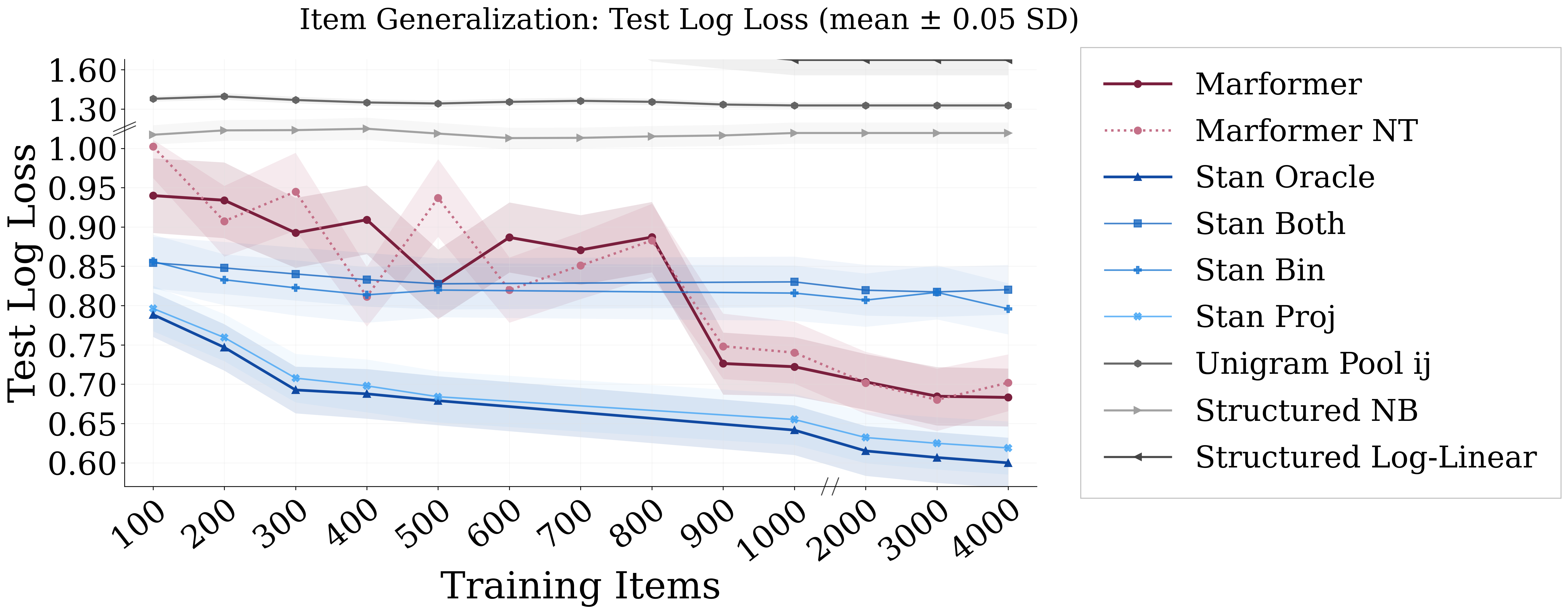}
   \caption{Item Generalization}
   \label{fig:dom3-full-ll-item}
 \end{subfigure}
 \hfill
 \begin{subfigure}{0.48\linewidth}
   \centering
   \includegraphics[width=\linewidth]{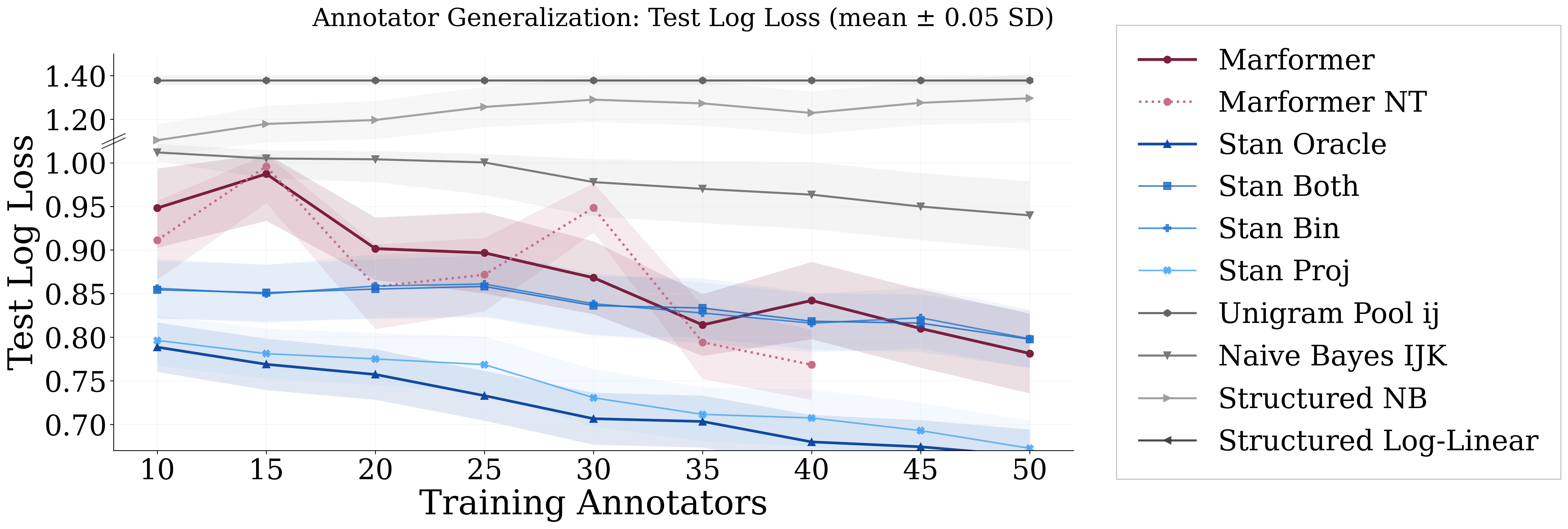}
   \caption{Judge Generalization}
   \label{fig:dom3-full-ll-anno}
 \end{subfigure}
 \caption{Full log-loss curves including all baselines. Empirical baselines (Unigram Pool, Structured NB, Log-Linear) generally have higher log-loss than the Marformer and Stan variants at larger training sizes in both settings.}
 \label{fig:dom3-full-ll}
\end{figure}

\begin{figure}[h!]
 \centering
 \begin{subfigure}{0.48\linewidth}
   \centering
   \includegraphics[width=\linewidth]{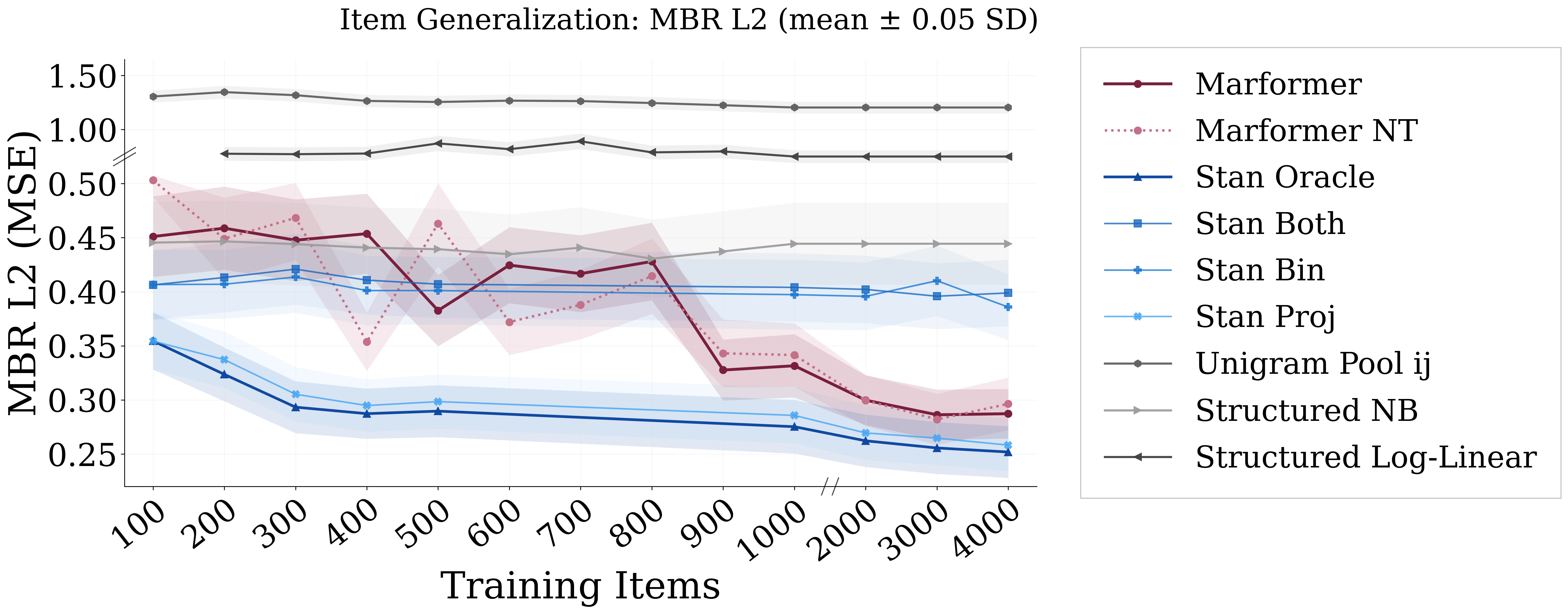}
   \caption{Item Generalization}
   \label{fig:dom3-full-mbr-item}
 \end{subfigure}
 \hfill
 \begin{subfigure}{0.48\linewidth}
   \centering
   \includegraphics[width=\linewidth]{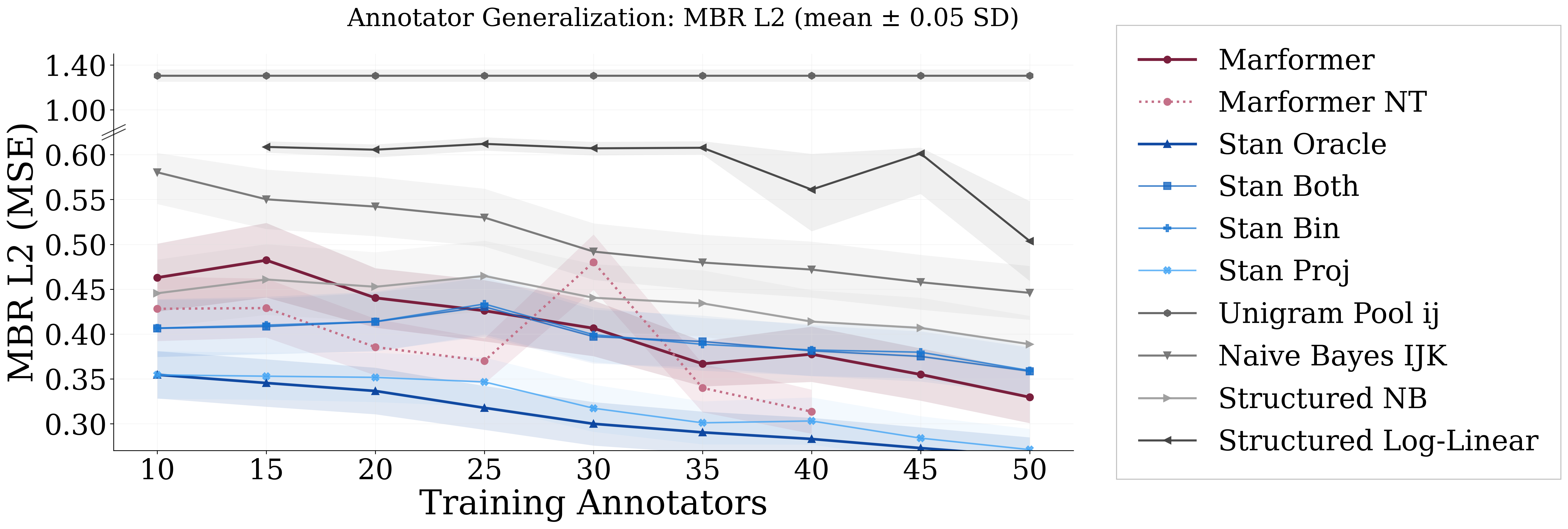}
   \caption{Judge Generalization}
   \label{fig:dom3-full-mbr-anno}
 \end{subfigure}
 \caption{Full MBR-L2 curves including all baselines. The empirical baselines generally have higher squared error than the Marformer and Stan variants at larger training sizes in both settings.}
 \label{fig:dom3-full-mbr}
\end{figure}

\begin{figure}[h!]
 \centering
 \begin{subfigure}{0.48\linewidth}
   \centering
   \includegraphics[width=\linewidth]{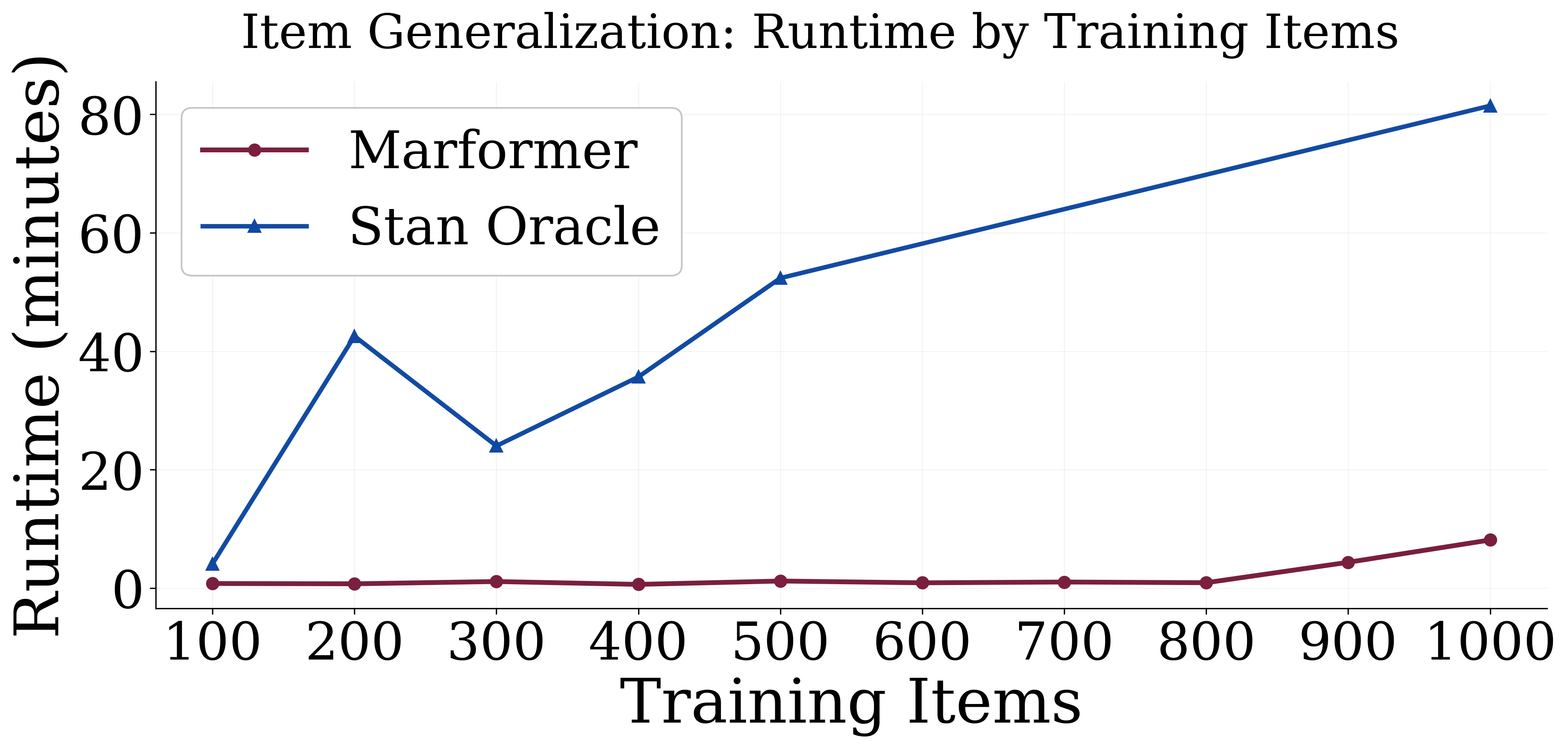}
   \caption{Item Generalization}
   \label{fig:dom3-runtime-item}
 \end{subfigure}
 \hfill
 \begin{subfigure}{0.48\linewidth}
   \centering
   \includegraphics[width=\linewidth]{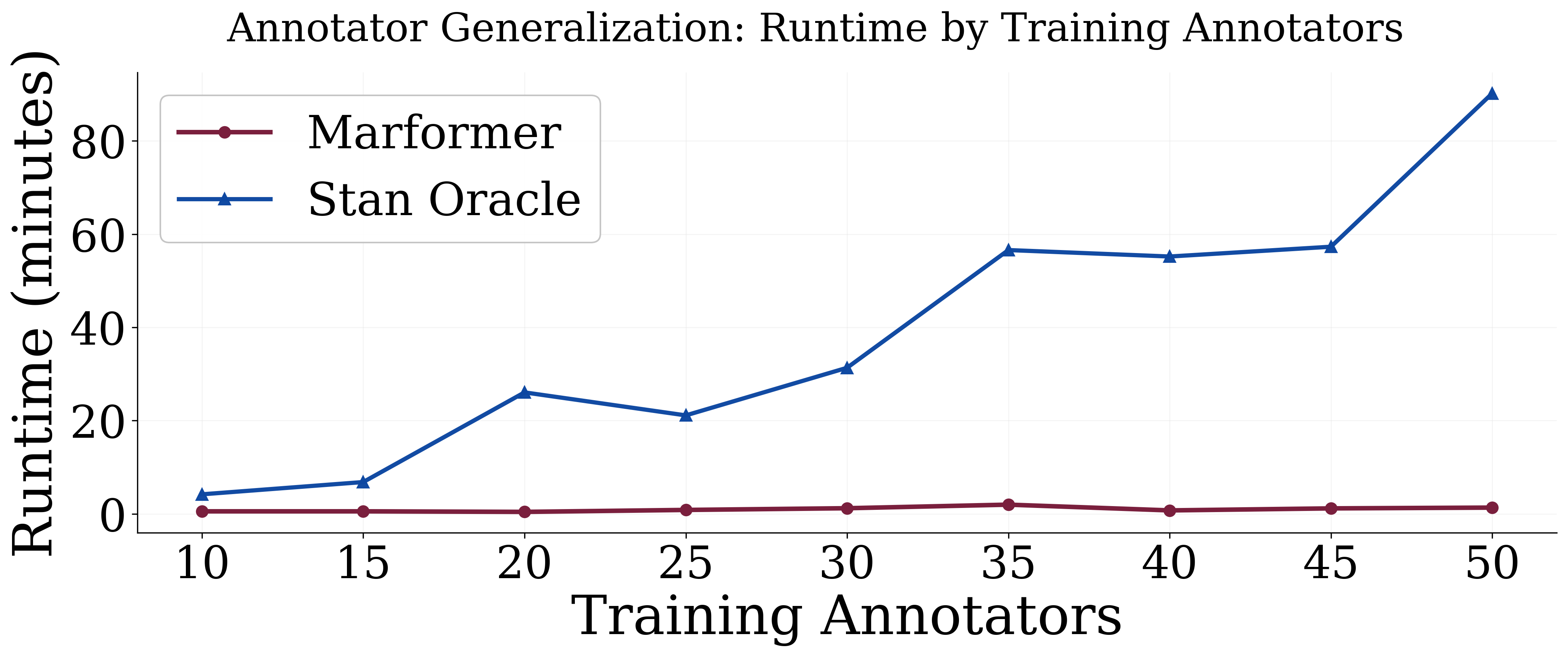}
   \caption{Judge Generalization}
   \label{fig:dom3-runtime-anno}
 \end{subfigure}
 \caption{Synthetic annotation runtime (minutes). \textsc{Marformer} costs are hypothetical estimates
of a shortened validation search plus the measured refit, using the accounting in
\cref{sec:results-annot}; Stan costs cover its
measured HMC fit.}
 \label{fig:dom3-runtime}
\end{figure}

\subsection*{Stan HMC configuration}
\label{sec:stan-convergence}
The predictive fits use one HMC chain, 300 warmup iterations and 500 post-warmup samples,
\texttt{adapt\_delta}=0.85, \texttt{max\_treedepth}=12, and random initialization.

\subsection{Stan Posterior Predictive Calculation}
\label{sec:stan-posterior-predictive}

For completeness, we give the calculation used to obtain the categorical predictive distribution from each posterior sample in the Stan CPM.  For posterior sample $\vphi^{(m)}$, let $z_{\ratesub}^{(m)}$, $s_{\judgeidx\critidx}^{(m)}$, $\tau_{\judgeidx\critidx,c}^{(m)}$, and $\sigma_{\mathrm{meas}}^{(m)}$ denote the corresponding latent score, normalization scale, ordinal thresholds, and measurement-noise parameter defined by the CPM in \cref{sec:true-annot}.  The probability assigned to category $c$ is
\[
q_{\vphi^{(m)}}(X_{\ratesub}=c\mid\obs{\vx})
=
\Phi
\left(
\frac{
\tau_{\judgeidx\critidx,c+1}^{(m)}
-
z_{\ratesub}^{(m)}/s_{\judgeidx\critidx}^{(m)}
}{
\sigma_{\mathrm{meas}}^{(m)}/s_{\judgeidx\critidx}^{(m)}
}
\right)
-
\Phi
\left(
\frac{
\tau_{\judgeidx\critidx,c}^{(m)}
-
z_{\ratesub}^{(m)}/s_{\judgeidx\critidx}^{(m)}
}{
\sigma_{\mathrm{meas}}^{(m)}/s_{\judgeidx\critidx}^{(m)}
}
\right).
\]
Averaging these probabilities across posterior samples gives the posterior predictive marginal reported in the main text.

\section{Additional Results: Domain 3 (Real Data)}
\label{sec:additional-domain3-real}

\begin{figure}[h!]
  \centering
  \includegraphics[width=0.6\linewidth]{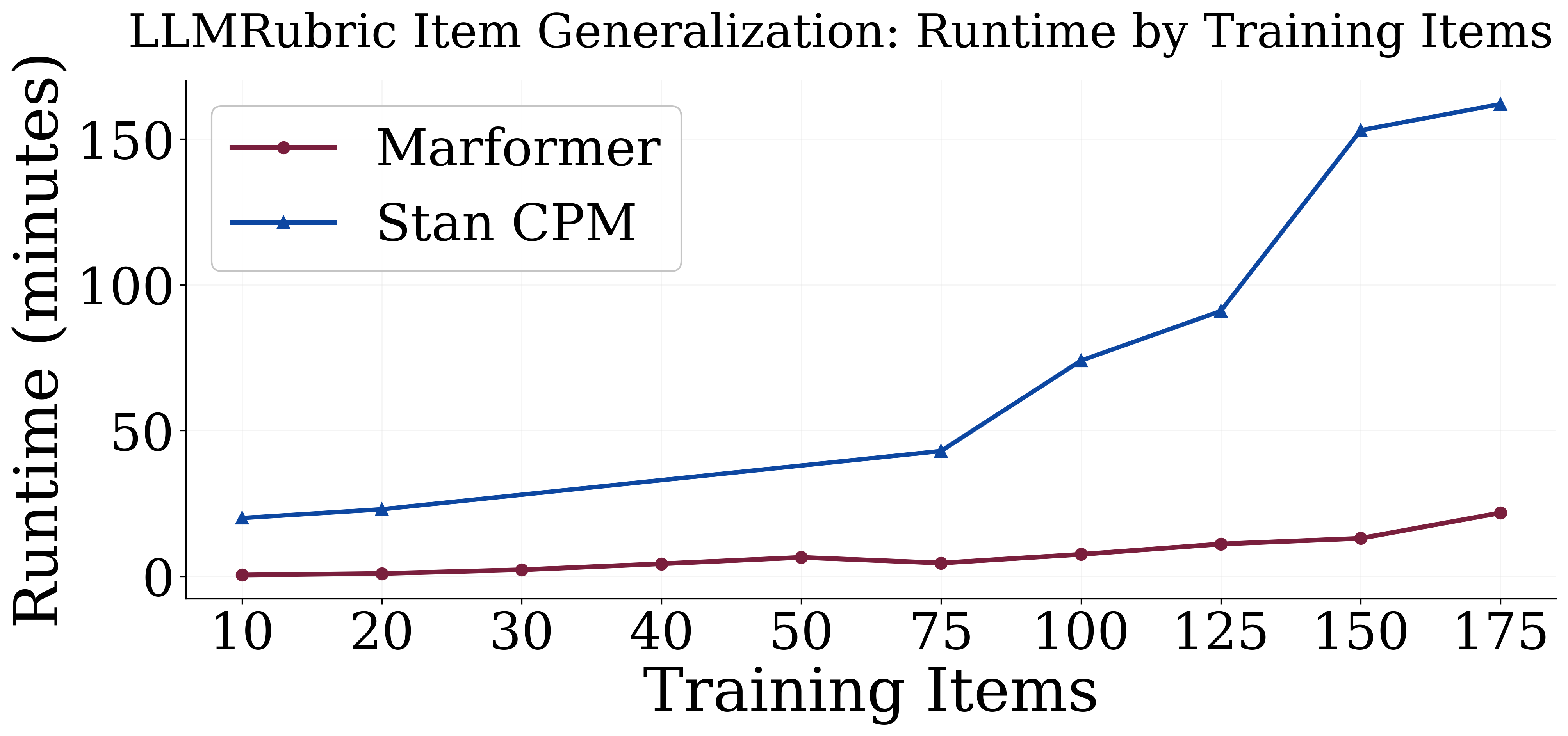}
  \caption{Wall-clock runtime (minutes) for \textsc{Marformer} and Stan CPM on LLMRubric as a function of training items. Stan CPM is run with the HMC configuration described in \cref{sec:stan-convergence}. At $\nitem_{\mathrm{train}}=175$, the recorded training times are approximately 22 minutes
for \textsc{Marformer} and 162 minutes for Stan CPM. The \textsc{Marformer} uses best-validation checkpoint
selection without refitting.}
  \label{fig:llmrubric-runtime}
\end{figure}

\subsection*{Stan CPM Posterior Hyperparameters}
\label{sec:llmrubric-hyperparams}

\cref{tab:llmrubric-hyperparams} reports the posterior marginal distributions of the five Stan CPM scale hyperparameters, inferred from the full LLMRubric training set ($\nitem_{\mathrm{train}}=175$) with weakly informative priors. The relatively large posterior mean for $\sigma_v$ (${\approx}11.6$) compared to $\sigma_u$ (${\approx}2.1$) and $\sigma_\text{uit}$ (${\approx}1.2$) indicates that prototype-level variation is the dominant source of judge-to-judge differences, with criterion--prototype interactions at a much smaller scale. The small measurement noise $\sigma_\text{meas}$ (${\approx}0.38$) suggests that the ordinal thresholds capture most of the rating variability. These posterior distributions informed the prior scale choices for synthetic data generation in \cref{sec:true-annot}.

\begin{table}[H]
\centering
\small
\caption{Posterior marginal summaries of Stan CPM scale hyperparameters estimated from LLMRubric ($\nitem_{\mathrm{train}}=175$, weakly informative priors). Mean, SD, and percentiles are computed from HMC posterior samples.}
\label{tab:llmrubric-hyperparams}
\begin{tabular}{llrrrrr}
\toprule
\textbf{Parameter} & \textbf{Role} & \textbf{Mean} & \textbf{SD} & \textbf{5th \%ile} & \textbf{Median} & \textbf{95th \%ile} \\
\midrule
$\sigma_u$ & Criterion vector scale & 2.127 & 0.475 & 1.542 & 2.014 & 3.065 \\
$\sigma_v$ & Prototype vector scale & 11.588 & 1.713 & 9.028 & 11.445 & 14.564 \\
$\sigma_\text{uit}$ & Criterion--prototype interaction scale & 1.242 & 0.254 & 0.874 & 1.234 & 1.716 \\
$\sigma_\text{meas}$ & Measurement noise & 0.383 & 0.043 & 0.317 & 0.381 & 0.453 \\
$\kappa$ & Rating scale concentration & 5.201 & 0.417 & 4.563 & 5.186 & 5.958 \\
\bottomrule
\end{tabular}
\end{table}

\clearpage
\section{Logit Lens Analysis}
\label{sec:logit-lens}

At each layer $\ell$, the \textsc{Marformer} maintains a
parameter slice $\vpsi_a\psup{\ell}$, which is read
off at the final layer $L$ as the predicted rating
distribution. For missing ratings, this slice
initially encodes a uniform distribution; attention
layers then update it using the observed ratings.
To examine how these predictions develop, we apply
logit and tuned lenses~\citep{belrose2023tunedlens}
to the transductive Domain~3 model trained at
$\nitem_{\mathrm{train}}=500$, with $L=8$ layers.
We evaluate all readouts on the same 316 missing
ratings in the test corner used for the main
item-generalization results, as detailed in \cref{sec:data-annot}.

We construct three readouts at each layer
$\ell\in\{0,\ldots,L\}$. The \defn{Param Lens}
interprets the parameter slice directly, applying
softmax to its $C=4$ rating logits. The
\defn{Tuned Param Lens} instead learns an affine
map from that slice to rating logits, minimizing KL
divergence to the final model's predicted
distributions. This asks whether an intermediate
representation can already recover those
predictions through a learned readout. The
\defn{Tuned 80d-Lens} uses the same procedure on
the full $d=80$-dimensional residual stream,
$\vh_a\psup{\ell}=[\vec{f}_a\psup{\ell};
\vpsi_a\psup{\ell}]$, to examine what the feature
slice adds. The \textsc{Marformer} is frozen while separate
probes are fitted for each layer. Probe inputs are
standardized, and the probes are trained for 300
epochs with AdamW at learning rate $10^{-2}$ on a
held-out set of training-missing ratings. Their training
targets are the final model's distributions,
not the true rating labels.

\begin{figure}[h!]
  \centering
  \includegraphics[width=0.85\linewidth]{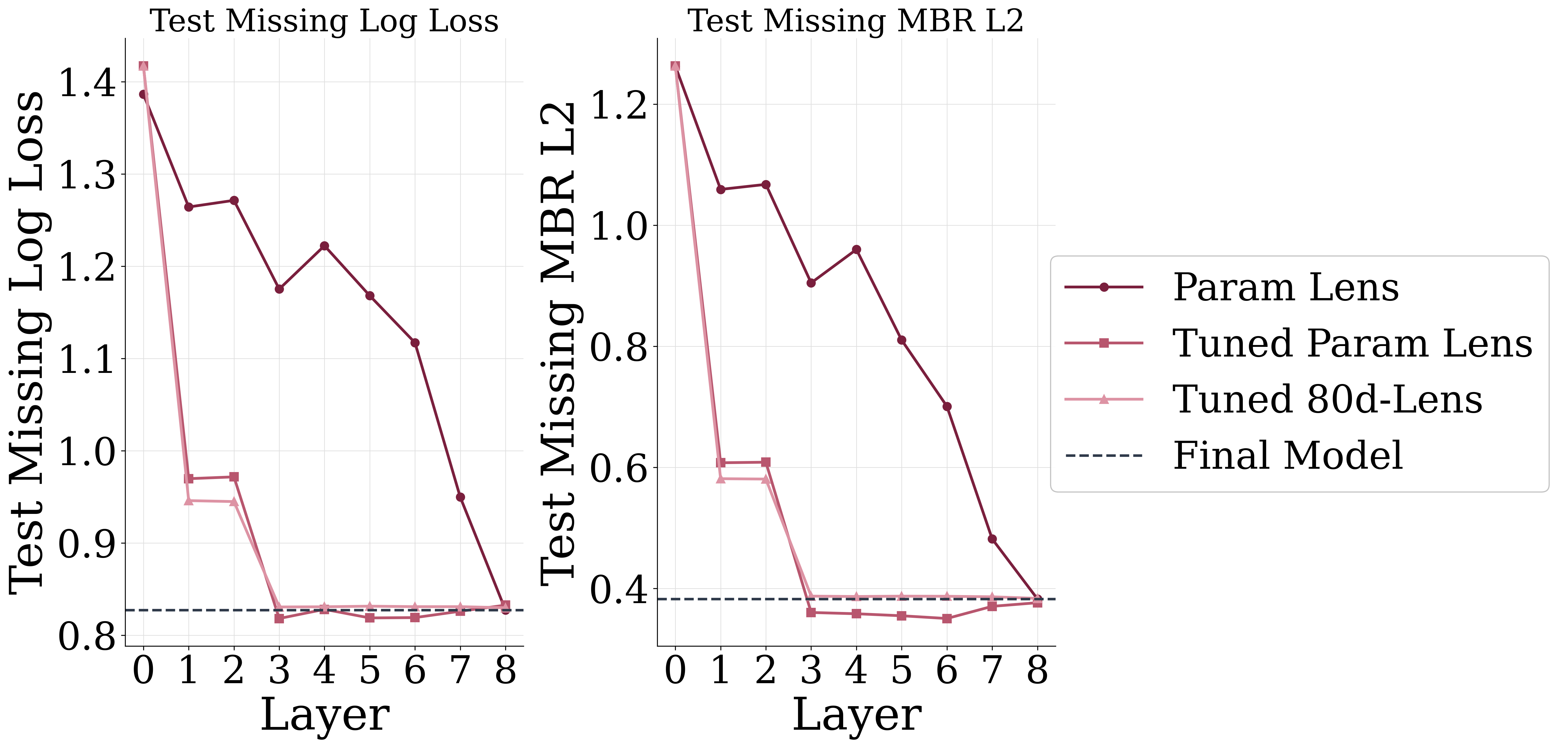}
  \caption{Log-loss and MBR-L2 for three
layerwise readout probes, evaluated on 316 missing test
ratings at $\nitem_{\mathrm{train}}=500$. Layer 0
is the input representation before attention, and
layer 8 is the final layer. The dashed line gives
the final model's performance; the raw Param Lens
at layer 8 reproduces this output.}
  \label{fig:logit-lens-metrics}
\end{figure}

\cref{fig:logit-lens-metrics} suggests that early
layers aggregate evidence in the latent representations,
while later layers gradually align this information
with the parameter coordinates.
After one attention layer, the Tuned 80d-Lens
achieves log-loss $0.946$, while the raw Param
Lens remains at $1.264$. The Tuned Param Lens
achieves $0.970$. Thus, useful predictions can
already be read from the hidden state even though
the raw parameter coordinates do not yet express
them. The full-state probe has slightly lower
loss than the parameter-only probe at this layer.

By layer 3, both tuned lenses have log-loss close
to the final model's $0.827$. From layers 3--8,
the Tuned 80d-Lens stays near $0.831$, whereas the
raw Param Lens improves overall but not
monotonically. In particular, its log-loss falls
from $1.117$ at layer 6 to $0.950$ at layer 7 and
$0.827$ at layer 8. The final two layers therefore
make substantial contributions to the raw
readout, even after a learned probe can recover
predictions close to the final model's. MBR-L2
shows the same broad pattern: the tuned readouts
approach the final model early, while the raw
readout continues to improve through later layers.

The two-stage improvement in the tuned readouts
suggests a possible message-passing interpretation.
One hypothesis is that the model performs two
rounds of a learned message-passing procedure.
Alternatively, the first attention layer may
aggregate direct evidence from observed ratings,
while subsequent layers incorporate indirect
evidence relayed through intermediate
representations, accounting for the further
improvement by layer 3.

\begin{figure}[h!]
  \centering
  \includegraphics[width=0.85\linewidth]{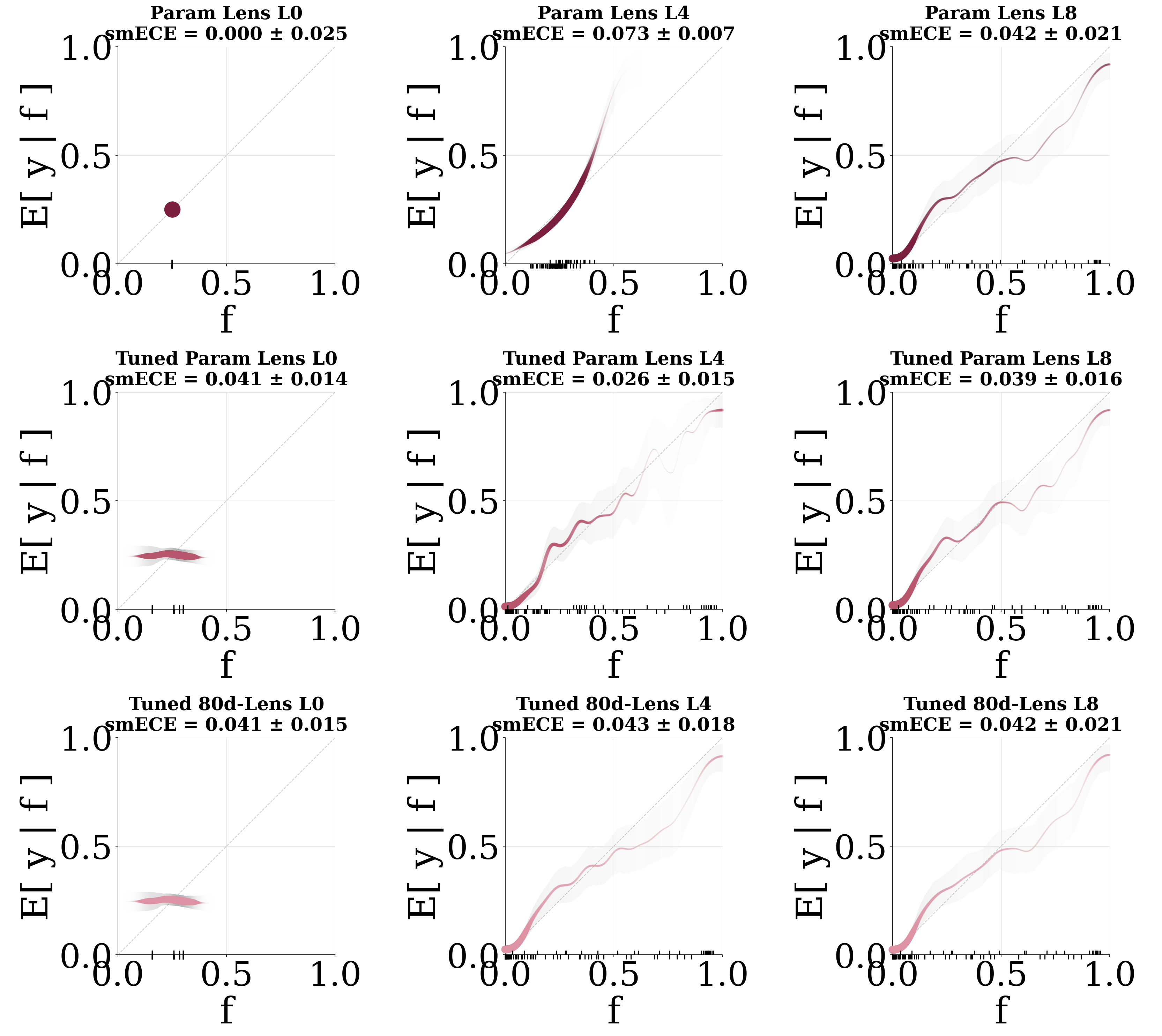}
  \caption{Calibration diagrams for the three
readouts (rows) at layers 0, 4 and 8 (columns), on
the same 316 missing test ratings. Predicted
probabilities for all four rating categories are
pooled. The diagonal indicates perfect
calibration; titles report smECE and \texttt{relplot}'s
reported calibration-error width from 500
bootstrap resamples. The layer-0 Param Lens
predicts $1/4$ for every category due to the initialization.}
  \label{fig:logit-lens-cal}
\end{figure}

Calibration also changes across depth, although all three readouts remain reasonably
well calibrated
(\cref{fig:logit-lens-cal}). At layer 4, the raw
Param Lens has smECE $0.073$, compared with
$0.026$ for the Tuned Param Lens and $0.043$ for
the Tuned 80d-Lens. At the output, the raw Param
Lens reaches $0.042$. Its smECE is not monotone
across layers: it ranges from $0.065$ to $0.078$
at layers 1--6, then falls to $0.063$ at layer 7
and $0.042$ at layer 8. As with log-loss, the last
two layers improve the raw predictions. The
uniform layer-0 readout has zero all-class smECE
but log-loss $\ln 4$, illustrating that
calibration alone does not measure how informative
a prediction is.

Together, these readouts show that predictions
close to the final model's can be recovered from
intermediate representations, while the raw
parameter slice requires further computation to
express them. This is consistent with the
\textsc{Marformer} learning an iterative refinement
procedure~\citep{pmlr-v15-stoyanov11a}: early
layers gather evidence, and later layers refine
the distribution that is read off directly at the
output.

\end{document}